\documentclass[final,5p,times,twocolumn]{elsarticle}

\usepackage{amsmath,amssymb,graphicx, multirow}
\usepackage{flushend}
\usepackage{float}
\usepackage{mathtools}
\usepackage{url}
\usepackage{color}
\newcommand{\tabincell}[2]{\begin{tabular}{@{}#1@{}}#2\end{tabular}} 

\begin{document}

\begin{frontmatter}

\title{Box-level Segmentation Supervised Deep Neural Networks for Accurate and Real-time Multispectral Pedestrian Detection}

\author[label1,label2]{Yanpeng Cao}
\ead{caoyp@zju.edu.cn}

\author[label2]{Dayan Guan}
\ead{11725001@zju.edu.cn}

\author[label2]{Yulun Wu}
\ead{3160105381@zju.edu.cn}

\author[label1,label2]{Jiangxin Yang}
\ead{yangjx@zju.edu.cn}

\author[label1,label2]{Yanlong Cao}
\ead{sdcaoyl@zju.edu.cn}

\author[label3]{Michael Ying Yang}
\ead{michael.yang@utwente.nl}

\address[label1]{State Key Laboratory of Fluid Power and Mechatronic Systems, School of  Mechanical Engineering, Zhejiang University, Hangzhou, China}
\address[label2]{Key Laboratory of Advanced Manufacturing Technology of Zhejiang Province, School of Mechanical Engineering,  Zhejiang University, Hangzhou, China}
\address[label3]{Scene Understanding Group, University of Twente, Hengelosestraat 99, 7514 AE Enschede, The Netherlands}

\begin{abstract}
Effective fusion of complementary information captured by multi-modal sensors (visible and infrared cameras) enables robust pedestrian detection under various surveillance situations (e.g. daytime and nighttime). In this paper, we present a novel box-level segmentation supervised learning framework for accurate and real-time multispectral pedestrian detection by incorporating features extracted in visible and infrared channels. Specifically, our method takes pairs of aligned visible and infrared images with easily obtained bounding box annotations as input and estimates accurate prediction maps to highlight the existence of pedestrians. It offers two major advantages over the existing anchor box based multispectral detection methods. Firstly, it overcomes the hyperparameter setting problem occurred during the training phase of anchor box based detectors and can obtain more accurate detection results, especially for small and occluded pedestrian instances. Secondly, it is capable of generating accurate detection results using small-size input images, leading to improvement of computational efficiency for real-time autonomous driving applications. Experimental results on KAIST multispectral dataset show that our proposed method outperforms state-of-the-art approaches in terms of both accuracy and speed.
\end{abstract}

\begin{keyword}
Multispectral data \sep Pedestrian detection \sep Deep neural networks \sep Box-level segmentation \sep Real-time application 
\end{keyword}

\end{frontmatter}

\section{INTRODUCTION}

Pedestrian detection has received much attention within the field of computer vision and robotics in recent years \cite{oren1997pedestrian, dalal2005histograms, dollar2012pedestrian, angelova2015pedestrian, geiger2012we, jafari2016real, cordts2016cityscapes, Zhang2017CVPR}. Given images captured in various real-world surveillance situations, pedestrian detectors are required to accurately locate human regions. It provides an important functionality to facilitate human-centric applications such as autonomous driving, video surveillance, and urban monitoring \cite{wu2016squeezedet, li2017unified, zhang2017towards, wang2014scene, li2017accurate, bu2005pedestrian, shirazi2017looking}.


\begin{figure}[!ht]
	\centering
	\begin{minipage}{0.48\linewidth}
		\centering
		{\includegraphics[width=1\linewidth,clip]{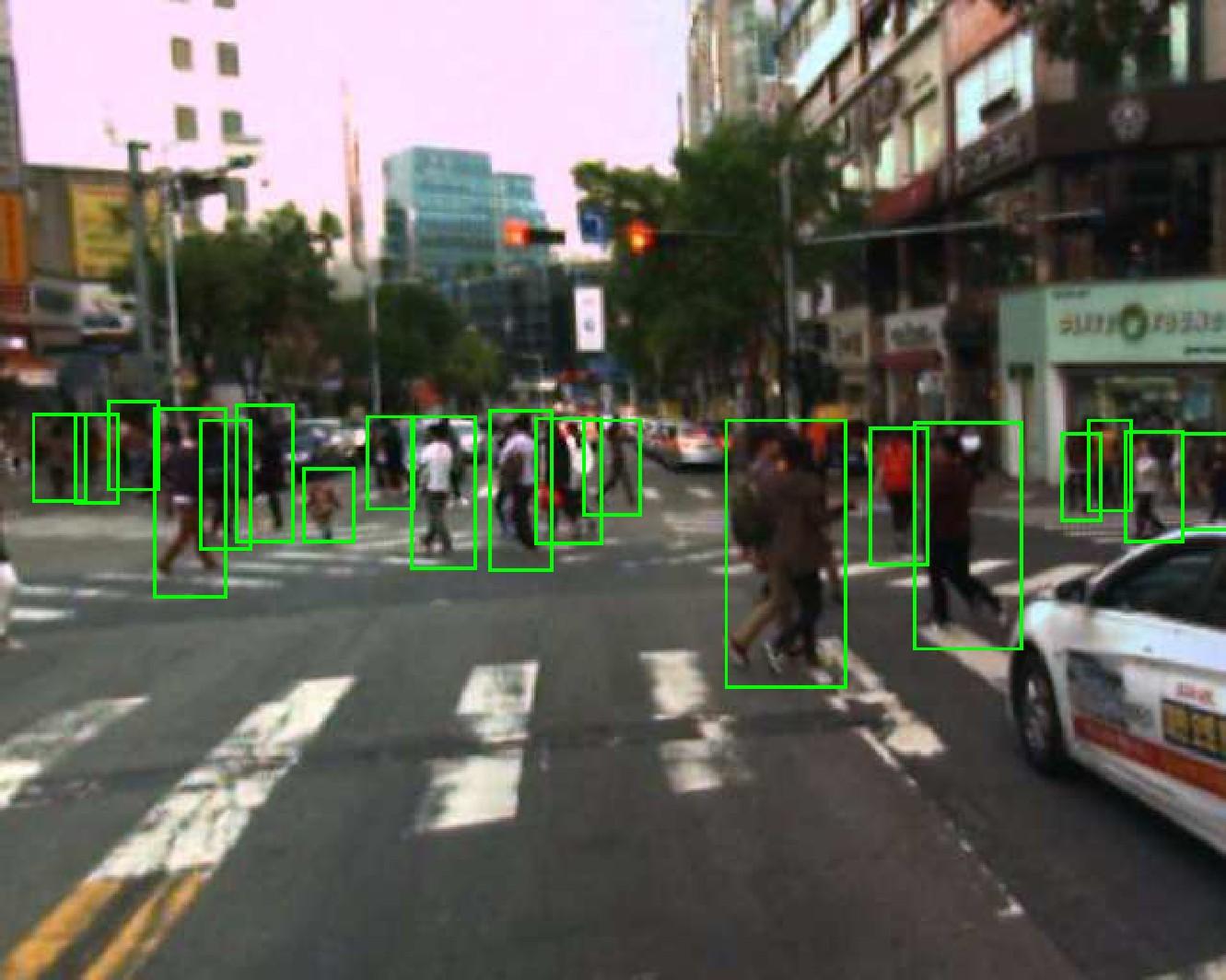}}
		{(a)}
	\end{minipage}
	\begin{minipage}{0.48\linewidth}
		\centering
		{\includegraphics[width=1\linewidth,clip]{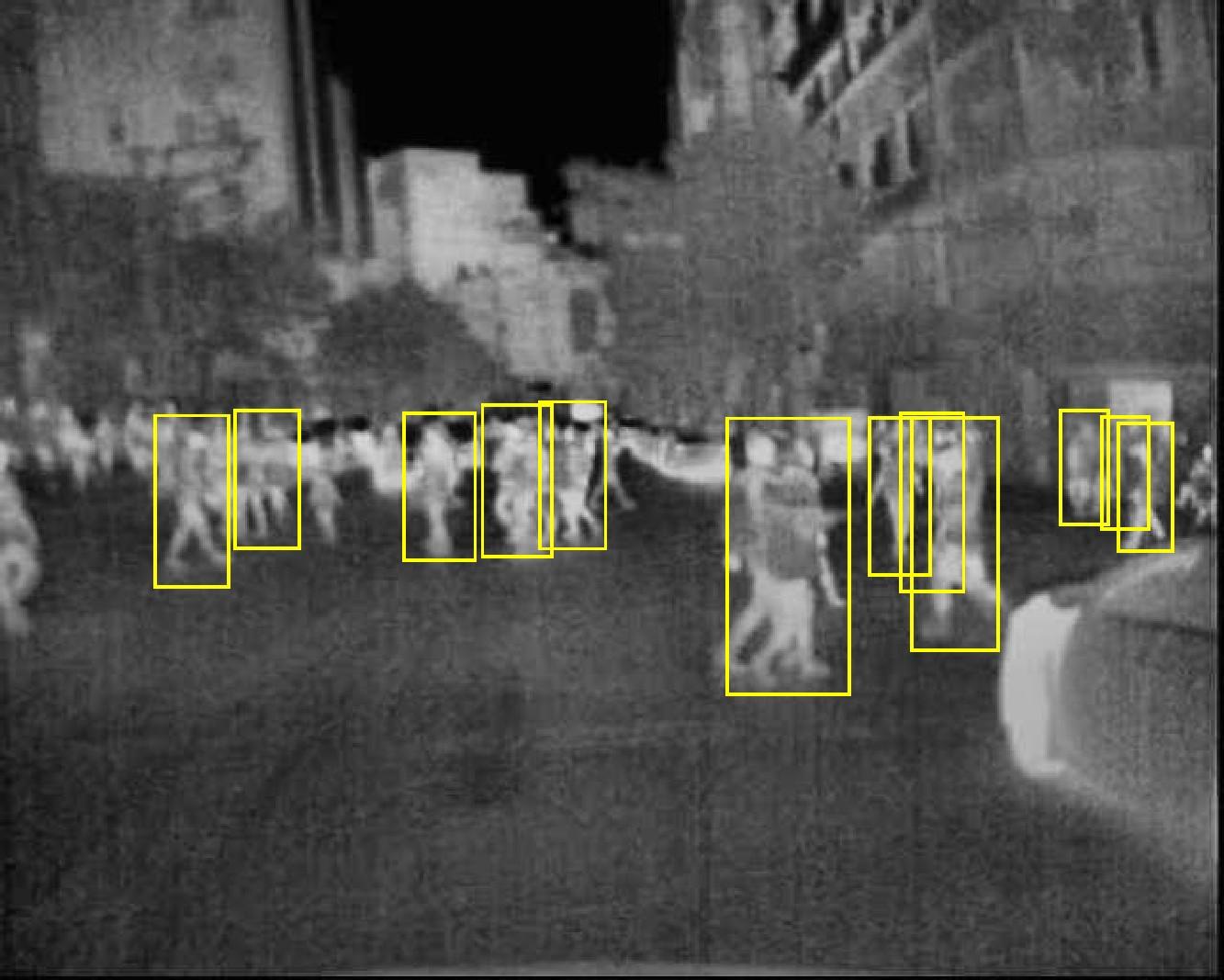}}
		{(b)}
	\end{minipage} \\
	\vspace{2mm}
	\begin{minipage}{0.48\linewidth}
		\centering
		{\includegraphics[width=1\linewidth,clip]{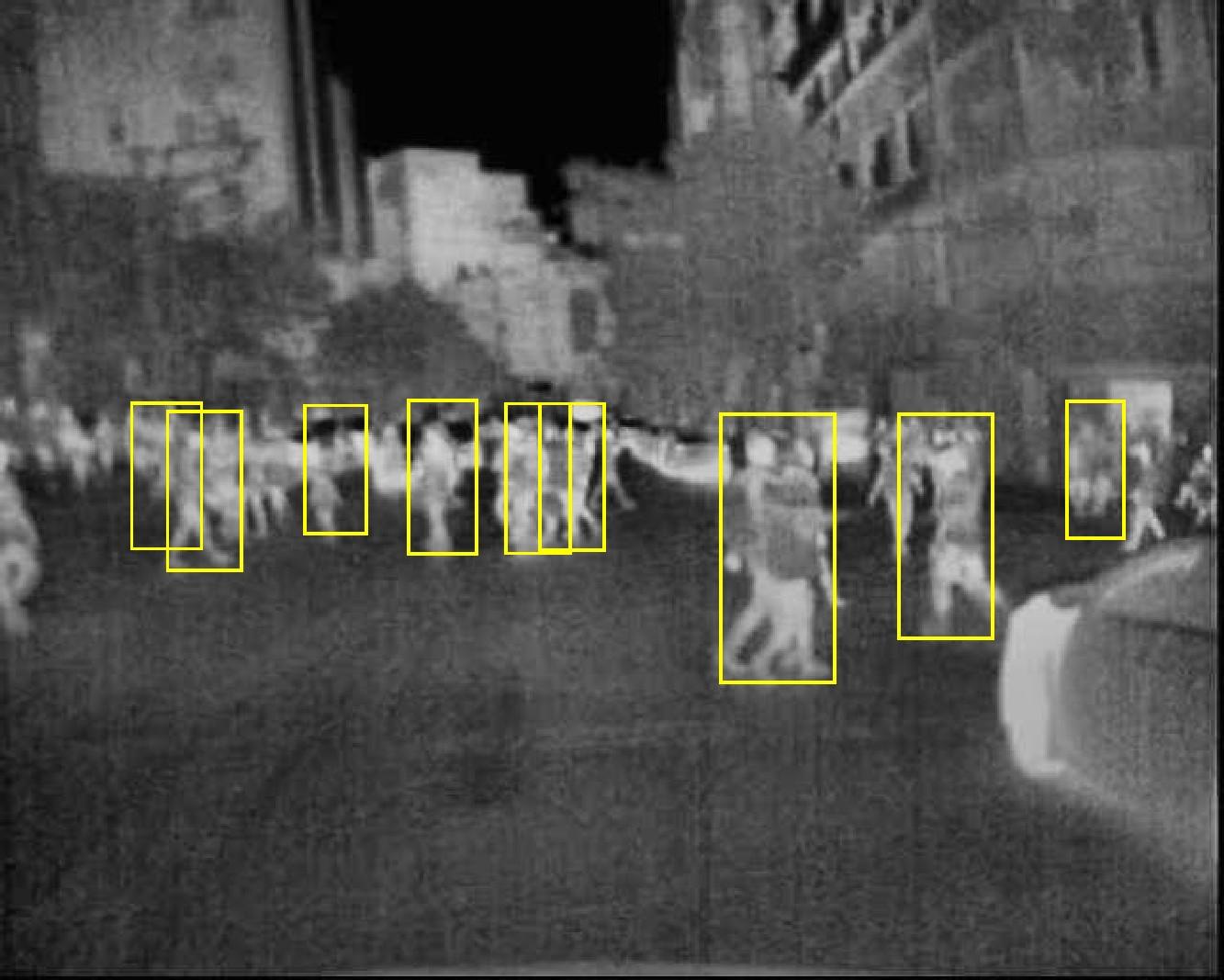}}
		{(c)}
	\end{minipage}
	\begin{minipage}{0.48\linewidth}
		
		\centering
		{\includegraphics[width=1\linewidth,clip]{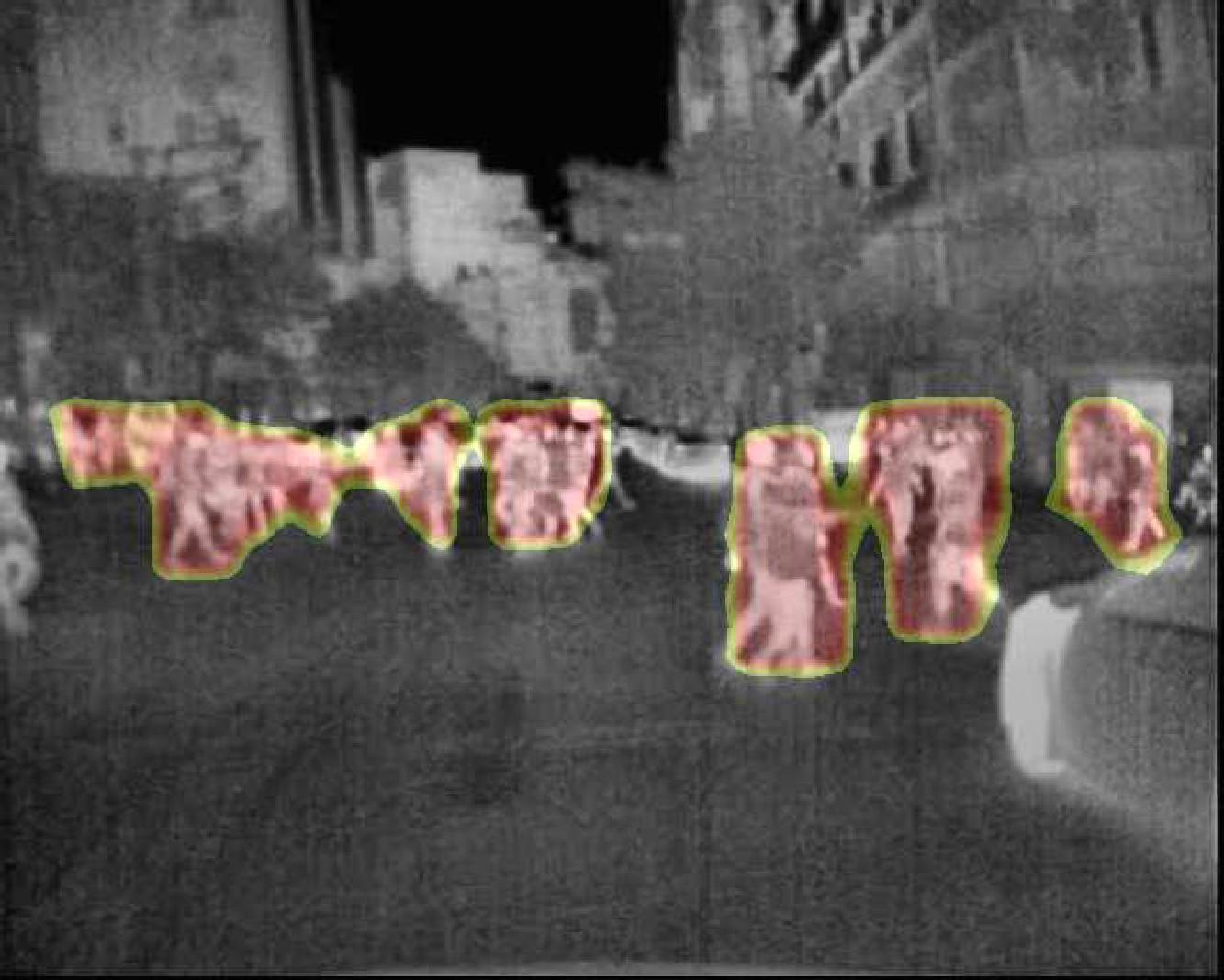}}
		{(d)}
	\end{minipage}
	\caption{(a) Ground truth detection results (displayed using the visible channel); (b) Bounding box detection results using 640$\times$512 images (displayed using the thermal channel); (c) Bounding box detection results using 320$\times$256 images; (d) Detection results of our proposed method using 320$\times$256 images. Note that green bounding boxes show ground truth boxes, yellow bounding boxes show bounding box detections. {A score threshold of 0.5 is used to display the detections.} It is observed that the proposed box-level segmentation supervised learning framework produces more accurate detection results and successfully localizes far-scale human targets even when the input is small-size images. All images are resized to the same resolution for visualization.} 
	\label{fig_1}
\end{figure}

Although significant improvements have been accomplished during recent years, it still remains a challenging task to develop a robust pedestrian detector ready for practical applications. It can be noticed that most existing pedestrian detection methods are based on visible information alone. Their performances are sensitive to changes of the environmental brightness (daytime or nighttime). To overcome the aforementioned limitations, multispectral information (\emph{e.g.} visible and infrared), which can supply complementary information about the targets of interest, are considering to build more robust pedestrian detectors under various illumination conditions. In the past few years, multispectral pedestrian detection solutions are developed by many research works to achieve more accurate and stable pedestrian detection results for around-the-clock application \cite{leykin2007thermal, krotosky2008person, torabi2012iterative, oliveira2015multimodal, hwang2015multispectral, gonzalez2016pedestrian}. 


It is noted that most existing multispectral pedestrian detection approaches are built upon anchor box based detectors such as region proposal networks (RPN) \cite{zhang2016faster} or Faster R-CNN \cite{ren2017faster}, localizing each human target using a bounding box. During the training phase, a large number of anchor boxes are needed to ensure sufficient overlap with most ground truth boxes and will cause severe imbalance between positive and negative anchor boxes and slow down the training process \cite{lin2018focal}.  Moreover, the state-of-the-art pedestrian detection techniques only perform well when the input is large-size images. Their performances will drop significantly when applied to small-size images since it is difficult to make use of anchor boxes to generate positive samples for small-size targets. A simple solution is to increase the size of input images and human targets through image up-scaling, however such practice will adversely decrease the computational efficiency which is critical for real-time autonomous driving applications. 

To overcome the problems mentioned above, we present a novel box-level segmentation supervised learning framework for accurate and real-time multispectral pedestrian detection. Our approach takes pairs of aligned visible and infrared images with easily obtained bounding box annotations as input and computes heat maps to predict the existence of human targets. In Fig.~\ref{fig_1}, we show some comparative detection results of our method with the state-of-the-art anchor box based detector. It is noticed that the proposed box-level segmentation supervised learning framework produces more accurate detection results, successfully locating far-scale human targets even when the input is small-size images. It is also worth mentioning that our proposed method can process more than 30 images per second on a single NVIDIA Geforce Titan X GPU which is sufficient for real-time applications in autonomous vehicles. The contributions of this work are as follows.

Overall, the \textbf{contributions} of this paper are summarized as follows:
\begin{itemize}
	\item[1] Our box-level segmentation supervised framework completely eliminates the complex hyperparameter settings of anchor boxes (e.g., box size, aspect ratio, stride, and intersection-over-union threshold) required in existing anchor box based detectors. To the best of our knowledge, this is the first attempt to train { deep learning based} multispectral pedestrian detectors without using anchor boxes.
	\item[2] We demonstrate that box-level approximate segmentation masks provide better supervision information than anchored boxes to train two-stream deep neural networks for distinguishing pedestrians from the background, particularly for small human targets. As the result, our method is capable of generating accurate detection results even using small-size input images.
	\item[3] Our method achieves significantly higher detection accuracy compared with the state-of-the-art multispectral pedestrian detectors \cite{konig2017fully, Liu2016BMVC, guan2018fusion, guan2018exploiting, li2018multispectral}. Moreover, this efficient framework can process more than 30 images per second on a single NVIDIA Geforce Titan X GPU to facilitate real-time applications in autonomous vehicles.
\end{itemize}

The remainder of our paper is structured as follows. Section~\ref{related} reviews existing research work on multispectral pedestrian detection. The details of our proposed box-level segmentation supervised deep neural networks are presented in Section~\ref{method}. An extensive evaluation of our method and experimental comparison of methods for multispectral pedestrian detection are provided in Section~\ref{experiment}. We conclude our paper in Section~\ref{conclusion}.


\section{RELATED WORKS}
\label{related}

Pedestrian detection facilitates various applications in robotics, automotive safety, surveillance, and autonomous vehicles. A large variety of visible-channel pedestrian detectors have been proposed. Schindler et al. \cite{schindler2010automatic} developed a visual stereo system, which consists of various probabilistic models to fuse evidence from 3D points and 2D images, for accurate detection and tracking of pedestrians in urban traffic scenes. Piotr et al. \cite{Piotr2009ICF} developed the Integrate Channel Features (ICF) detector using feature pyramids and boosted classifiers for visible images. 
{The feature representations of ICF have been further improved through various techniques, including aggregated channel features (ACF) \cite{dollar2014fast}, locally decorrelated channel features (LDCF) \cite{nam2014local}, Checkerboards \cite{Zhang2015Checkerboards} etc. }
Klinger et al. \cite{klinger2017probabilistic} addressed the problems of target occlusion and imprecise visual observation by building up a new predictive model on the basis of Gaussian process regression, and by combining generic object detection with instance-specific classification for refined localization. Object detection based on deep neural networks \cite{girshick2015fast, ren2017faster, He2017ICCV} have achieved state-of-the-art results on various challenging benchmarks, thus they have been adopted for the task of human-target detection. Li et al. \cite{li2015scale} developed a scale-aware fast region-based convolutional neural networks (SAF R-CNN) method which combines a large-size sub-network and a small-size one into a unified architecture using a
scale-aware weighting mechanism to capture unique pedestrian features at different scales. Zhang et al. \cite{zhang2016faster} proposed an effective baseline for pedestrian detection using region proposal networks (RPN) followed by boosted classifiers, which utilizes high-resolution convolutional feature maps generated by RPN for classification. Mao et al. \cite{Mao2017CVPR} proposed a powerful deep neural networks framework by implementing representations of channel features to boost pedestrian detection accuracy without extra inputs in inference. Brazil et al. \cite{Brazil2017ICCV} developed an effective segmentation infusion network to improve pedestrian detection performance through the joint training of target detection and semantic segmentation.


Recently, multispectral pedestrian detection becomes a promising solution to narrow the gap between automatic pedestrian detectors and human observers. Multi-modal sensors (visible and infrared) supply complementary information about the targets of interest thus lead to more robust and accurate detection results. Hwang et al. \cite{hwang2015multispectral} published the first large-scale multispectral pedestrian dataset (KAIST) which contains well-aligned visible and infrared image pairs with dense pedestrian annotations. Wagner et al. \cite{wagner2016multispectral} presented the first application of deep neural networks for multispectral pedestrian detection. Two decision networks, one for early-fusion and the other for late-fusion, were proposed to classify the proposals generated by ACF+T+THOG \cite{hwang2015multispectral} and achieved more accurate detections. Liu et al. \cite{Liu2016BMVC} systematically evaluated the performance of four ConvNet fusion architectures which integrate two-branch ConvNets on different DNNs stages and found the optimal architecture is the Halfway Fusion model that merges two-branch ConvNets on the middle-level convolutional features. K{\"o}nig et al. \cite{konig2017fully} adopt the architecture of RPN+BDT \cite{zhang2016faster} to build Fusion RPN+BDT, which merges the two-branch RPN on the middle-level convolutional features, for multispectral pedestrian detection. Recently, researchers explore illumination information of a scene and proposed illumination-aware weighting mechanism to boost multispectral pedestrian detection performances \cite{guan2018fusion, li2018illumination}. Guan et al. \cite{guan2018exploiting} presented a unified multispectral fusion framework for joint training of semantic segmentation and target detection. More accurate detection results were obtained by infusing the multispectral semantic segmentation masks as supervision for learning human-related features. Li et al. \cite{li2018multispectral} further deployed subsequent multispectral classification network to distinguish pedestrian instances from hard negatives. 


It is noted that most existing multispectral pedestrian detection approaches are built upon anchor box based detectors such as region proposal networks (RPN) \cite{zhang2016faster} or Faster R-CNN \cite{ren2017faster}, using a number of bounding boxes to localize human pedestrians. However, the use of anchor boxes will cause severe imbalance between positive and negative training samples \cite{lin2018focal} and involve complex hyperparameter settings (e.g., box size, aspect ratio, stride, and intersection-over-union threshold) \cite{law2018cornernet}. Our method is very different from the existing anchor box based multispectral pedestrian detectors \cite{konig2017fully, Liu2016BMVC, li2018illumination, guan2018fusion, guan2018exploiting, li2018multispectral} in two major aspects. Firstly, we make use of the ground truth bounding boxes (manually annotated) to generate coarse box-level segmentation masks, which are utilized to replace the anchor bounding boxes for the training of two-stream deep neural networks to learn human-relative characteristic features. Secondly, our method estimates a prediction heat map instead of a number of bounding boxes to localize pedestrians in the surrounding space, which can be easily used to support perceptive autonomous driving applications such as path planning or collision avoidance. It is worth mentioning that a large number of semantic segmentation techniques have been proposed to generate accurate boundary between foreground objects and background regions without using anchor boxes \cite{ha2017mfnet, balloch2018unbiasing, jegou2017one}. However, these methods typically require the supervision of pixel-level accuracy mask annotations which are very time-consuming to obtain. Many researchers attempted to achieve competitive semantic segmentation accuracy by only using the easily obtained bounding box annotations \cite{dai2015boxsup, rajchl2017deepcut}. These methods involve iterative updates to gradually improve the accuracy of segmentation masks, which are slow and not suitable for real-time autonomous driving applications.

\section{Our Approach}

\begin{figure}[!h]
	\centering{\includegraphics[width=0.8\linewidth]{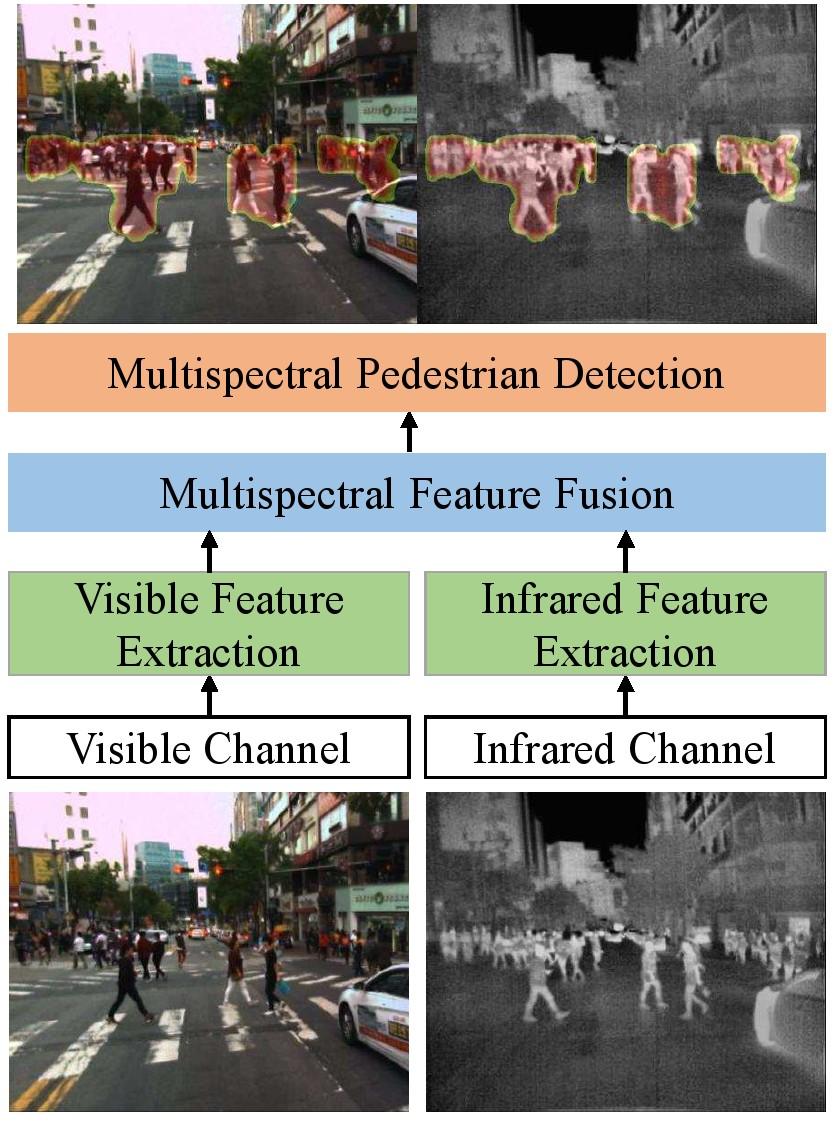}}
	\caption{The workflow of our proposed box-level segmentation supervised deep neural networks for multispectral pedestrian detection. Please note our method generates a prediction heat map { (a score threshold of 0.5 is used to display the detected pedestrian regions)} instead of a number of bounding boxes to localize pedestrians in the scene. Best viewed in color.}
	\label{fig2}
\end{figure}

We propose a novel box-level segmentation supervised framework for multispectral pedestrian detection. Given pairs of well-aligned visible and infrared images, we make use of two-stream deep neural networks to extract semantic features in individual channels. Visible and infrared feature maps are combined through the concatenation operation and then utilized to estimate heat maps to predict the existence of pedestrians as illustrated in Fig.~\ref{fig2}. 
{Note that image regions corresponding to human targets produce high confident scores (larger than 0.5). }

\label{method}

\subsection{Network Architecture}

\begin{figure}[!ht]
	\centering
	\begin{minipage}[t]{0.49\linewidth}
		\centering
		{\includegraphics[width=1\linewidth,clip]{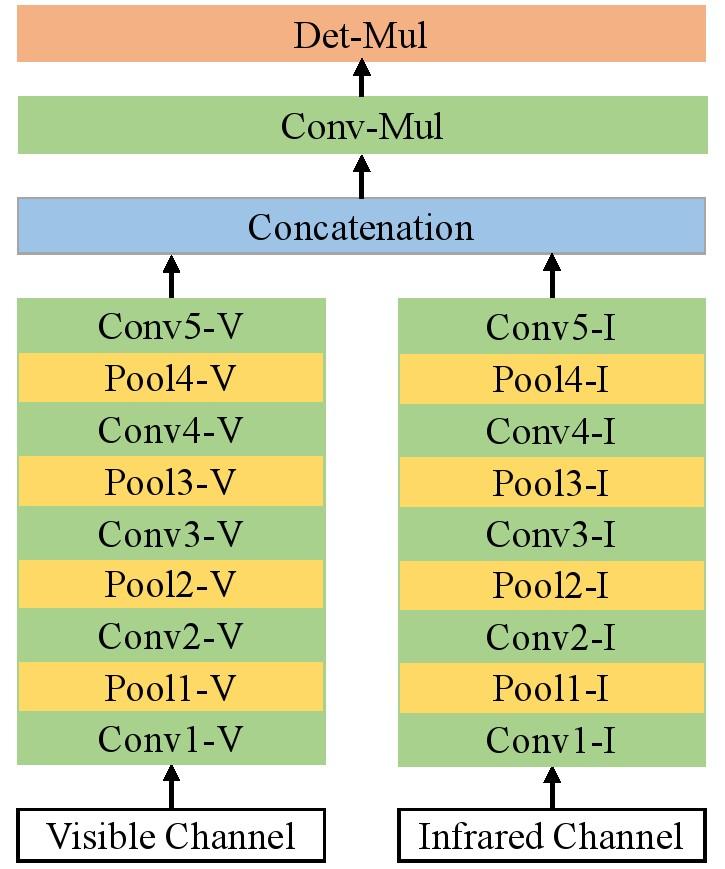}}\\
		{(a) MFFN }
	\end{minipage}
	\begin{minipage}[t]{0.49\linewidth}
		\centering
		{\includegraphics[width=1\linewidth,clip]{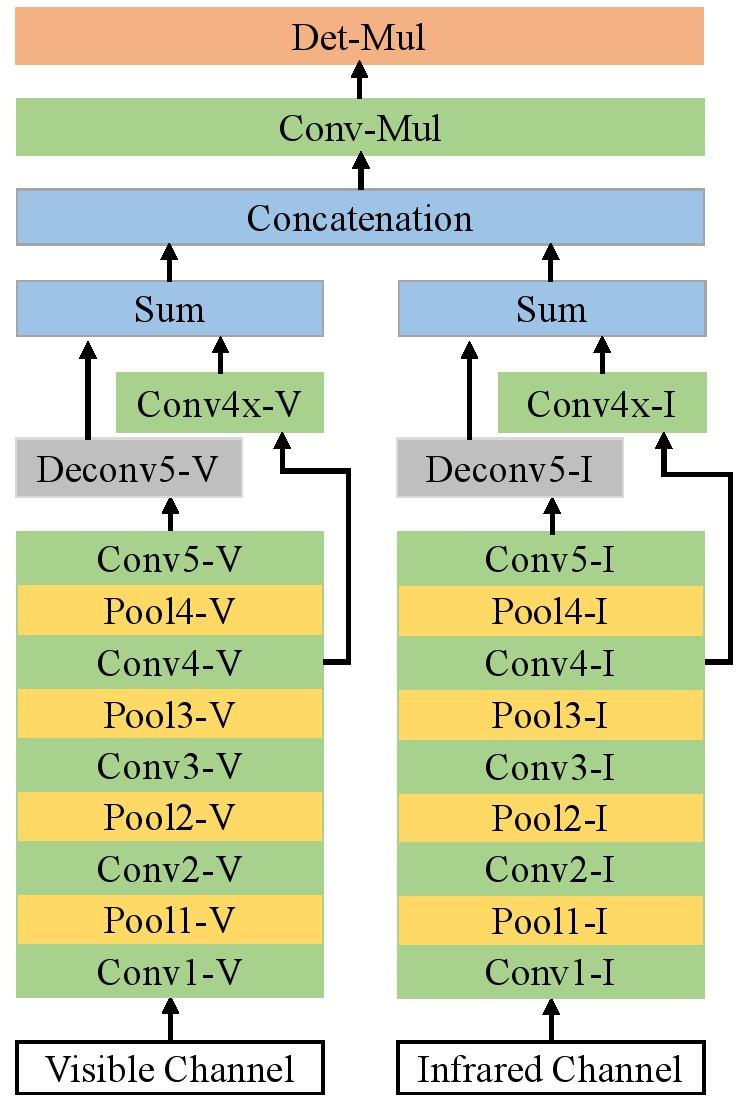}}\\
		{(b) HMFFN}
	\end{minipage}
	\caption{Illustration of (a) MFFN and (b) HMFFN architectures. Note that green boxes represent convolutional layers, yellow boxes represent pooling layers, blue boxes represent fusion layers, gray boxes represent deconvolutional layers, and orange boxes represent soft-max layers. Best viewed in color.}
	\label{fig3}
\end{figure}

Fig.~\ref{fig3} (a) shows a baseline architecture of our proposed multispectral feature fusion network (MFFN) for pedestrian detection.  Given a pair of well-aligned visible and infrared images, we make use of the two-stream deep convolutional neural networks presented by Liu \emph{et al.} \cite{liu2016multispectral} to extract semantic feature maps in individual channels. Note that each feature extraction stream consists of five convolutional layers and pooling ones (Conv1-V to Conv5-V in the visible stream and Conv1-I to Conv5-I in the infrared stream) which adopts the architectures of Conv1-5 from VGG-16 \cite{simonyan2014very}. The two single-channel feature maps are then fused using the concatenation layer followed by a $1\times1$ convolutional layer (Conv-Mul) to learn two-channel multispectral semantic features. We use a softmax layer (Det-Mul) to estimate the heat map to predict the location of pedestrians. 

Inspired by the recent success of top-down architecture with lateral connections for object detection and segmentation \cite{pinheiro2016learning, lin2017feature}, we design another hierarchical multispectral feature fusion network (HMFFN) and its architecture is shown in Fig.~\ref{fig3}(b). It is noted that the HMFFN architecture makes use of skip connections to associate the middle-level feature maps (output of Conv4-V/I layers) with the high-level ones (output of Conv5-V/I layers). The deconvolutional layers (Deconv5-V/I) are deployed to increase the spatial resolution of high-level feature maps by a factor of 2. Then, the upsampled high-level feature maps are merged with the corresponding middle-level ones (which undergoes $1\times1$ convolutional layers Conv4x-V/I to reduce channel dimensions) by element-wise addition. In deep convolutional neural networks, outputs of deeper layers encode high-level semantic information while shallower layers outputs capture rich low-level spatial patterns \cite{lin2017feature, hou2017deeply}. Therefore, the proposed HMFFN architecture, combining feature maps from different levels, is capable of extracting informative multi-scale feature maps to achieve more accurate detection results. The comparative evaluation of MFFN and HMFFN architectures are provided in Sec.~\ref{MFFNvsHMFFN}.

\subsection{Box-level segmentation for Supervised Training}

\begin{figure}[!ht]
	\centering
	\begin{minipage}{0.45\linewidth}
		\centering
		{\includegraphics[width=1\linewidth,clip]{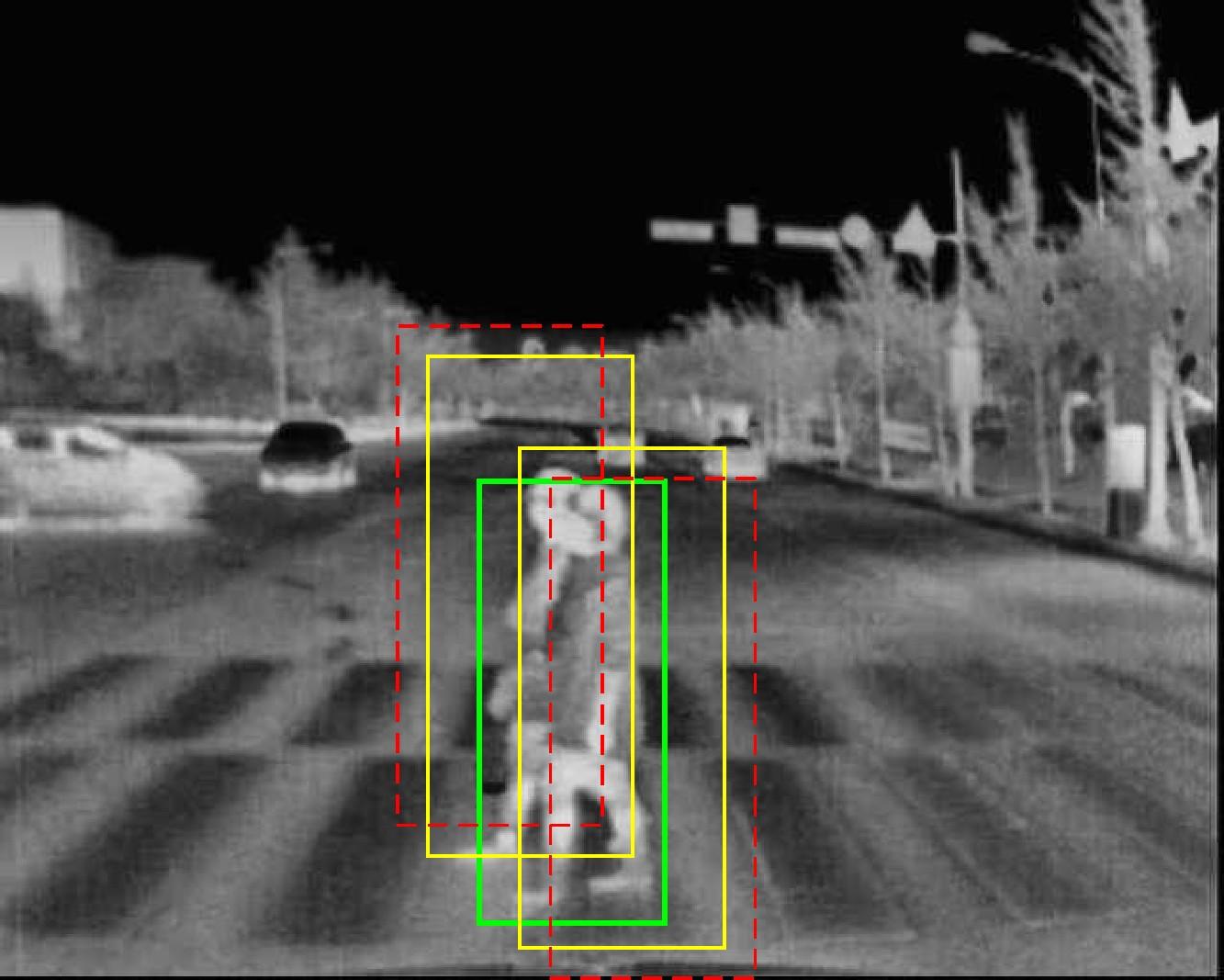}}
		{(a)}
	\end{minipage}
	\begin{minipage}{0.45\linewidth}
		\centering
		{\includegraphics[width=1\linewidth,clip]{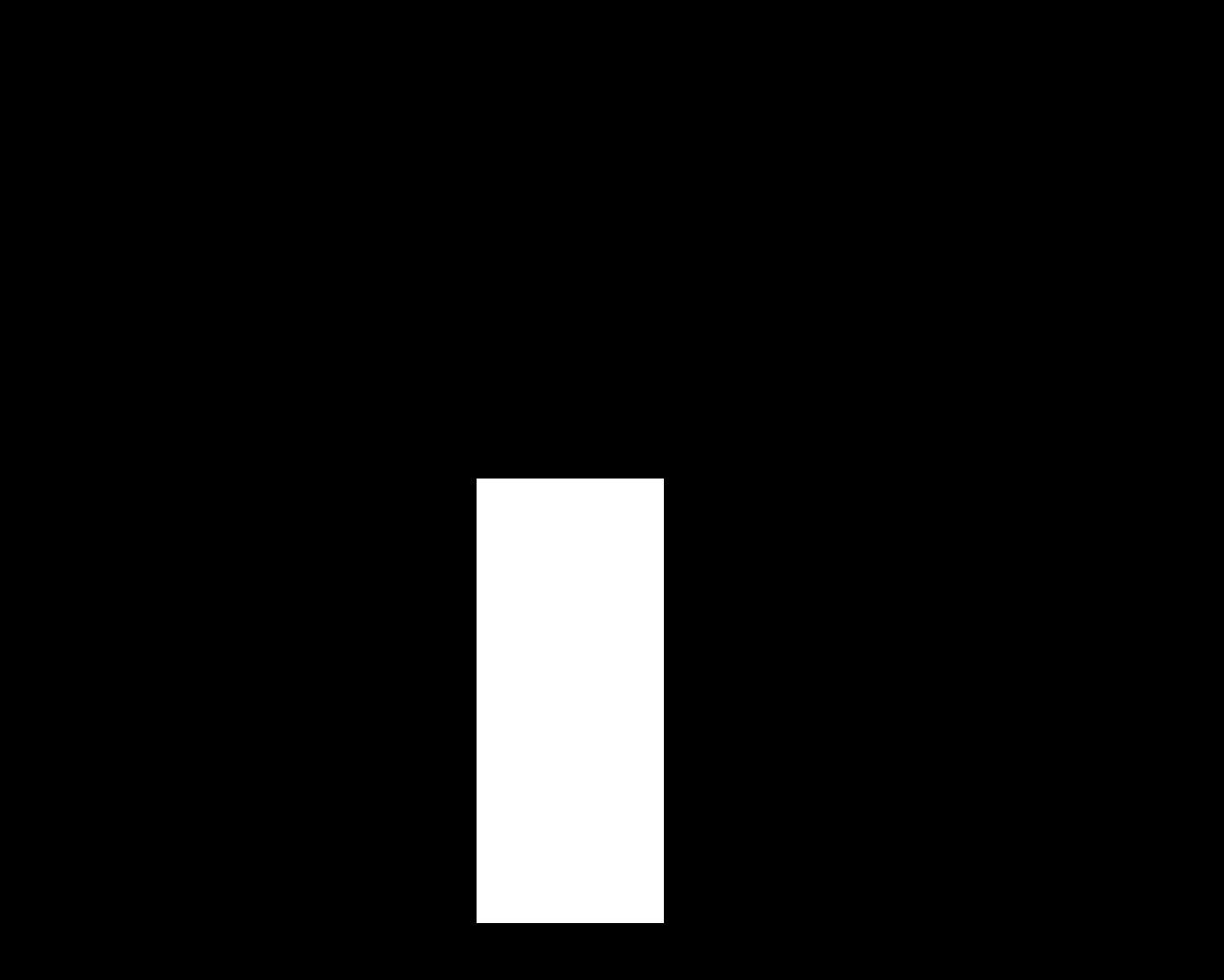}}
		{(b)}
	\end{minipage}
	\caption{Illustration of generating training labels using (a) Anchor boxes; (b) Box-level segmentation masks. The use of anchor boxes involves complex hyperparameter settings (e.g., box size, aspect ratio, stride, and intersection-over-union threshold). In comparison, our proposed method generates an unambiguous box-level segmentation mask for learning human-relative features. Note that green bounding boxes (BBs) represent BB ground truth, yellow BBs represent positive training samples, and red BBs in dashed line represent negative training samples. Best viewed in color.}
	\label{fig4}
\end{figure}

A common step of state-of-the-art anchor box based detectors is generating a large number of anchor boxes with various sizes and aspect ratios as potential detection candidates as illustrated in Fig.~\ref{fig4} (a). However, the use of anchor boxes involves complex hyperparameter settings (e.g., box size, aspect ratio, stride, and intersection-over-union threshold) \cite{law2018cornernet} and causes severe imbalance between positive and negative training samples \cite{lin2018focal}. Moreover, it is difficult to make use of discretely distributed anchor boxes (using a large stride) to generate positive samples for small-size targets. In comparison, our proposed method takes the easily obtained bounding box annotation as input and generates an unambiguous box-level segmentation mask for the training of two-stream deep neural networks to learn human-relative characteristic features as illustrated in Fig.~\ref{fig4} (b). In our implementation, the obtained box-level segmentation masks are down-scaled to match with the size of final multispetral feature maps (outputs of the concatenation layer) through bilinear interpolation. It is worth mentioning that it is a challenging task to obtain pixel-level accurate annotations for visible and infrared image pairs since it is difficult to obtain perfectly aligned and synchronized multispectral data \cite{hwang2015multispectral}. Therefore, we attempt to explore the easily obtained bounding box annotations as an alternative of supervision to train deep convolutional neural networks for multispectral target detection.

Let $\{(X,Y)\}$ denote the training images $X=\{x_{i},i=1,...,M\}$ (M pixels) with box-level approximate segmentation masks $Y=\{y_{i},i=1,...,M\}$, where $y_{i}=1$ denotes the foreground pixel and $y_{i}=0$ is the background pixel. The parameters $\theta$ of multispectral pedestrian detector are updated by minimizing the cross-entropy loss which is defined as
\begin{equation}
\begin{aligned}
\mathcal{L}(\theta) = -\sum_{i\in{Y_{+}}}{\text{log }\text{Pr}(y_{i}=1|X;\theta)} \\ -\sum_{i\in{Y_{-}}}{\text{log }\text{Pr}(y_{i}=0|X;\theta)} ,
\label{eq1}
\end{aligned}
\end{equation}
where $Y_{+}$ and $Y_{-}$ represent the foreground and background pixels respectively, and $\text{Pr}(y_{i}|X;\theta)\in[0,1]$ is the confidence score of the prediction that measures probability of the pixel belong to pedestrian regions. The confidence score is calculated utilizing the softmax function as
\begin{equation}
\text{Pr}(y_{i}=1|X;\theta)=\frac{e^{s_{1}}}{e^{s_{0}}+e^{s_{1}}},
\label{eq2}
\end{equation}
\begin{equation}
\text{Pr}(y_{i}=0|X;\theta)=\frac{e^{s_{0}}}{e^{s_{0}}+e^{s_{1}}},
\label{eq3}
\end{equation}
where $s_{0}$ and $s_{1}$ are the computed values in our two-channel feature maps. The optimal parameters $\theta^{*}$ are obtained by minimizing the loss function $\mathcal{L}(\theta)$ through the gradient descent optimization algorithm as
\begin{equation}
\theta^{*}=\mathop{\arg\min}_{\theta}\mathcal{L}(\theta).
\label{eq4}
\end{equation}
The output of our method is a full-size prediction heat map in which human target regions yields high confident scores {(larger than 0.5)} while the background regions produce low ones. Such perceptive information is useful for many autonomous driving applications such as path planning or collision avoidance. In comparison, it is difficult/impractical to use a number of bounding boxes to identify individual pedestrians in crowded urban scenes. Visual comparisons are provided in Fig.~\ref{fig_1}.

\section{EXPERIMENTS}
\label{experiment}

\begin{table*}[ht]
	\centering
	\caption{ Quantitative performance {(pixel-level AP \cite{salton1986introduction})} of MFFN and HMFFN for different sizes of input images ($640 \times 512$, $480 \times 384$, and $320 \times 256$). }
	\fontsize{8pt}{8pt}\selectfont
	\begin{tabular}{ccccccccccc}
		\hline
		Model &\tabincell{c}{Reasonable\\all} &\tabincell{c}{Reasonable\\day} &\tabincell{c}{Reasonable\\night} &\tabincell{c}{Near\\scale} &\tabincell{c}{Medium\\scale}  &\tabincell{c}{Far\\scale}  &\tabincell{c}{No\\occlusion} &\tabincell{c}{Partial\\occlusion} &\tabincell{c}{Heavy\\occlusion} &\tabincell{c}{Inference\\speed (fps)}\\
		\hline
		MFFN-640 &0.844 &0.849 &0.836 &0.812 &0.736 &0.163 &0.816 &0.373 &0.169 &12.4 \\
		HMFFN-640 &0.854 &0.865 &0.836 &0.797 &0.785 &0.166 &0.832 &0.391 &0.171 &10.8 \\
		MFFN-480 &0.825	&0.837 &0.812	&0.799 &0.705	&0.100 &0.790	&0.328 &0.152 &20.3 \\
		HMFFN-480 &0.843 &0.866 &0.805 &0.796 &0.764 &0.148 &0.818 &0.373 &0.152 &18.5 \\
		MFFN-320 &0.748	&0.757 &0.740	&0.756 &0.546	&0.043 &0.697	&0.243 &0.110 &40.0 \\
		HMFFN-320 &0.817 &0.825 &0.808 &0.779 &0.696 &0.111 &0.779 &0.345 &0.140 &38.3 \\
		\hline
	\end{tabular}
	\label{tab1}
\end{table*}

\begin{figure*}[ht]
	
	\fontsize{8pt}{8pt}\selectfont
	\begin{minipage}{0.105\linewidth}
		\centering {Near scale\\no occlusion}
		\centering {\includegraphics[width=1\linewidth,clip]{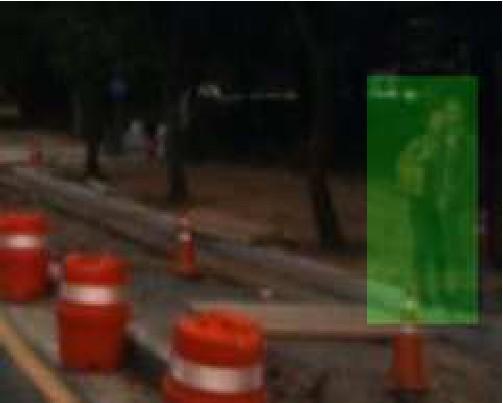}}
	\end{minipage}
	\begin{minipage}{0.105\linewidth}
		\centering {Near scale\\partial occlusion}
		\centering {\includegraphics[width=1\linewidth,clip]{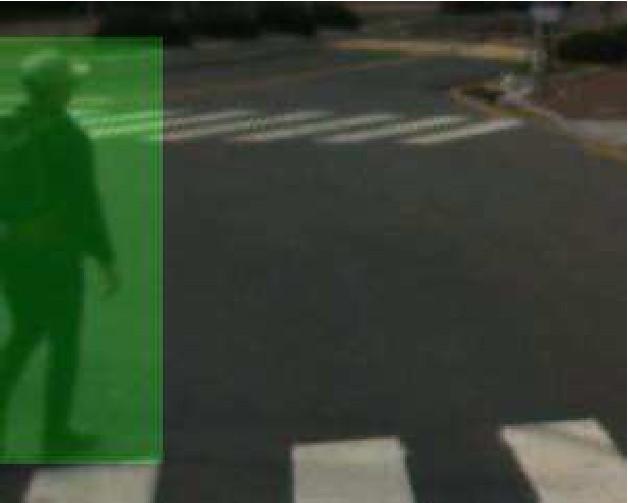}}
	\end{minipage}
	\begin{minipage}{0.105\linewidth}
		\centering {Near scale \\heavy occlusion}
		\centering {\includegraphics[width=1\linewidth,clip]{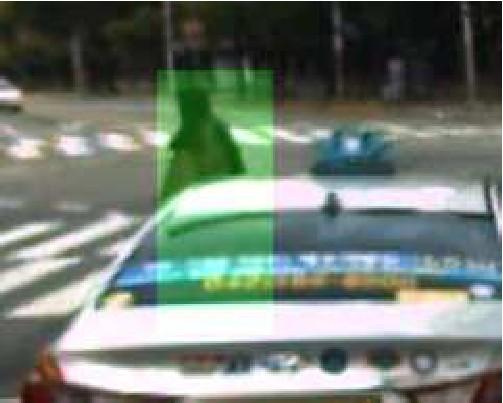}}
	\end{minipage}
	\begin{minipage}{0.105\linewidth}
		\centering {Medium scale\\no occlusion}
		\centering {\includegraphics[width=1\linewidth,clip]{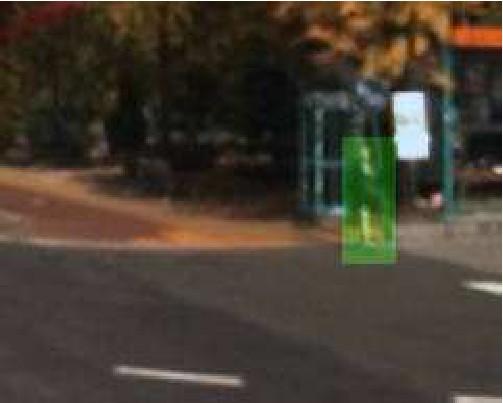}}
	\end{minipage}
	\begin{minipage}{0.105\linewidth}
		\centering {Medium scale\\partial occlusion}
		\centering {\includegraphics[width=1\linewidth,clip]{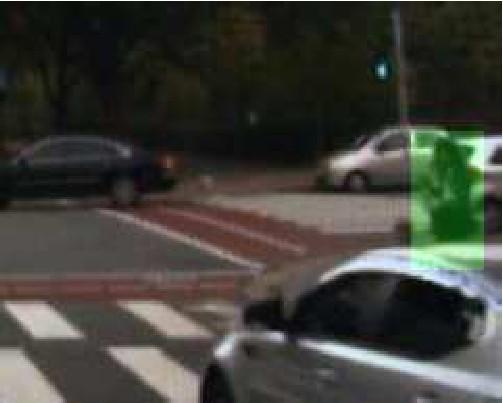}}
	\end{minipage}
	\begin{minipage}{0.105\linewidth}
		\centering {Medium scale\\heavy occlusion}
		\centering {\includegraphics[width=1\linewidth,clip]{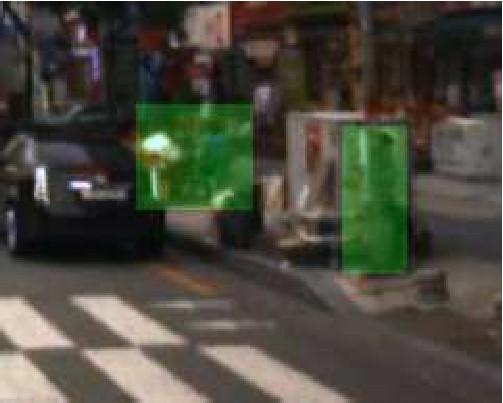}}
	\end{minipage}
	\centering
	\begin{minipage}{0.105\linewidth}
		\centering {Far scale\\no occlusion}
		\centering {\includegraphics[width=1\linewidth,clip]{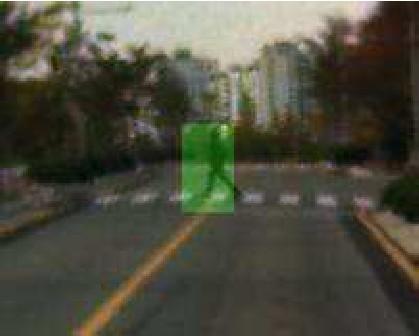}}
	\end{minipage}
	\begin{minipage}{0.105\linewidth}
		\centering {Far scale\\partial occlusion}
		\centering {\includegraphics[width=1\linewidth,clip]{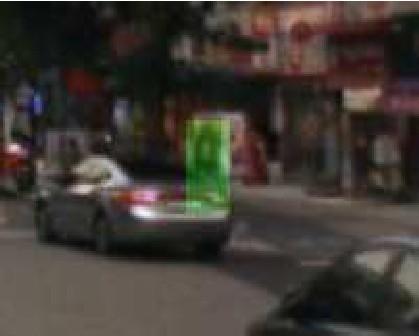}}
	\end{minipage}
	\begin{minipage}{0.105\linewidth}
		\centering {Far scale\\heavy occlusion}
		\centering {\includegraphics[width=1\linewidth,clip]{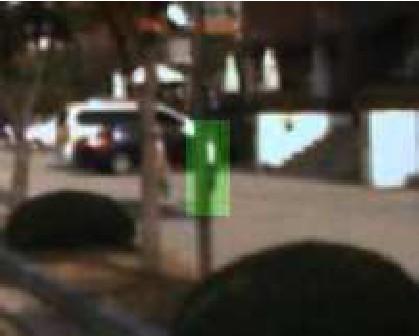}}
	\end{minipage}
	\vspace{1mm}\\
		
	\begin{minipage}{0.105\linewidth}
		\centering {\includegraphics[width=1\linewidth,clip]{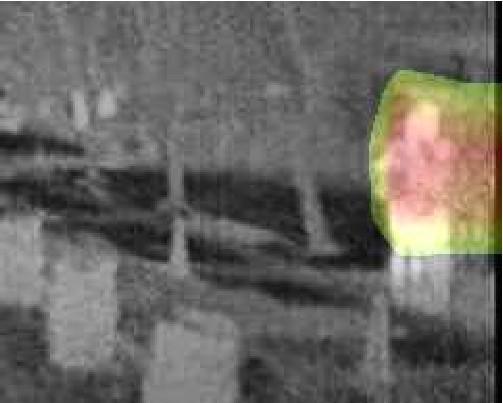}}
	\end{minipage}
	\begin{minipage}{0.105\linewidth}
		\centering {\includegraphics[width=1\linewidth,clip]{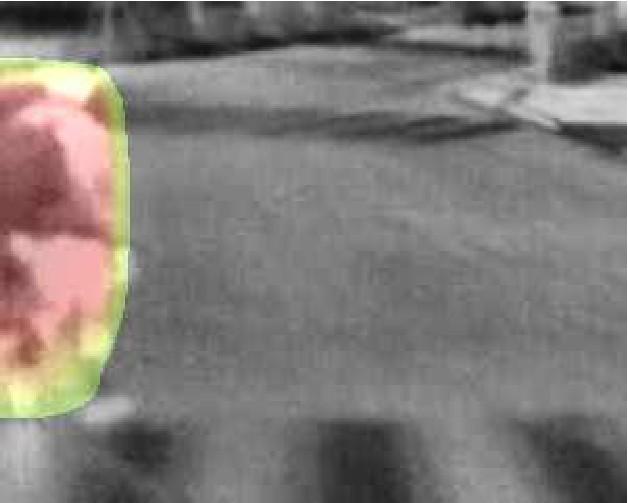}}
	\end{minipage}
	\begin{minipage}{0.105\linewidth}
		\centering {\includegraphics[width=1\linewidth,clip]{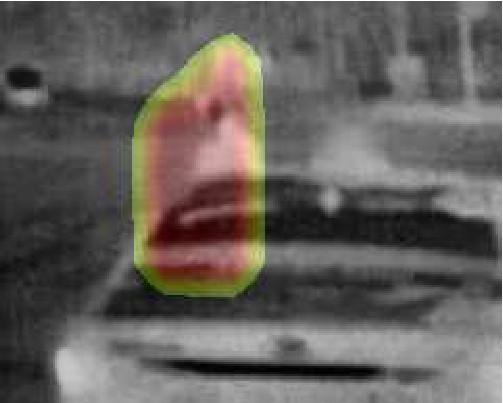}}
	\end{minipage}
	\begin{minipage}{0.105\linewidth}
		\centering {\includegraphics[width=1\linewidth,clip]{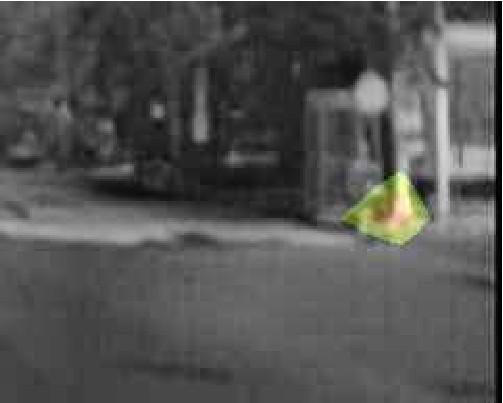}}
	\end{minipage}
	\begin{minipage}{0.105\linewidth}
		\centering {\includegraphics[width=1\linewidth,clip]{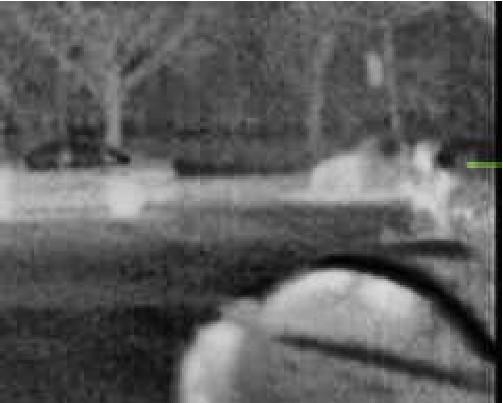}}
	\end{minipage}
	\begin{minipage}{0.105\linewidth}
		\centering {\includegraphics[width=1\linewidth,clip]{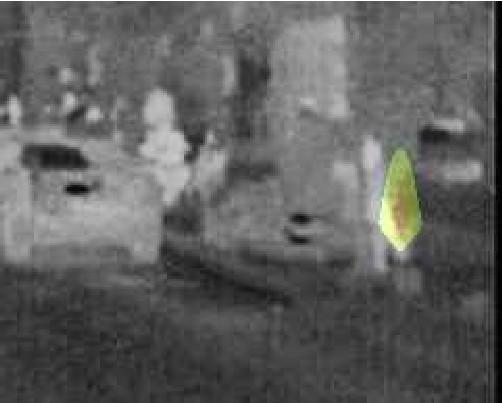}}
	\end{minipage}
	\centering
	\begin{minipage}{0.105\linewidth}
		\centering {\includegraphics[width=1\linewidth,clip]{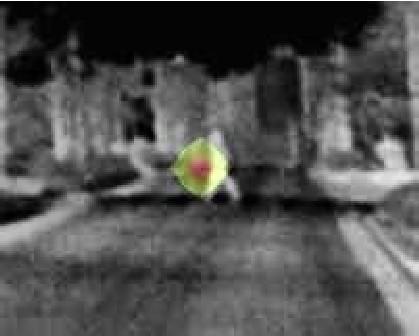}}
	\end{minipage}
	\begin{minipage}{0.105\linewidth}
		\centering {\includegraphics[width=1\linewidth,clip]{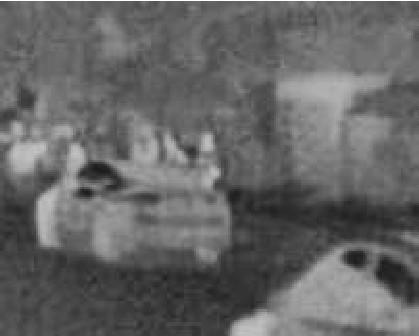}}
	\end{minipage}
	\begin{minipage}{0.105\linewidth}
		\centering {\includegraphics[width=1\linewidth,clip]{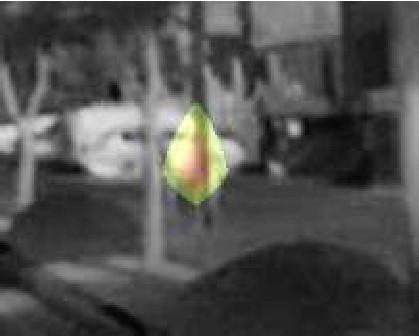}}
	\end{minipage}
	\vspace{1mm}\\

	\begin{minipage}{0.105\linewidth}
		\centering {\includegraphics[width=1\linewidth,clip]{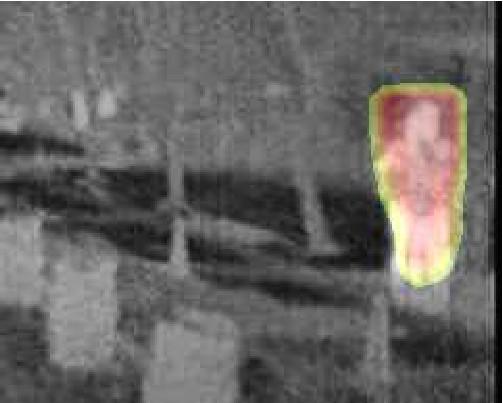}}
	\end{minipage}
	\begin{minipage}{0.105\linewidth}
		\centering {\includegraphics[width=1\linewidth,clip]{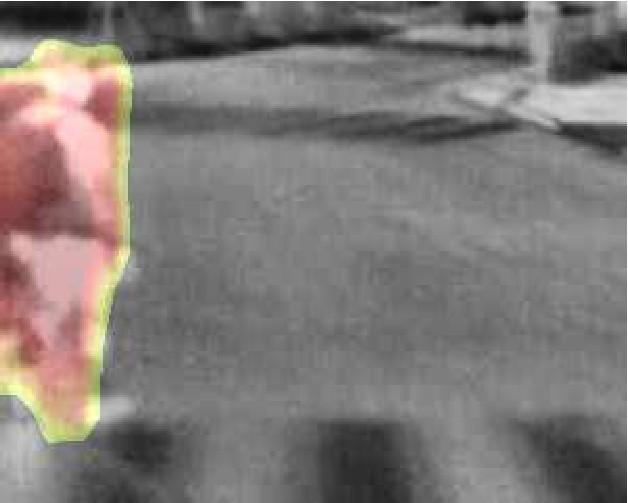}}
	\end{minipage}
	\begin{minipage}{0.105\linewidth}
		\centering {\includegraphics[width=1\linewidth,clip]{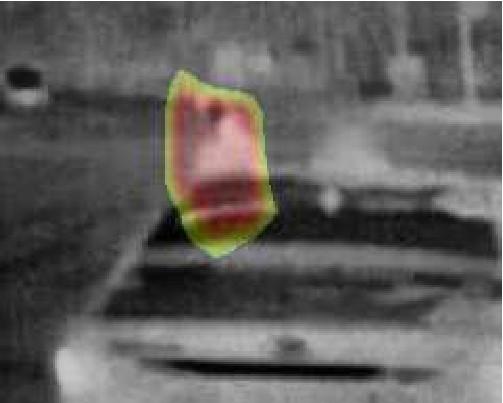}}
	\end{minipage}
	\begin{minipage}{0.105\linewidth}
		\centering {\includegraphics[width=1\linewidth,clip]{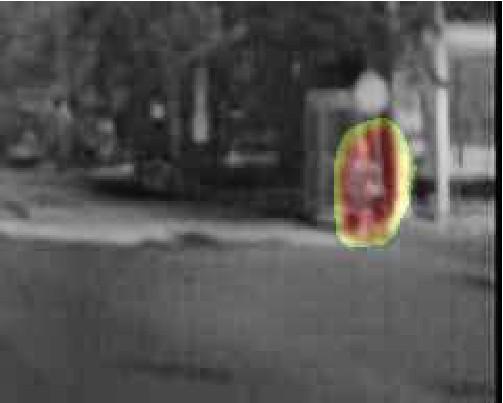}}
	\end{minipage}
	\begin{minipage}{0.105\linewidth}
		\centering {\includegraphics[width=1\linewidth,clip]{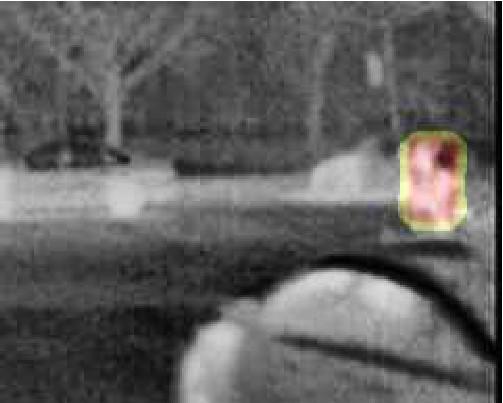}}
	\end{minipage}
	\begin{minipage}{0.105\linewidth}
		\centering {\includegraphics[width=1\linewidth,clip]{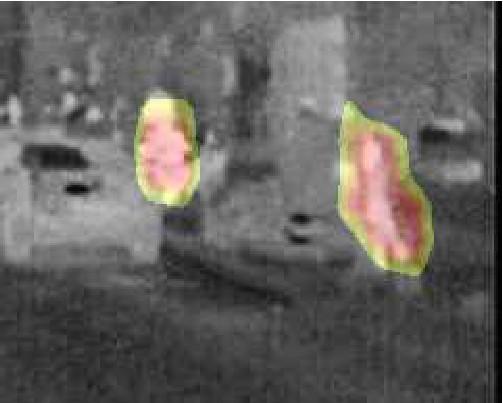}}
	\end{minipage}
	\centering
	\begin{minipage}{0.105\linewidth}
		\centering {\includegraphics[width=1\linewidth,clip]{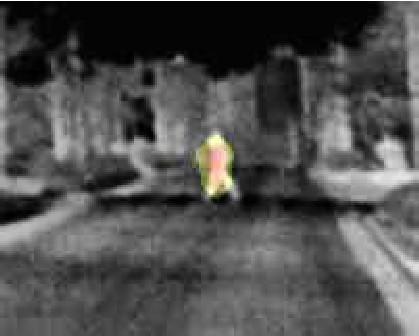}}
	\end{minipage}
	\begin{minipage}{0.105\linewidth}
		\centering {\includegraphics[width=1\linewidth,clip]{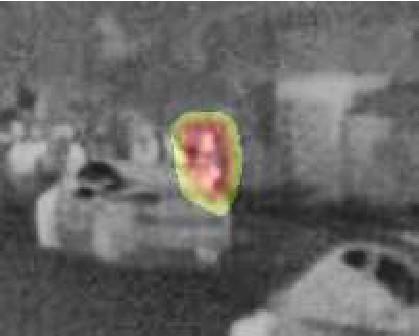}}
	\end{minipage}
	\begin{minipage}{0.105\linewidth}
		\centering {\includegraphics[width=1\linewidth,clip]{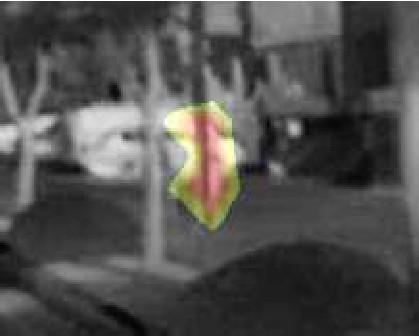}}
	\end{minipage}
	\vspace{1mm}\\
	
	{\bf (a) Daytime}
	\vspace{1mm}\\
	
		\begin{minipage}{0.105\linewidth}
		\centering {\includegraphics[width=1\linewidth,clip]{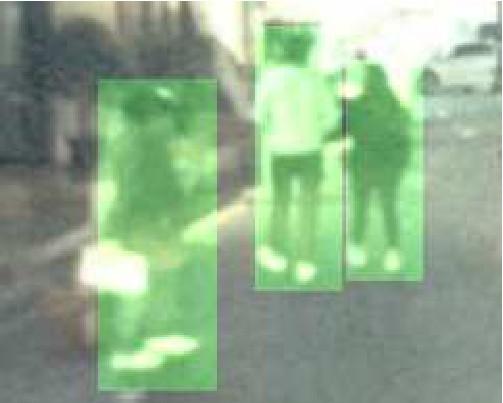}}
	\end{minipage}
	\begin{minipage}{0.105\linewidth}
		\centering {\includegraphics[width=1\linewidth,clip]{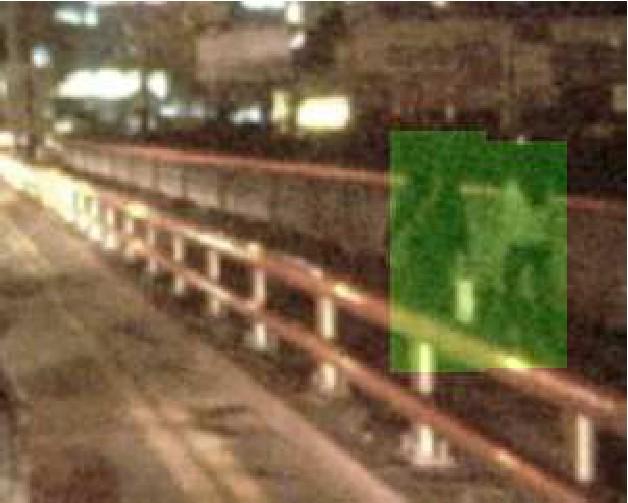}}
	\end{minipage}
	\begin{minipage}{0.105\linewidth}
		\centering {\includegraphics[width=1\linewidth,clip]{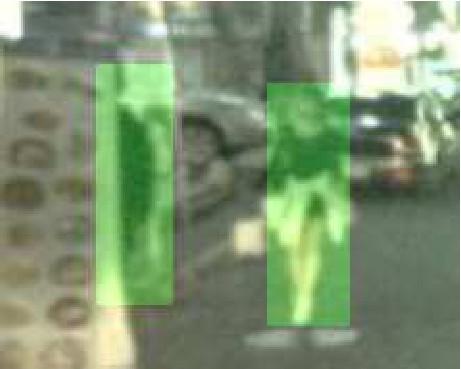}}
	\end{minipage}
	\begin{minipage}{0.105\linewidth}
		\centering {\includegraphics[width=1\linewidth,clip]{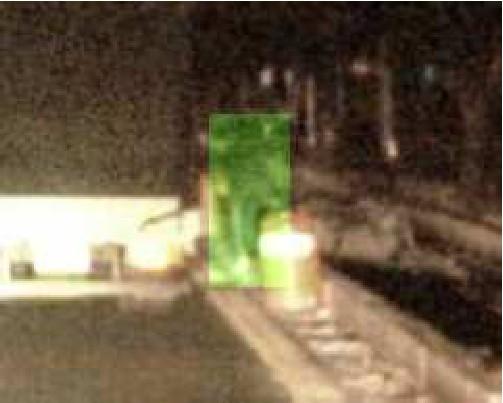}}
	\end{minipage}
	\begin{minipage}{0.105\linewidth}
		\centering {\includegraphics[width=1\linewidth,clip]{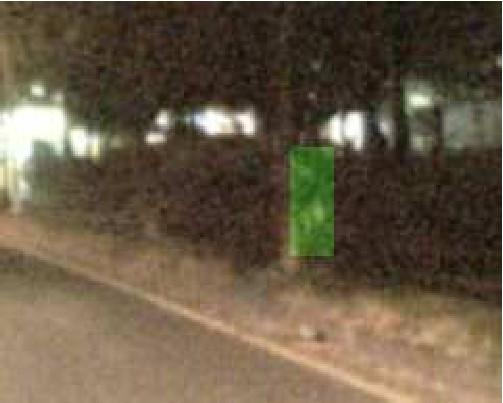}}
	\end{minipage}
	\begin{minipage}{0.105\linewidth}
		\centering {\includegraphics[width=1\linewidth,clip]{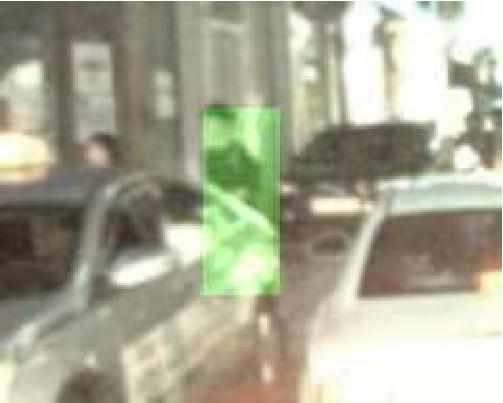}}
	\end{minipage}
	\centering
	\begin{minipage}{0.105\linewidth}
		\centering {\includegraphics[width=1\linewidth,clip]{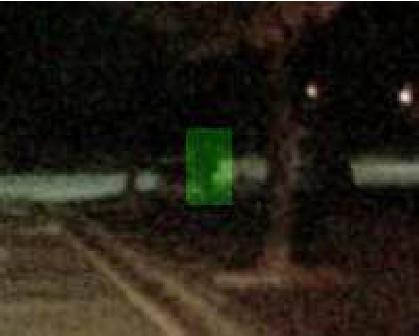}}
	\end{minipage}
	\begin{minipage}{0.105\linewidth}
		\centering {\includegraphics[width=1\linewidth,clip]{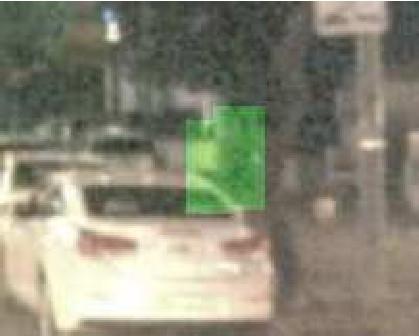}}
	\end{minipage}
	\begin{minipage}{0.105\linewidth}
		\centering {\includegraphics[width=1\linewidth,clip]{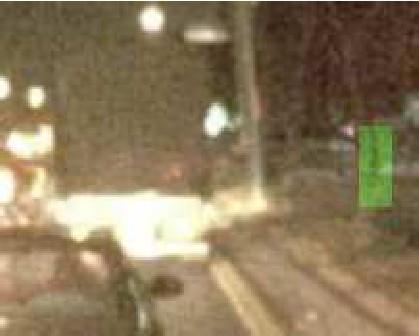}}
	\end{minipage}
	\vspace{1mm}\\
	
		\begin{minipage}{0.105\linewidth}
		\centering {\includegraphics[width=1\linewidth,clip]{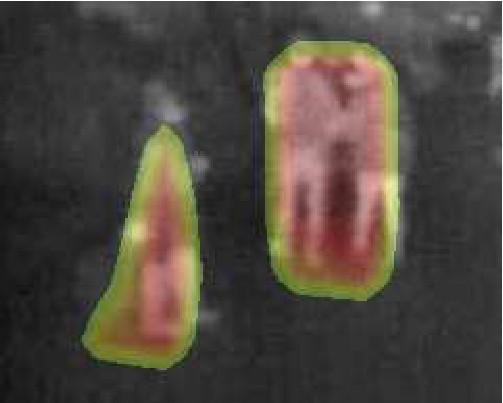}}
	\end{minipage}
	\begin{minipage}{0.105\linewidth}
		\centering {\includegraphics[width=1\linewidth,clip]{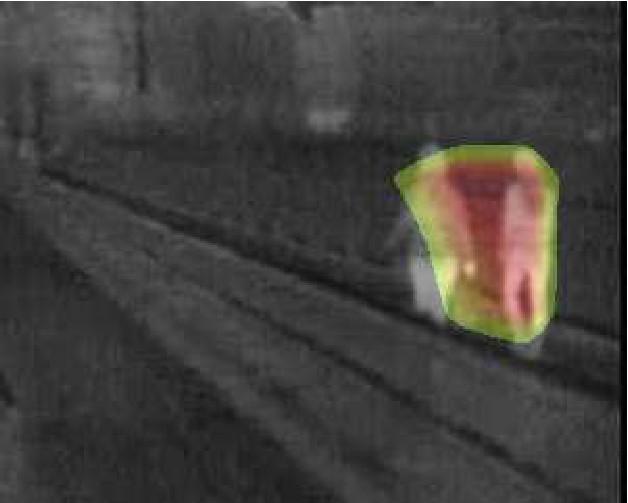}}
	\end{minipage}
	\begin{minipage}{0.105\linewidth}
		\centering {\includegraphics[width=1\linewidth,clip]{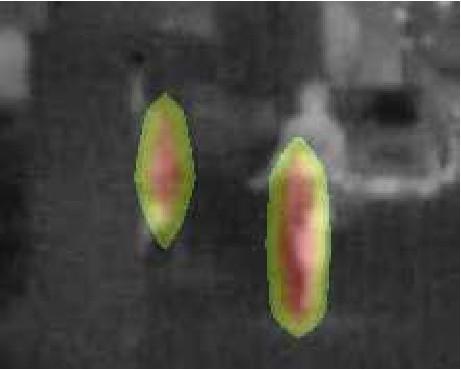}}
	\end{minipage}
	\begin{minipage}{0.105\linewidth}
		\centering {\includegraphics[width=1\linewidth,clip]{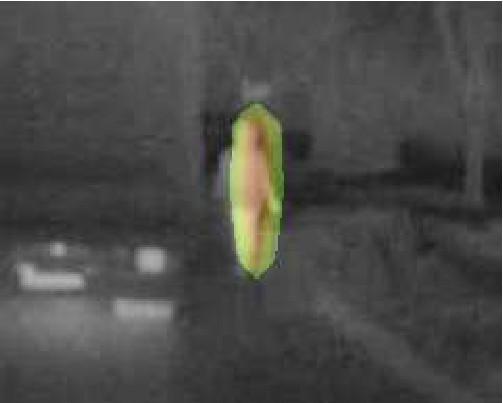}}
	\end{minipage}
	\begin{minipage}{0.105\linewidth}
		\centering {\includegraphics[width=1\linewidth,clip]{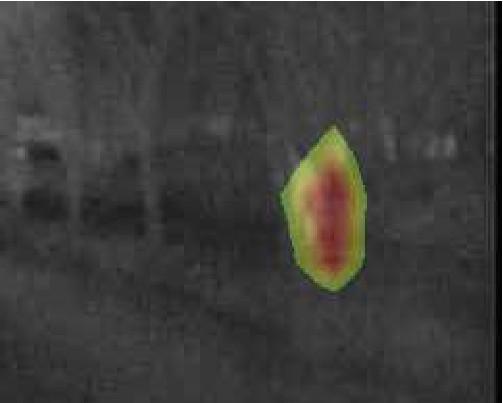}}
	\end{minipage}
	\begin{minipage}{0.105\linewidth}
		\centering {\includegraphics[width=1\linewidth,clip]{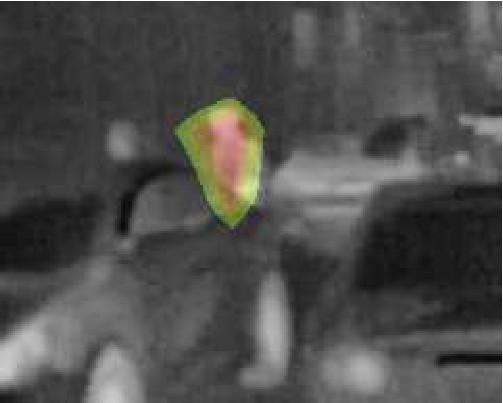}}
	\end{minipage}
	\centering
	\begin{minipage}{0.105\linewidth}
		\centering {\includegraphics[width=1\linewidth,clip]{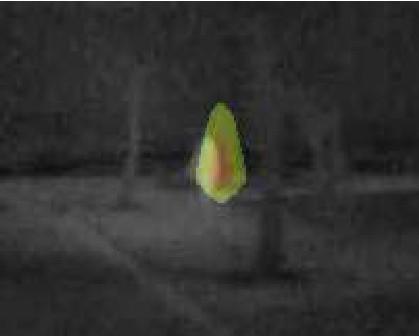}}
	\end{minipage}
	\begin{minipage}{0.105\linewidth}
		\centering {\includegraphics[width=1\linewidth,clip]{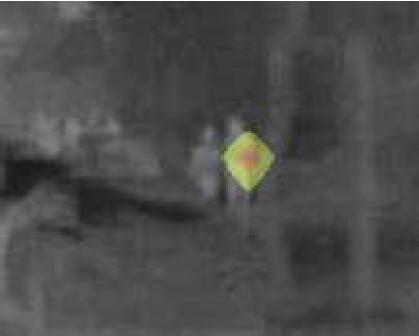}}
	\end{minipage}
	\begin{minipage}{0.105\linewidth}
		\centering {\includegraphics[width=1\linewidth,clip]{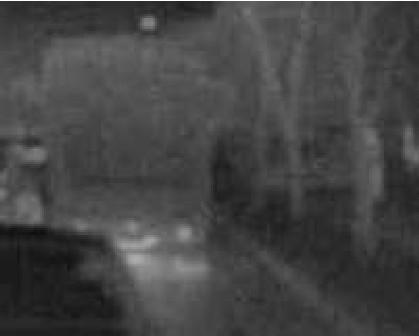}}
	\end{minipage}
	\vspace{1mm}\\
	
		\begin{minipage}{0.105\linewidth}
		\centering {\includegraphics[width=1\linewidth,clip]{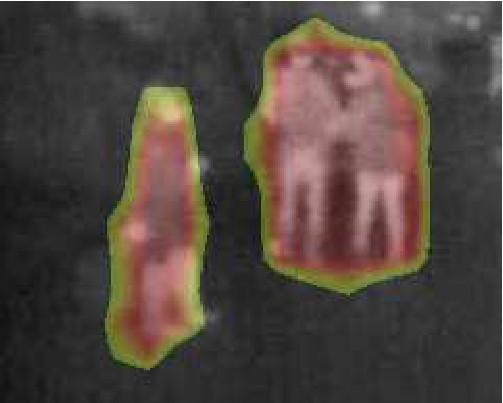}}
	\end{minipage}
	\begin{minipage}{0.105\linewidth}
		\centering {\includegraphics[width=1\linewidth,clip]{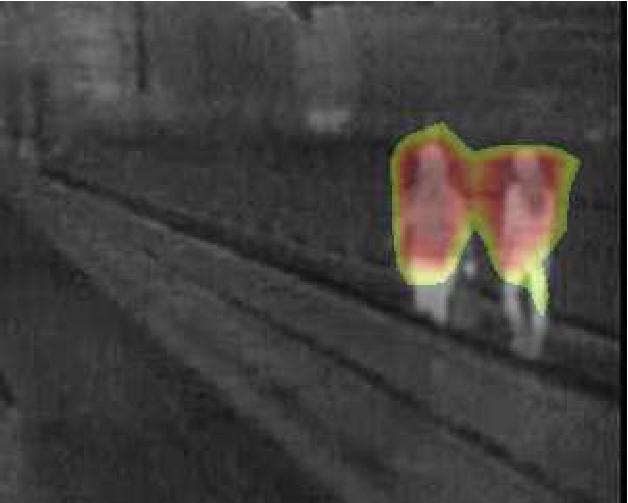}}
	\end{minipage}
	\begin{minipage}{0.105\linewidth}
		\centering {\includegraphics[width=1\linewidth,clip]{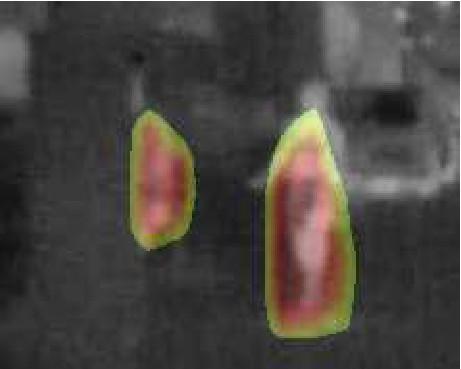}}
	\end{minipage}
	\begin{minipage}{0.105\linewidth}
		\centering {\includegraphics[width=1\linewidth,clip]{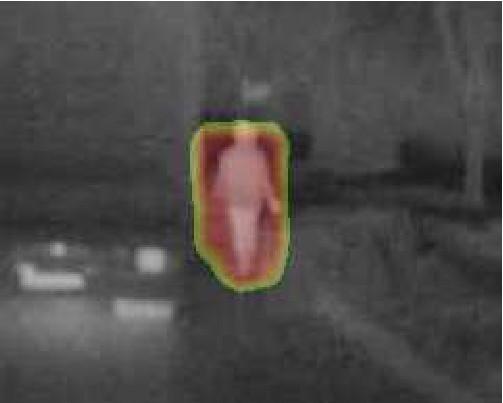}}
	\end{minipage}
	\begin{minipage}{0.105\linewidth}
		\centering {\includegraphics[width=1\linewidth,clip]{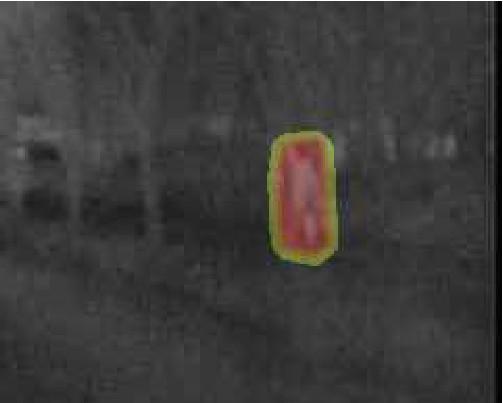}}
	\end{minipage}
	\begin{minipage}{0.105\linewidth}
		\centering {\includegraphics[width=1\linewidth,clip]{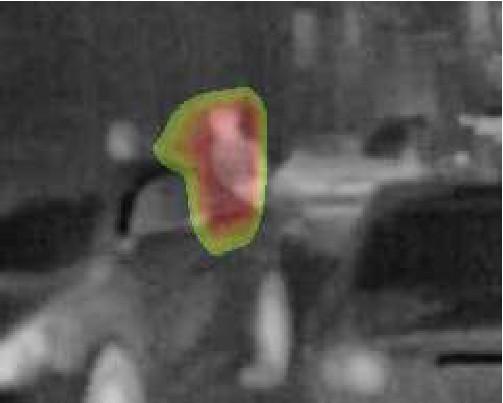}}
	\end{minipage}
	\centering
	\begin{minipage}{0.105\linewidth}
		\centering {\includegraphics[width=1\linewidth,clip]{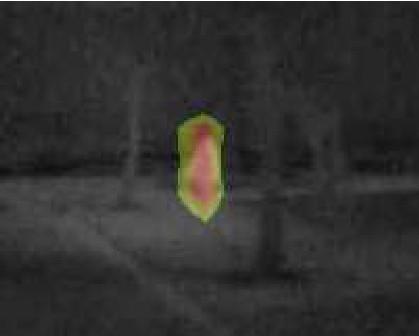}}
	\end{minipage}
	\begin{minipage}{0.105\linewidth}
		\centering {\includegraphics[width=1\linewidth,clip]{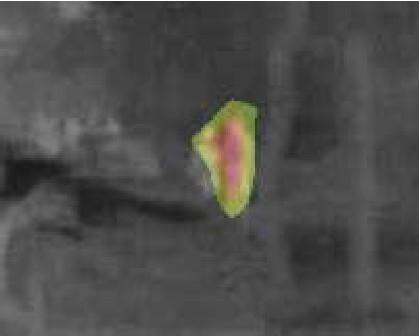}}
	\end{minipage}
	\begin{minipage}{0.105\linewidth}
		\centering {\includegraphics[width=1\linewidth,clip]{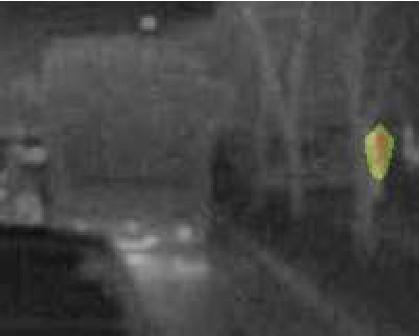}}
	\end{minipage}
	\vspace{1mm}\\
	
	{\bf (b) Nighttime}\\
	
	\centering
	\caption{Qualitative comparison of multispectral pedestrian detection results of MFFN-320 and HMFFN-320 in the KAIST testing images captured in (a) daytime and (b) nighttime scenes. First row shows the ground truth (displaying using the visible channel) and the others show detection results of MFFN-320 and HMFFN-320 respectively (displaying using the infrared channel). Note that the green regions represent ground-truth annotation masks which are generated based on manually labeled bounding boxes, and the detected pedestrian targets are visualized using the heat map representation with a 0.5 threshold. Best viewed in color.}
	\label{fig7}
\end{figure*}

\begin{table*}[ht]
	\centering
	\caption{ Quantitative performance {(pixel-level AP \cite{salton1986introduction})} of our proposed box-based segmentation supervised detectors (HMFFN) with the  anchor box based detectors (RPN-HMFFN) for different sizes of input images ($640 \times 512$, $480 \times 384$, and $320 \times 256$). }
	\fontsize{8pt}{8pt}\selectfont
	\begin{tabular}{ccccccccccc}
		\hline
		Model &\tabincell{c}{Reasonable\\all} &\tabincell{c}{Reasonable\\day} &\tabincell{c}{Reasonable\\night} &\tabincell{c}{Near\\scale} &\tabincell{c}{Medium\\scale}  &\tabincell{c}{Far\\scale}  &\tabincell{c}{No\\occlusion} &\tabincell{c}{Partial\\occlusion} &\tabincell{c}{Heavy\\occlusion} &\tabincell{c}{Inference\\speed (fps)}\\
		\hline
		RPN-HMFFN-640 &0.756	&0.761	&0.741	&0.607	&0.662	&0.065	&0.705	&0.263	&0.149 &9.4 \\
		HMFFN-640 &0.854 &0.865 &0.836 &0.797 &0.785 &0.166 &0.832 &0.391 &0.171 &10.8 \\
		RPN-HMFFN-480 &0.75	&0.755	&0.743	&0.591	&0.64	&0.046	&0.7	&0.282	&0.142 &16.5 \\
		HMFFN-480 &0.843 &0.866 &0.805 &0.796 &0.764 &0.148 &0.818 &0.373 &0.152 &18.5 \\
		RPN-HMFFN-320 &0.718	&0.717	&0.713	&0.638	&0.571	&0.057	&0.672	&0.225	&0.124 &32.0 \\
		HMFFN-320 &0.817 &0.825 &0.808 &0.779 &0.696 &0.111 &0.779 &0.345 &0.140 &38.3 \\
		\hline
	\end{tabular}
	\label{tab12}
\end{table*}

\begin{figure*}[!ht]
	
	\fontsize{8pt}{8pt}\selectfont
	\begin{minipage}{0.105\linewidth}
		\centering {Near scale\\no occlusion}
		\centering {\includegraphics[width=1\linewidth,clip]{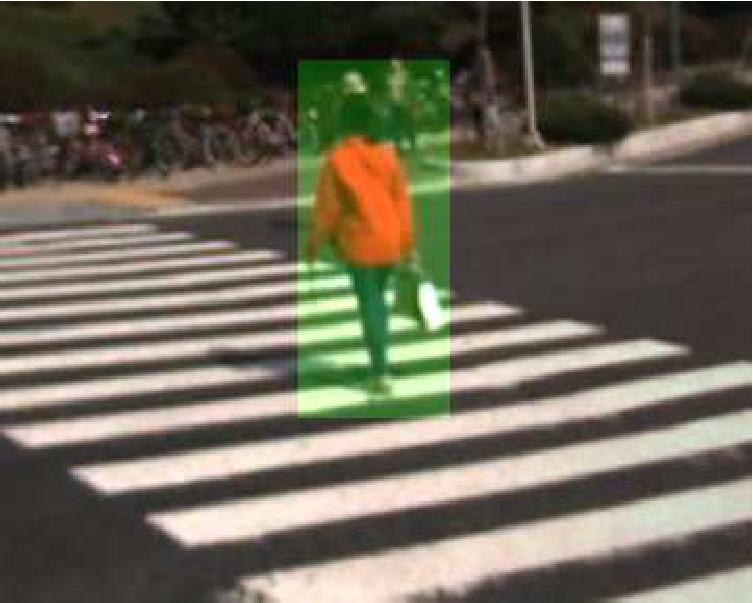}}
	\end{minipage}
	\begin{minipage}{0.105\linewidth}
		\centering {Near scale\\partial occlusion}
		\centering {\includegraphics[width=1\linewidth,clip]{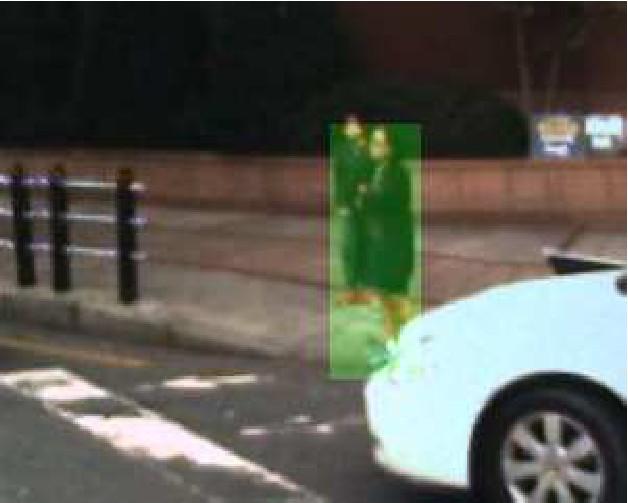}}
	\end{minipage}
	\begin{minipage}{0.105\linewidth}
		\centering {Near scale \\heavy occlusion}
		\centering {\includegraphics[width=1\linewidth,clip]{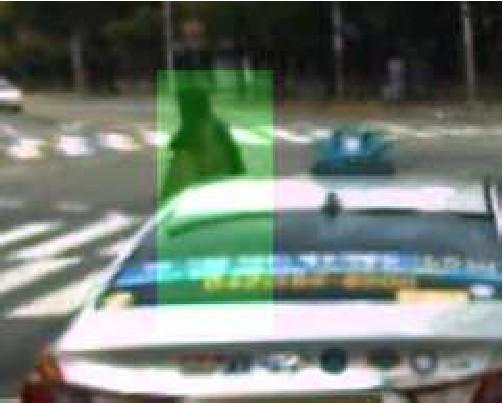}}
	\end{minipage}
	\begin{minipage}{0.105\linewidth}
		\centering {Medium scale\\no occlusion}
		\centering {\includegraphics[width=1\linewidth,clip]{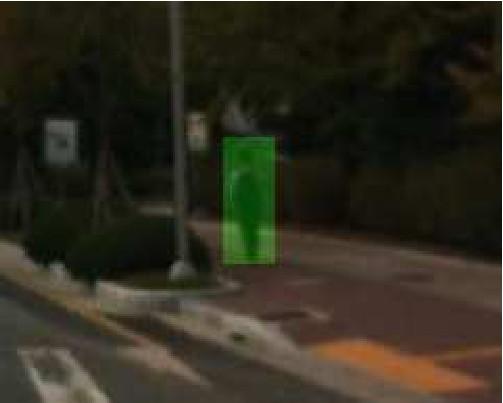}}
	\end{minipage}
	\begin{minipage}{0.105\linewidth}
		\centering {Medium scale\\partial occlusion}
		\centering {\includegraphics[width=1\linewidth,clip]{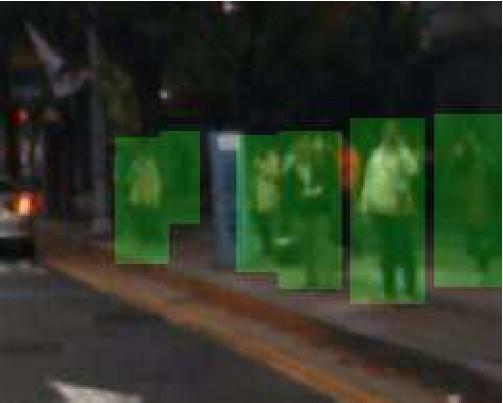}}
	\end{minipage}
	\begin{minipage}{0.105\linewidth}
		\centering {Medium scale\\heavy occlusion}
		\centering {\includegraphics[width=1\linewidth,clip]{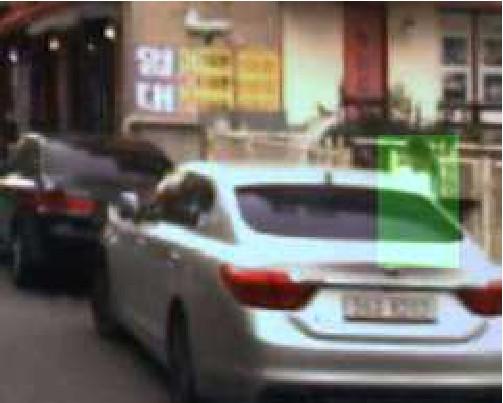}}
	\end{minipage}
	\centering
	\begin{minipage}{0.105\linewidth}
		\centering {far scale\\no occlusion}
		\centering {\includegraphics[width=1\linewidth,clip]{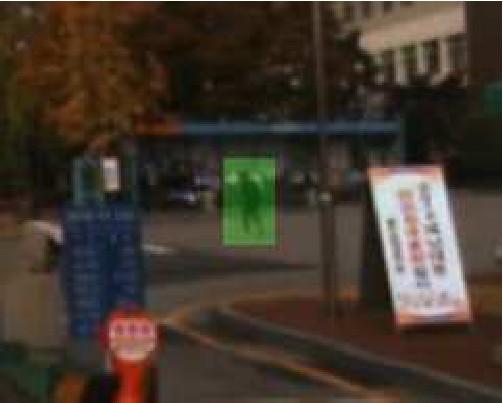}}
	\end{minipage}
	\begin{minipage}{0.105\linewidth}
		\centering {far scale\\partial occlusion}
		\centering {\includegraphics[width=1\linewidth,clip]{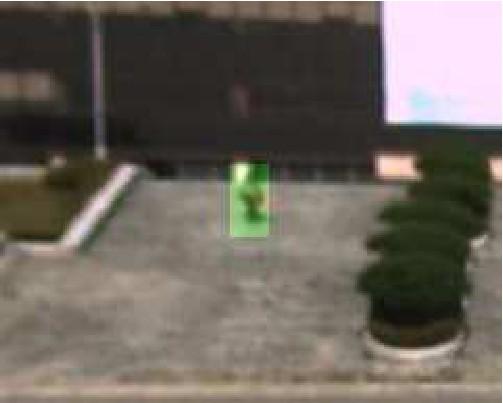}}
	\end{minipage}
	\begin{minipage}{0.105\linewidth}
		\centering {far scale\\heavy occlusion}
		\centering {\includegraphics[width=1\linewidth,clip]{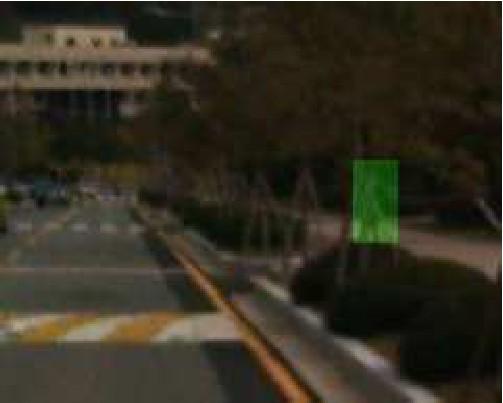}}
	\end{minipage}
	\vspace{1mm}\\
	
	\begin{minipage}{0.105\linewidth}
		\centering {\includegraphics[width=1\linewidth,clip]{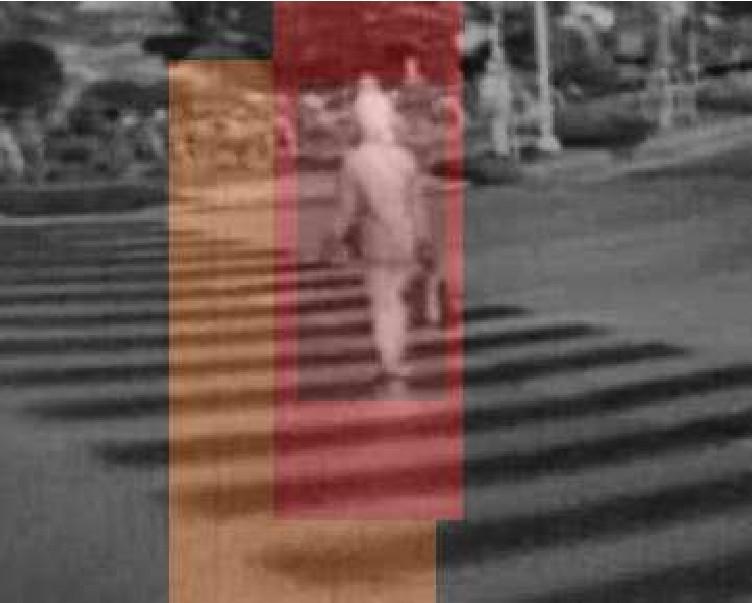}}
	\end{minipage}
	\begin{minipage}{0.105\linewidth}
		\centering {\includegraphics[width=1\linewidth,clip]{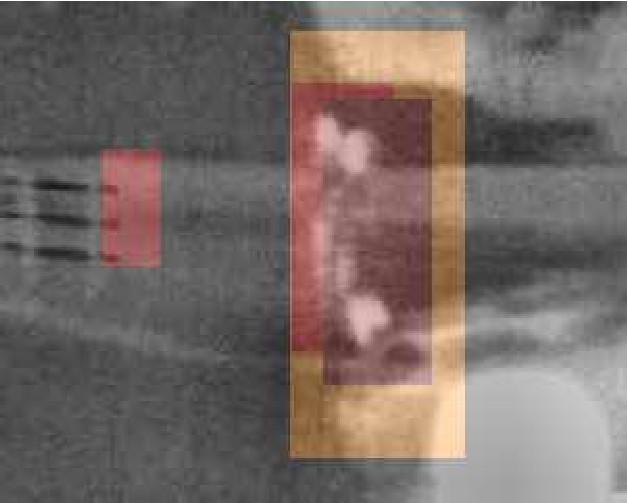}}
	\end{minipage}
	\begin{minipage}{0.105\linewidth}
		\centering {\includegraphics[width=1\linewidth,clip]{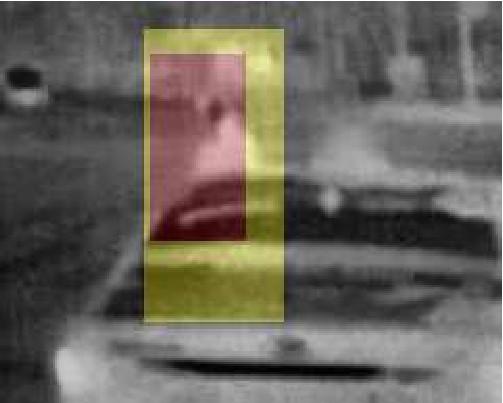}}
	\end{minipage}
	\begin{minipage}{0.105\linewidth}
		\centering {\includegraphics[width=1\linewidth,clip]{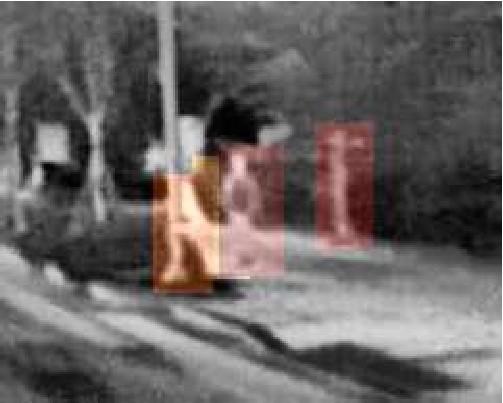}}
	\end{minipage}
	\begin{minipage}{0.105\linewidth}
		\centering {\includegraphics[width=1\linewidth,clip]{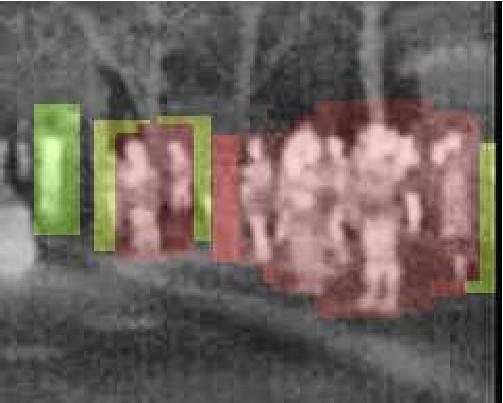}}
	\end{minipage}
	\begin{minipage}{0.105\linewidth}
		\centering {\includegraphics[width=1\linewidth,clip]{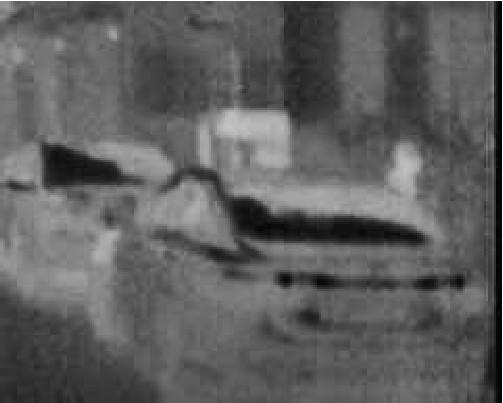}}
	\end{minipage}
	\centering
	\begin{minipage}{0.105\linewidth}
		\centering {\includegraphics[width=1\linewidth,clip]{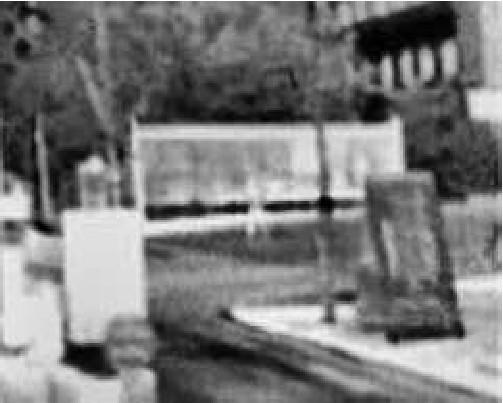}}
	\end{minipage}
	\begin{minipage}{0.105\linewidth}
		\centering {\includegraphics[width=1\linewidth,clip]{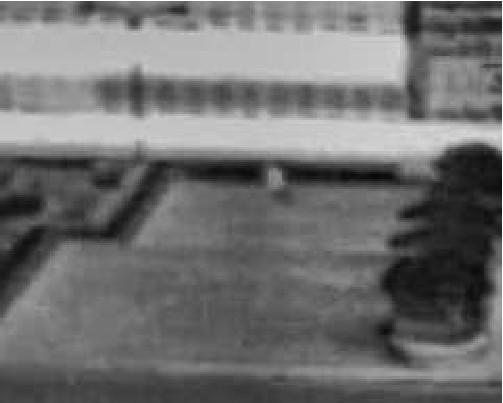}}
	\end{minipage}
	\begin{minipage}{0.105\linewidth}
		\centering {\includegraphics[width=1\linewidth,clip]{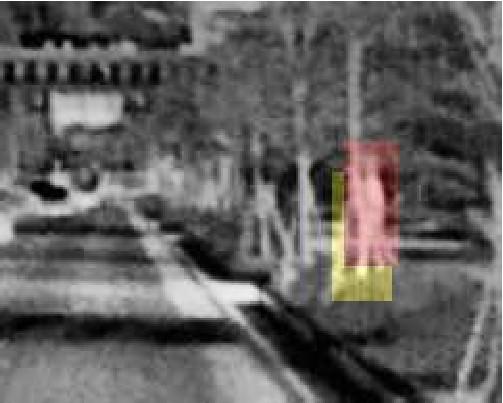}}
	\end{minipage}
	\vspace{1mm}\\
	
	\begin{minipage}{0.105\linewidth}
		\centering {\includegraphics[width=1\linewidth,clip]{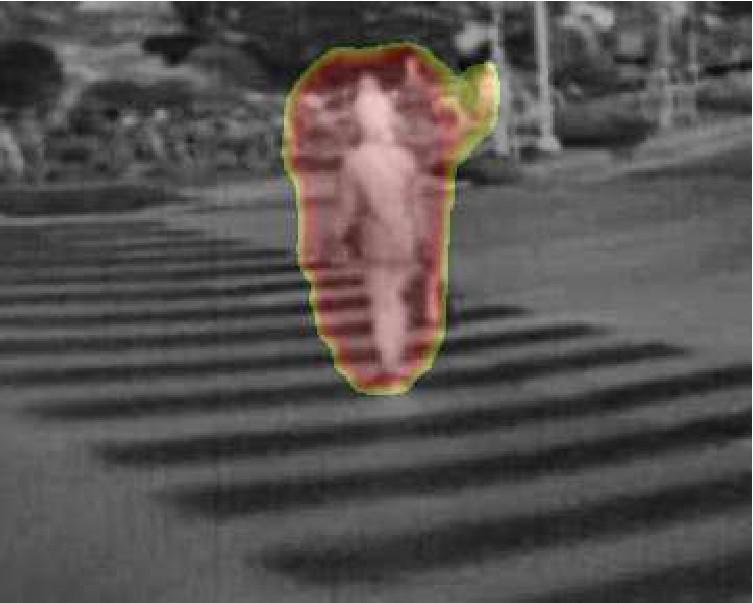}}
	\end{minipage}
	\begin{minipage}{0.105\linewidth}
		\centering {\includegraphics[width=1\linewidth,clip]{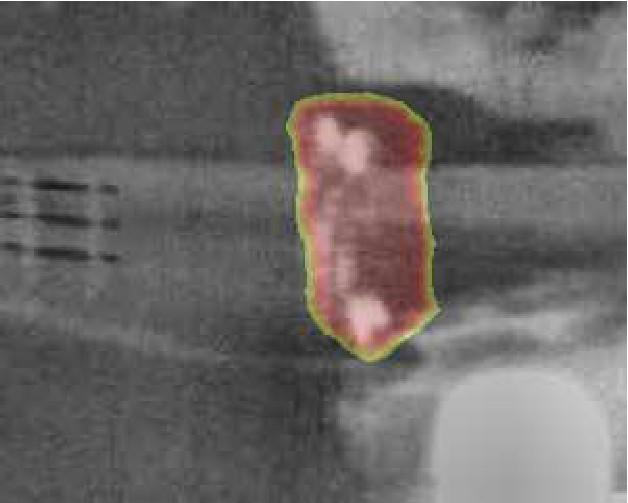}}
	\end{minipage}
	\begin{minipage}{0.105\linewidth}
		\centering {\includegraphics[width=1\linewidth,clip]{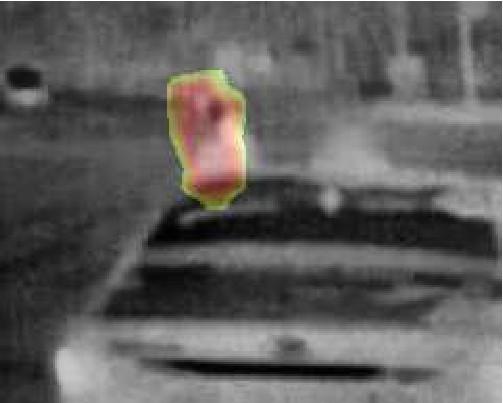}}
	\end{minipage}
	\begin{minipage}{0.105\linewidth}
		\centering {\includegraphics[width=1\linewidth,clip]{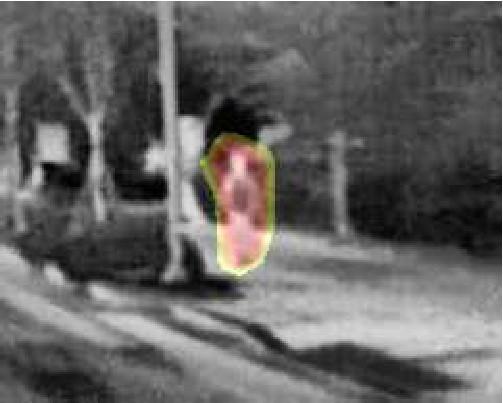}}
	\end{minipage}
	\begin{minipage}{0.105\linewidth}
		\centering {\includegraphics[width=1\linewidth,clip]{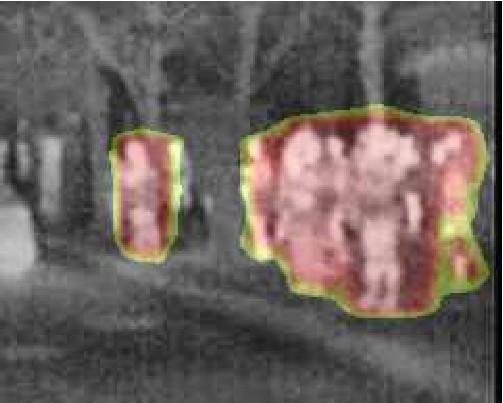}}
	\end{minipage}
	\begin{minipage}{0.105\linewidth}
		\centering {\includegraphics[width=1\linewidth,clip]{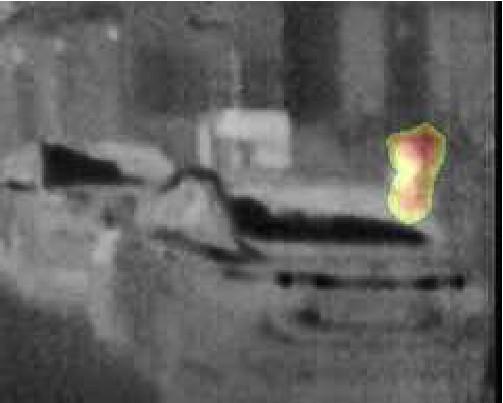}}
	\end{minipage}
	\centering
	\begin{minipage}{0.105\linewidth}
		\centering {\includegraphics[width=1\linewidth,clip]{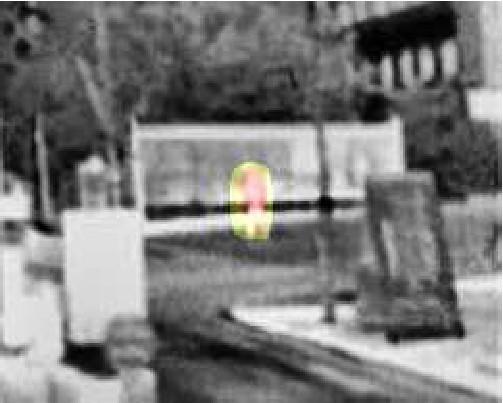}}
	\end{minipage}
	\begin{minipage}{0.105\linewidth}
		\centering {\includegraphics[width=1\linewidth,clip]{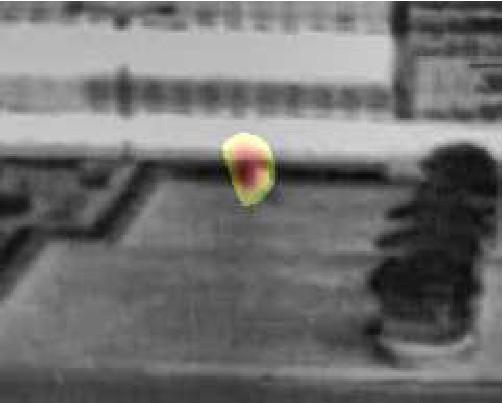}}
	\end{minipage}
	\begin{minipage}{0.105\linewidth}
		\centering {\includegraphics[width=1\linewidth,clip]{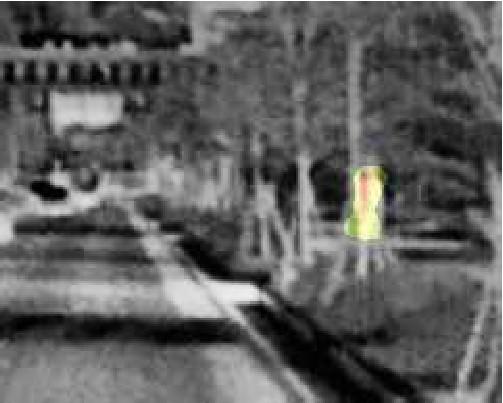}}
	\end{minipage}
	\vspace{1mm}\\
	
	{\bf (a) Daytime}
	\vspace{1mm}\\
	
	\begin{minipage}{0.105\linewidth}
		\centering {\includegraphics[width=1\linewidth,clip]{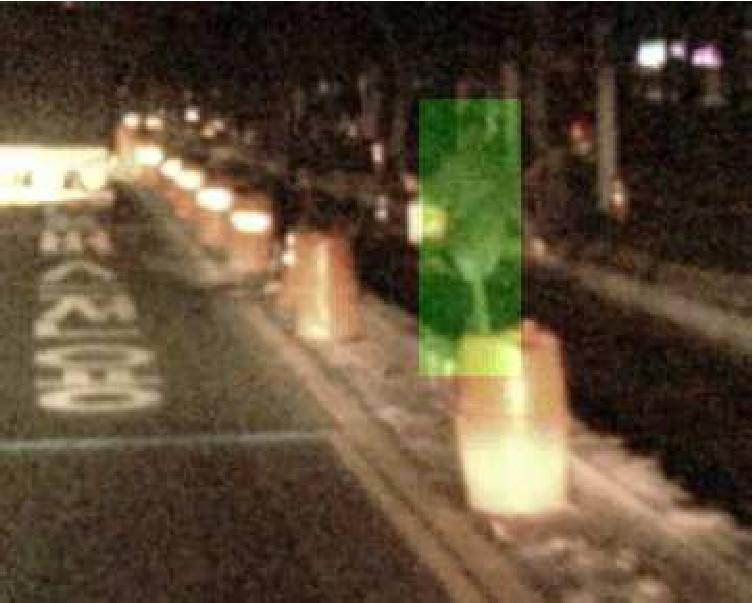}}
	\end{minipage}
	\begin{minipage}{0.105\linewidth}
		\centering {\includegraphics[width=1\linewidth,clip]{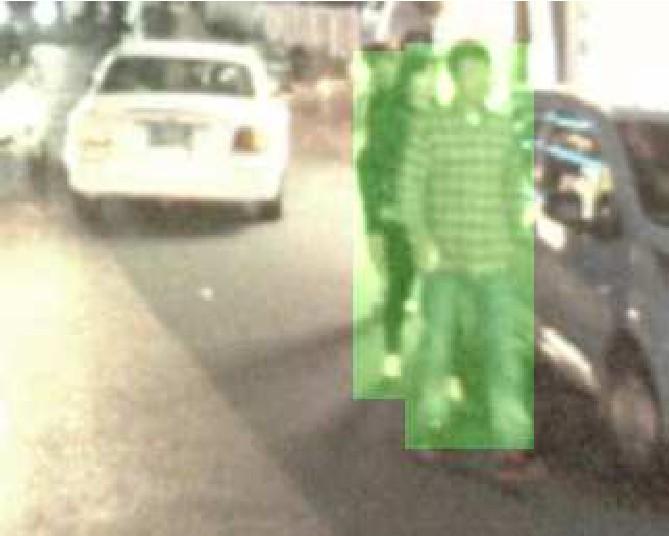}}
	\end{minipage}
	\begin{minipage}{0.105\linewidth}
		\centering {\includegraphics[width=1\linewidth,clip]{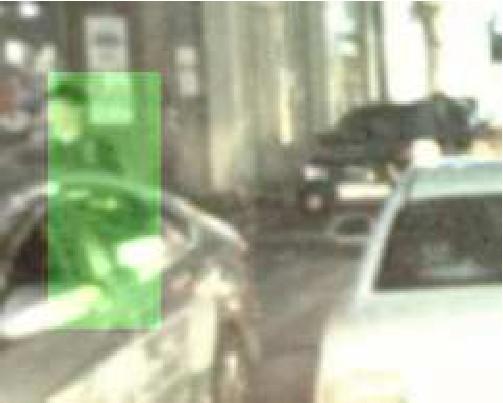}}
	\end{minipage}
	\begin{minipage}{0.105\linewidth}
		\centering {\includegraphics[width=1\linewidth,clip]{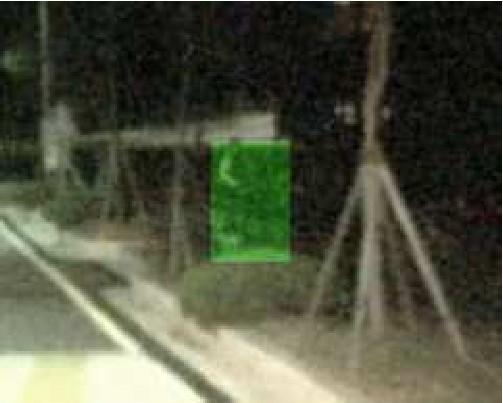}}
	\end{minipage}
	\begin{minipage}{0.105\linewidth}
		\centering {\includegraphics[width=1\linewidth,clip]{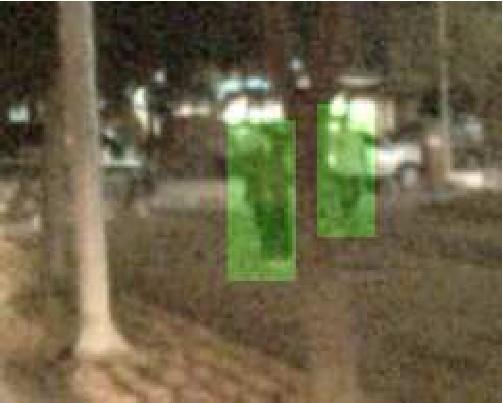}}
	\end{minipage}
	\begin{minipage}{0.105\linewidth}
		\centering {\includegraphics[width=1\linewidth,clip]{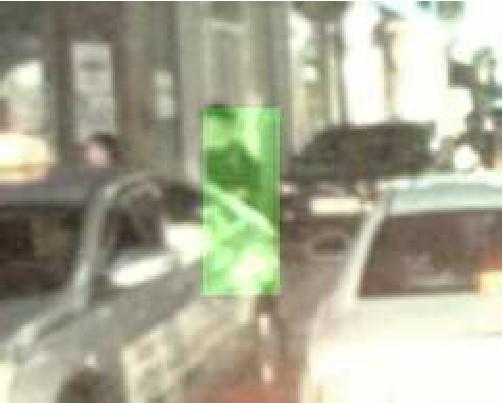}}
	\end{minipage}
	\centering
	\begin{minipage}{0.105\linewidth}
		\centering {\includegraphics[width=1\linewidth,clip]{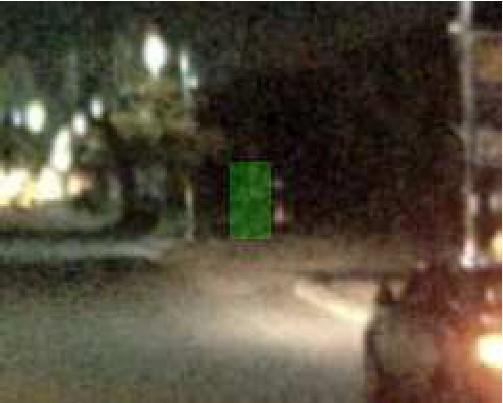}}
	\end{minipage}
	\begin{minipage}{0.105\linewidth}
		\centering {\includegraphics[width=1\linewidth,clip]{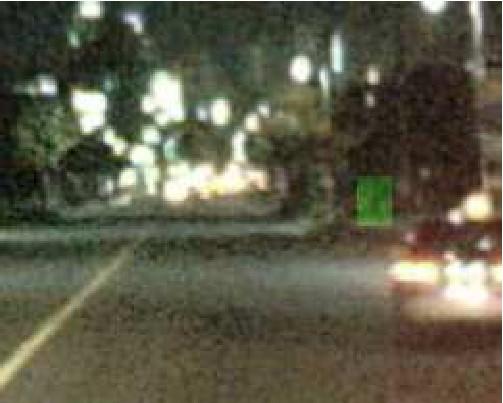}}
	\end{minipage}
	\begin{minipage}{0.105\linewidth}
		\centering {\includegraphics[width=1\linewidth,clip]{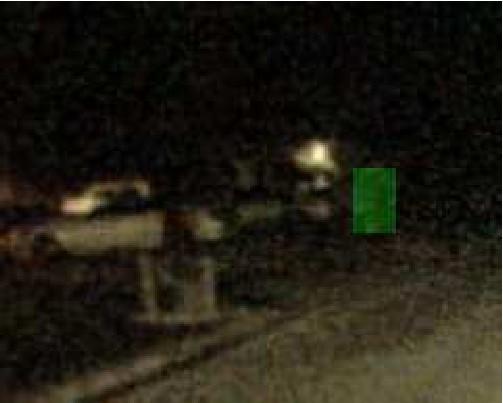}}
	\end{minipage}
	\vspace{1mm}\\
	
	\begin{minipage}{0.105\linewidth}
		\centering {\includegraphics[width=1\linewidth,clip]{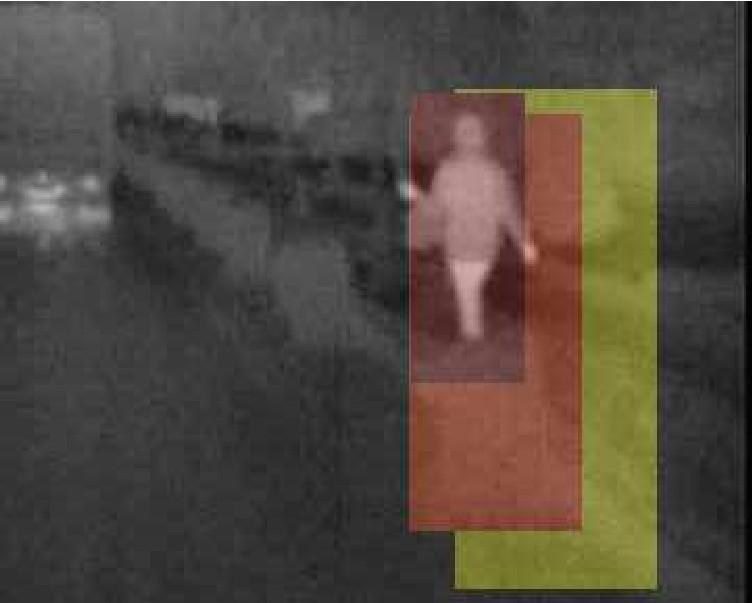}}
	\end{minipage}
	\begin{minipage}{0.105\linewidth}
		\centering {\includegraphics[width=1\linewidth,clip]{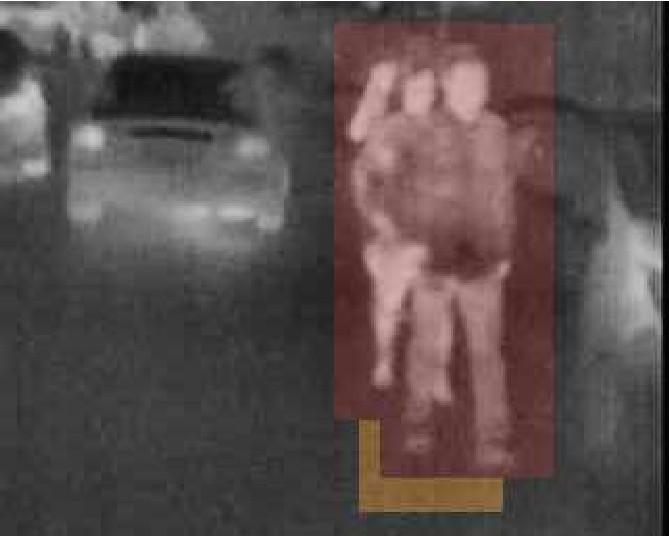}}
	\end{minipage}
	\begin{minipage}{0.105\linewidth}
		\centering {\includegraphics[width=1\linewidth,clip]{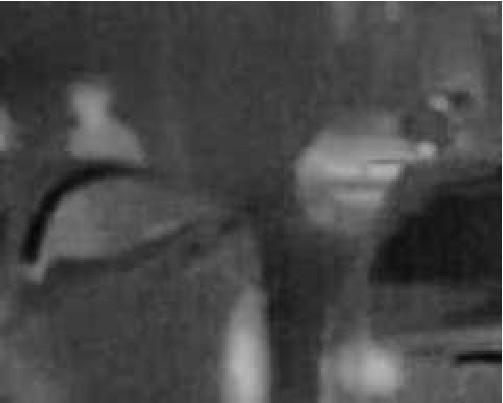}}
	\end{minipage}
	\begin{minipage}{0.105\linewidth}
		\centering {\includegraphics[width=1\linewidth,clip]{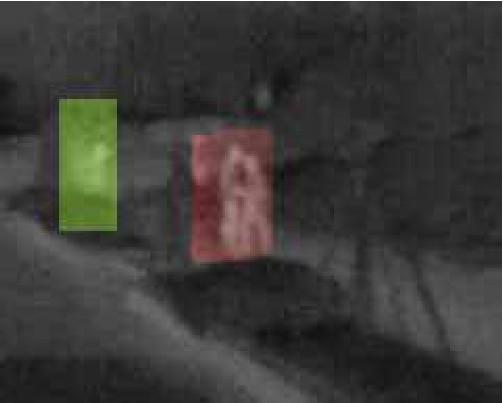}}
	\end{minipage}
	\begin{minipage}{0.105\linewidth}
		\centering {\includegraphics[width=1\linewidth,clip]{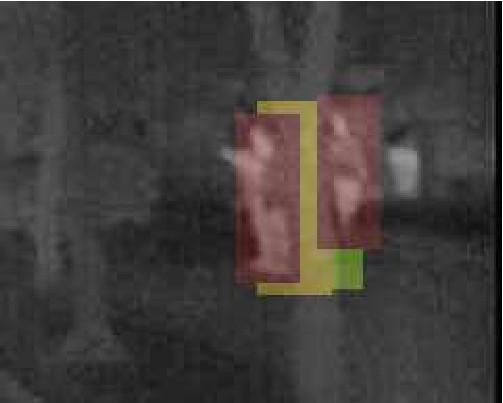}}
	\end{minipage}
	\begin{minipage}{0.105\linewidth}
		\centering {\includegraphics[width=1\linewidth,clip]{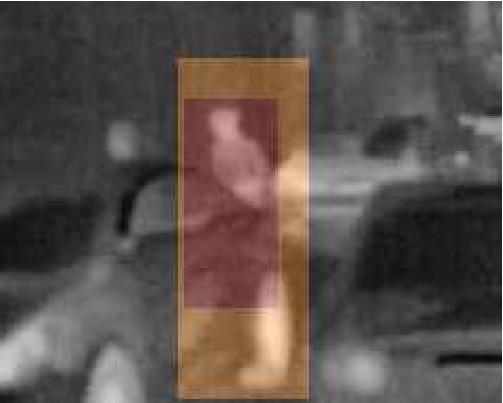}}
	\end{minipage}
	\centering
	\begin{minipage}{0.105\linewidth}
		\centering {\includegraphics[width=1\linewidth,clip]{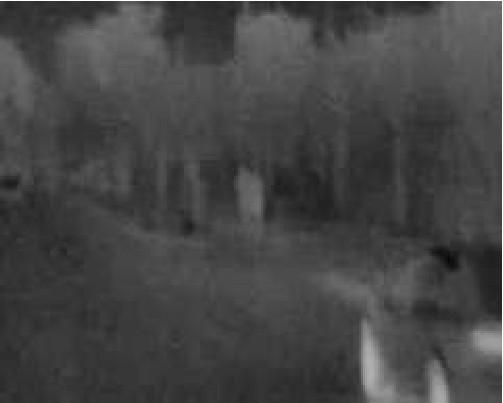}}
	\end{minipage}
	\begin{minipage}{0.105\linewidth}
		\centering {\includegraphics[width=1\linewidth,clip]{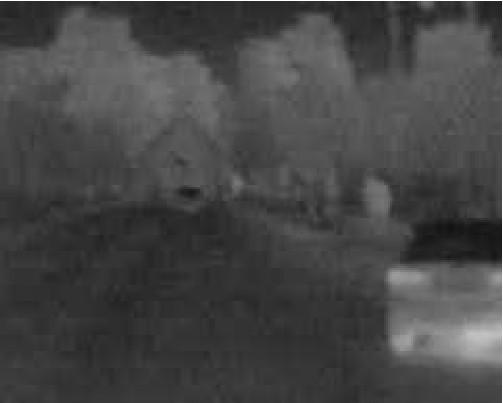}}
	\end{minipage}
	\begin{minipage}{0.105\linewidth}
		\centering {\includegraphics[width=1\linewidth,clip]{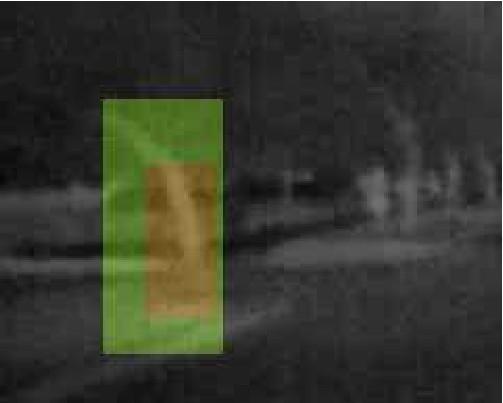}}
	\end{minipage}
	\vspace{1mm}\\
	
	\begin{minipage}{0.105\linewidth}
		\centering {\includegraphics[width=1\linewidth,clip]{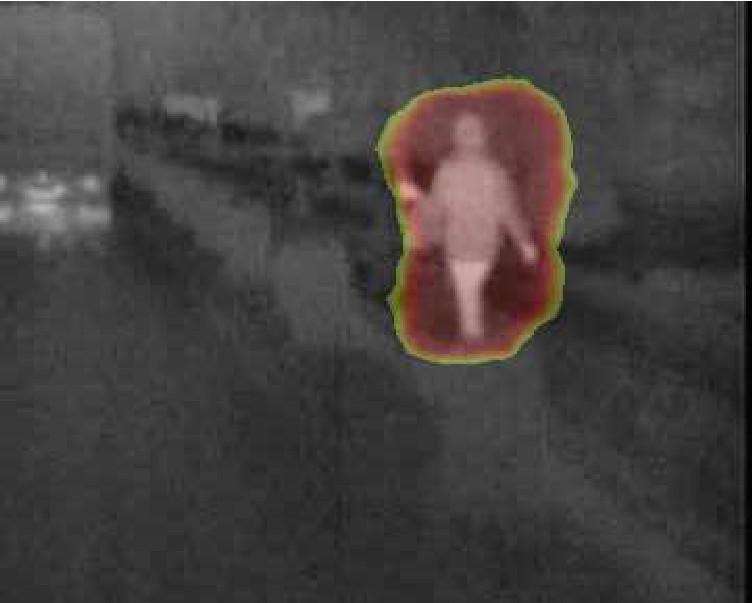}}
	\end{minipage}
	\begin{minipage}{0.105\linewidth}
		\centering {\includegraphics[width=1\linewidth,clip]{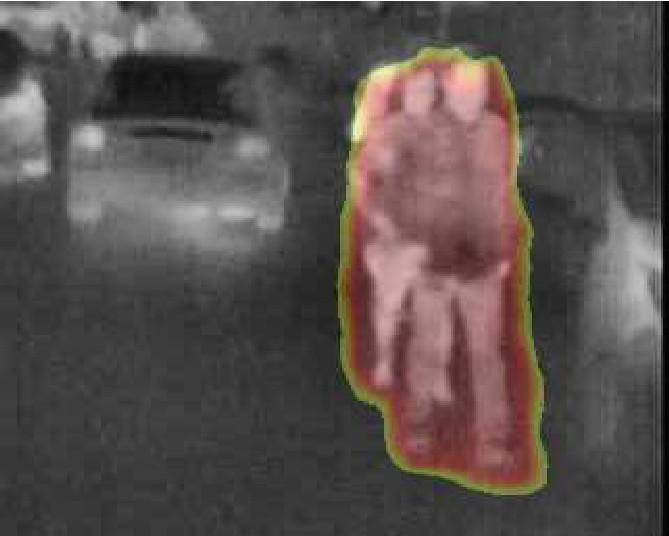}}
	\end{minipage}
	\begin{minipage}{0.105\linewidth}
		\centering {\includegraphics[width=1\linewidth,clip]{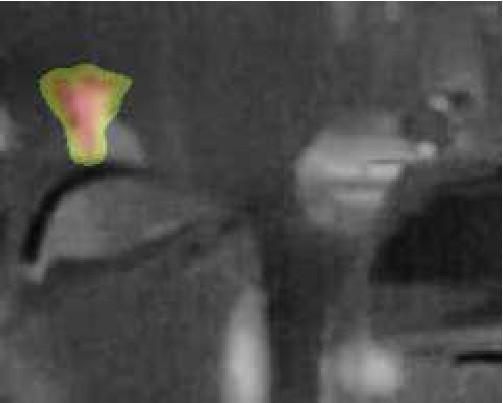}}
	\end{minipage}
	\begin{minipage}{0.105\linewidth}
		\centering {\includegraphics[width=1\linewidth,clip]{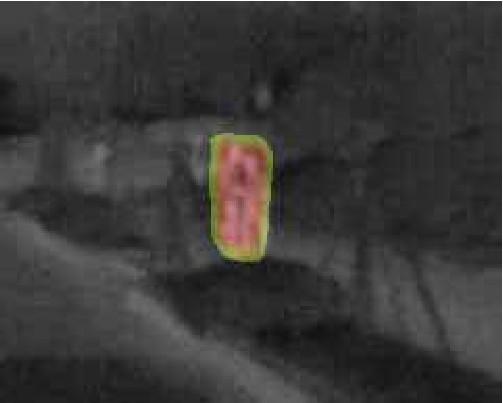}}
	\end{minipage}
	\begin{minipage}{0.105\linewidth}
		\centering {\includegraphics[width=1\linewidth,clip]{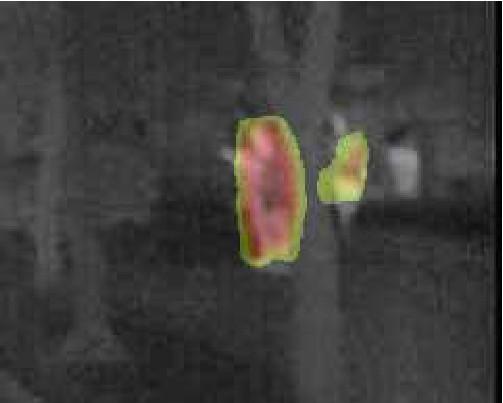}}
	\end{minipage}
	\begin{minipage}{0.105\linewidth}
		\centering {\includegraphics[width=1\linewidth,clip]{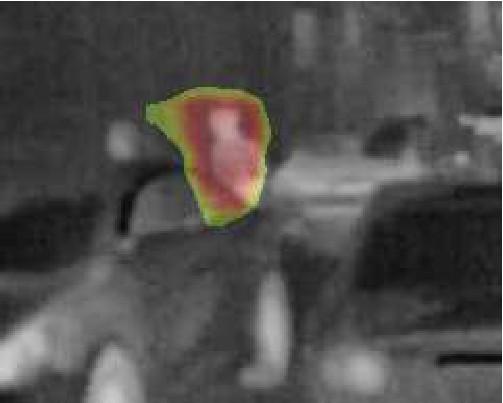}}
	\end{minipage}
	\centering
	\begin{minipage}{0.105\linewidth}
		\centering {\includegraphics[width=1\linewidth,clip]{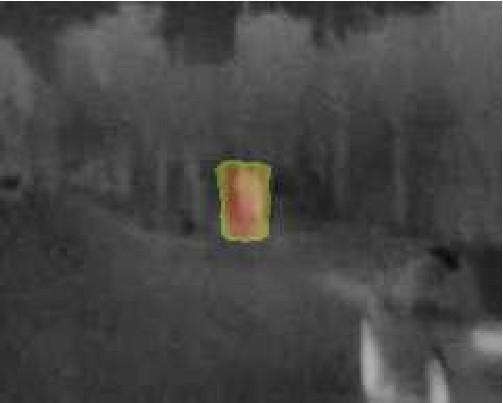}}
	\end{minipage}
	\begin{minipage}{0.105\linewidth}
		\centering {\includegraphics[width=1\linewidth,clip]{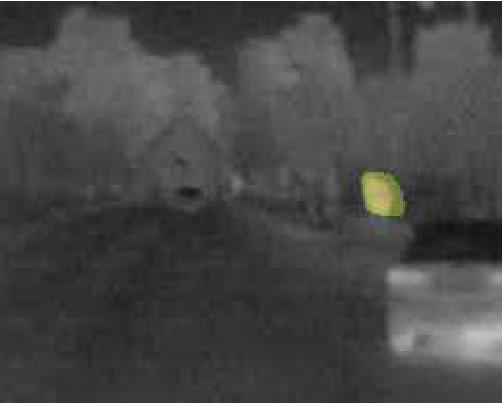}}
	\end{minipage}
	\begin{minipage}{0.105\linewidth}
		\centering {\includegraphics[width=1\linewidth,clip]{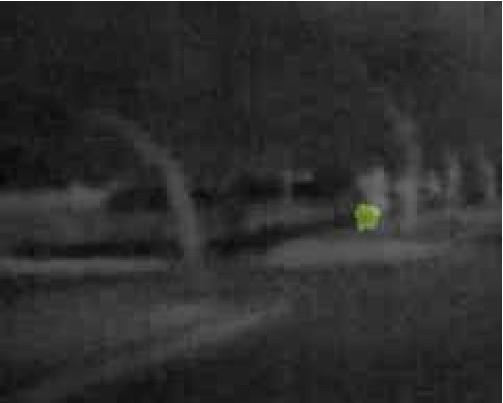}}
	\end{minipage}
	\vspace{1mm}\\
	
	{\bf (b) Nighttime}\\
	
	\centering
	\caption{Qualitative comparison of multispectral pedestrian detection results of RPN-HMFFN-640 and HMFFN-640 in the KAIST testing dataset. First row shows the ground truth (displaying using the visible channel) and the others show detection results of RPN-HMFFN-640 and HMFFN-640 respectively (displaying using the infrared channel). Note that the green regions represent ground-truth annotation masks which are generated based on manually labeled bounding boxes, and the detected pedestrian targets are visualized using the heat map representation with a 0.5 threshold. Best viewed in color.}
	\label{fig8}
\end{figure*}

\subsection{Dataset and Evaluation Metric}

All the detectors are evaluated using the public KAIST multispectral pedestrian benchmark \cite{hwang2015multispectral}. We notice that CVC-14 \cite{gonzalez2016pedestrian} is another newly published multispectral pedestrian benchmark consisting of infrared and visible gray image pairs. However, the multispectral image pairs were not properly aligned thus the pedestrian annotations are individually labeled in infrared and visible images. It should be noted that some annotations are only generated in the infrared/visible image on the CVC-14 dataset. To the best of our knowledge, KAIST multispectral pedestrian benchmark is the only available pedestrian dataset which contains large-scale and well-aligned visible-infrared image pairs with accurate manual annotations. 


Totally, KAIST training dataset consists of 50,172 well aligned visible-infrared image pairs ($640 \times 512$ resolution) captured in all-day traffic scenes with 13,853 pedestrian annotations. The training images are sampled in every 2 frames following the other multispectral pedestrian detection methods \cite{liu2016multispectral,konig2017fully,guan2018fusion,guan2018exploiting,li2018multispectral}. The KAIST testing dataset contains 2,252 image pairs with 1,356 pedestrian annotations. Since the original KAIST testing dataset contains many problematic annotations (e.g., inaccurate bounding boxes and missed human targets), we make use of the improved annotations provided by Liu et al. \cite{liu2018improved} for quantitative and qualitative evaluation. Specifically, we consider all reasonable, scale, and occlusion subset of the KAIST testing dataset \cite{hwang2015multispectral}. 

{

The output of our method is a full-size prediction heat map in which human target regions yields high confident scores while the background regions produce low ones. For a fair comparison, we transform the bounding box detection results with different prediction scores to the heat map representation, and the pixel-level average precision (AP) \cite{salton1986introduction,cordts2016cityscapes} is utilized to evaluate the quantitative performance of multispectral pedestrian detectors in the pixel-level. The computed detection results are compared with the ground-truth annotation masks which are generated based on manually labeled bounding boxes. Pixels located in the ground-truth bounding boxes are defined as foreground ones, while other pixels are defined as background ones. Given the heat map predictions, true positive (TP) is the number of correctly predicted foreground pixels, false positive (FP) is the number of incorrectly predicted background pixels, and false negative (FN) is the number of incorrectly foreground background pixels. Precision is calculated as TP/(TP+FP) and recall is computed as TP/(TP+FN). The AP depicts the shape of the precision/recall curve, and is defined as the mean precision at a number of equally spaced recall levels by varying the threshold on detection scores.}
In our implementation, we average the precision values at 100 recall levels equally spaced between 0 and 1.

\subsection{Implementation Details}
The image-centric training and testing strategy are applied to generate mini-batches without using image pyramids. The batch size is set to 1 according to the method presented by Guan \emph{et al.} \cite{guan2018exploiting}. Each stream of the feature extraction layers in MFFN and HMFFN are initialized using the weights and bias of VGG-16 net \cite{simonyan2014very}  pre-trained on the ImageNet dataset \cite{russakovsky2015imagenet}.  All the other convolutional layers use normalized initialization following the method presented by Xavier \cite{glorot2010understanding}. We utilize the Caffe \cite{jia2014caffe} deep learning framework to train and test our proposed multispectral pedestrian detectors. All the models are fine-tuned using stochastic gradient descent (SGD) \cite{zinkevich2010parallelized} for the first two epochs with the learning rate of 0.001 and one more epoch with the learning rate of 0.0001. Adjustable gradient clipping technique is used in training to suppress exploding gradients \cite{pascanu2013difficulty}.

\begin{table*}[ht]
	\centering
	\caption{Quantitative comparison of HMFFN-640 and HMFFN-320 with the current state-of-the-art methods \cite{konig2017fully, Liu2016BMVC, guan2018fusion, guan2018exploiting, li2018multispectral}. Input sizes of different models are Halfway Fusion - $750 \times 600$, Fusion RPN+BDT - $960 \times 768$, IATDNN+IAMSS - $960 \times 768$, FRPN-Sum+TSS - $960 \times 768$, MSDS-RCNN - $750 \times 600$, HMFFN-640 - $640 \times 512$, and HMFFN-320 - $320 \times 256$. The top three results are highlighted in { red}, {\color{green} green},and {\color{blue} blue}, respectively.}
	\fontsize{8pt}{8pt}\selectfont
	\begin{tabular}{ccccccccccc}
		\hline
		Model &\tabincell{c}{Reasonable\\all} &\tabincell{c}{Reasonable\\day} &\tabincell{c}{Reasonable\\night} &\tabincell{c}{Near\\scale} &\tabincell{c}{Medium\\scale}  &\tabincell{c}{Far\\scale}  &\tabincell{c}{No\\occlusion} &\tabincell{c}{Partial\\occlusion} &\tabincell{c}{Heavy\\occlusion} &\tabincell{c}{Inference\\speed (fps)}\\
		\hline
		{Halfway Fusion \cite{liu2016multispectral}} &0.702 &0.708 &0.691 &0.623 &0.583 &0.062 &0.695 &0.128 &0.037 &2.5 \\
		{Fusion RPN+BDT \cite{konig2017fully}} &0.755 &0.767 &0.731 &0.663 &\color{blue}0.681 &0.027 &0.700 &0.165 &0.030 &1.3 \\
		{IATDNN+IAMSS \cite{guan2018fusion}} &\color{blue}0.766 &\color{blue}0.772 &\color{blue}0.756 &0.614 &0.643 &0.043 &\color{blue}0.715 &0.263 &0.106 &4.0 \\
		{FRPN-Sum+TSS \cite{guan2018exploiting}} &0.765 &0.767 &0.750 &0.626 &0.638 &0.045 &0.714 &\color{blue}0.277 &\color{blue}0.116 &\color{blue}4.4 \\
		{MSDS-RCNN \cite{li2018multispectral}} &0.744 &0.750 &0.721 &\color{blue}0.670 &0.673 &\color{blue}0.068 &0.712 &0.206 &0.070 &\color{blue}4.4 \\
		\textbf{HMFFN-640 (ours)} &{0.854} &{0.865} &{0.836} &{0.797} &{0.785} &{0.166} &{0.832} &{0.391} &{0.171} &\color{green}{10.8} \\
		\textbf{HMFFN-320 (ours)} &\color{green}{0.817} &\color{green}{0.825} &\color{green}{0.808} &\color{green}{0.779} &\color{green}{0.696} &\color{green}{0.111} &\color{green}{0.779} &\color{green}{0.345} &\color{green}{0.140} &{38.3} \\
		\hline
	\end{tabular}
	\label{tab2}
\end{table*}

\begin{figure*}[!ht]
	
	\fontsize{8pt}{8pt}\selectfont
	\centering
	\begin{minipage}{0.16\linewidth}
		{Ground Truth}
		\centering
		{\includegraphics[width=1\linewidth,clip]{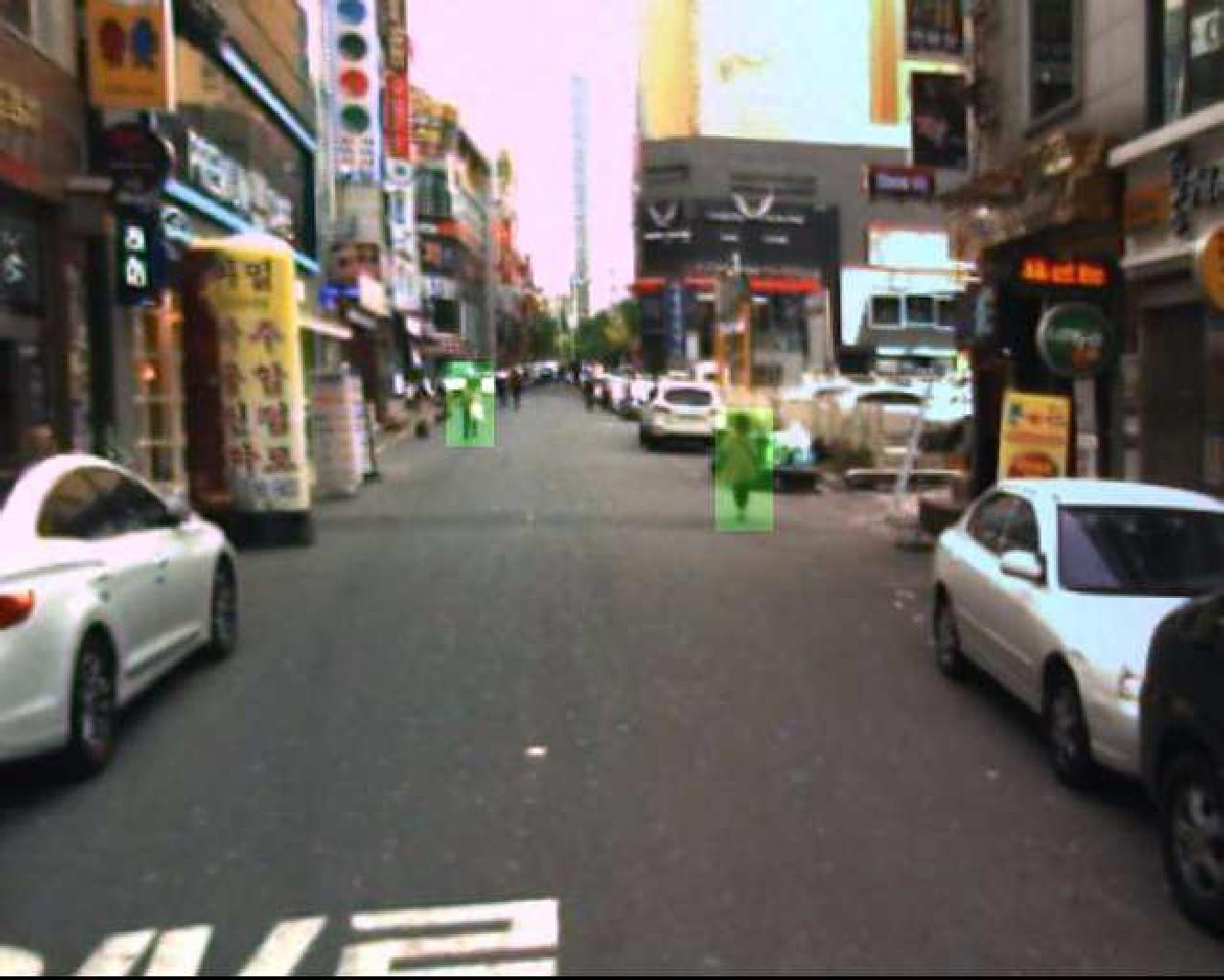}}
	\end{minipage}
	\begin{minipage}{0.16\linewidth}
		{FRPN+BDT \cite{konig2017fully}}
		\centering
		{\includegraphics[width=1\linewidth,clip]{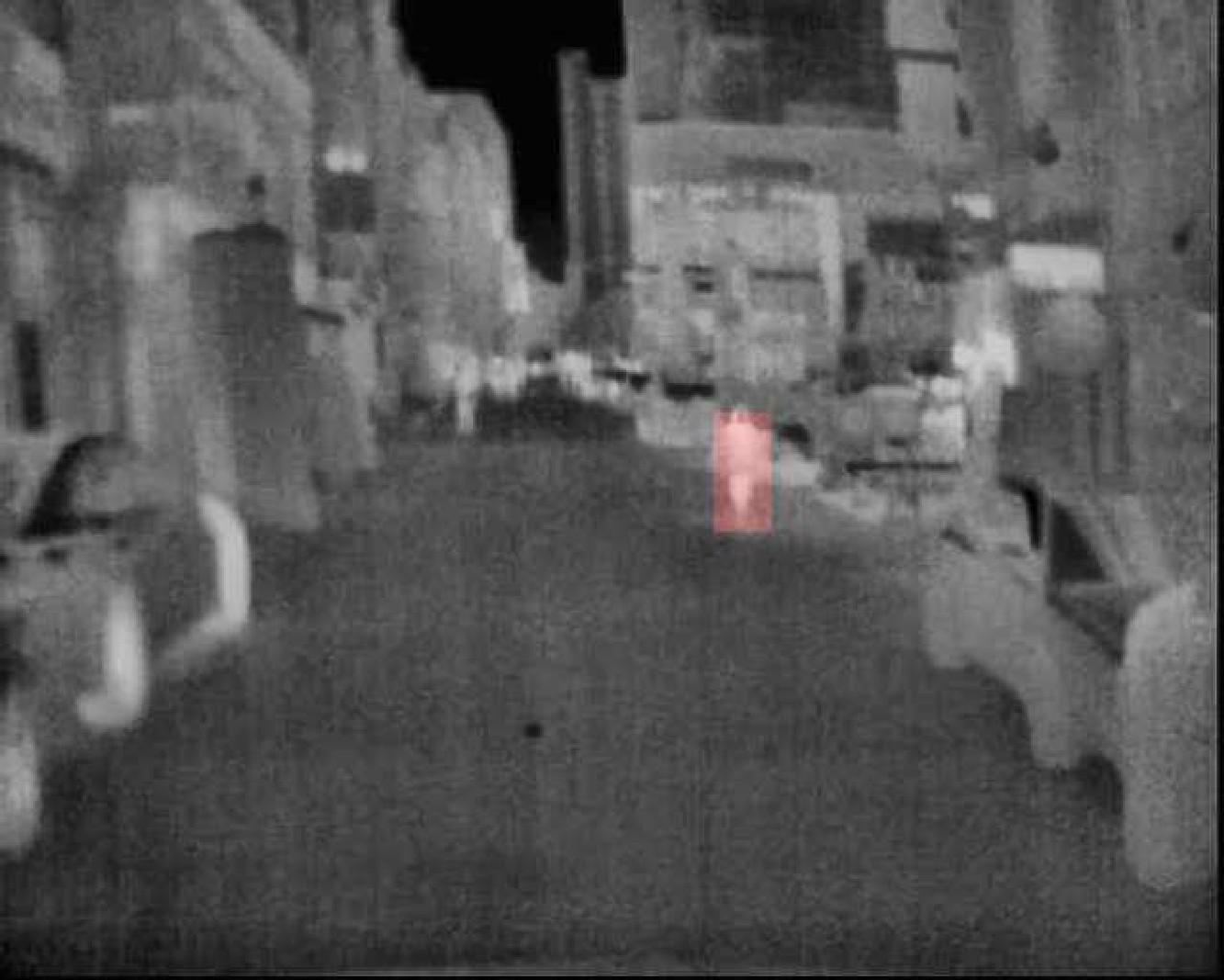}}
	\end{minipage}
	\begin{minipage}{0.16\linewidth}
		{IATDNN+IAMSS \cite{guan2018fusion}}
		\centering
		{\includegraphics[width=1\linewidth,clip]{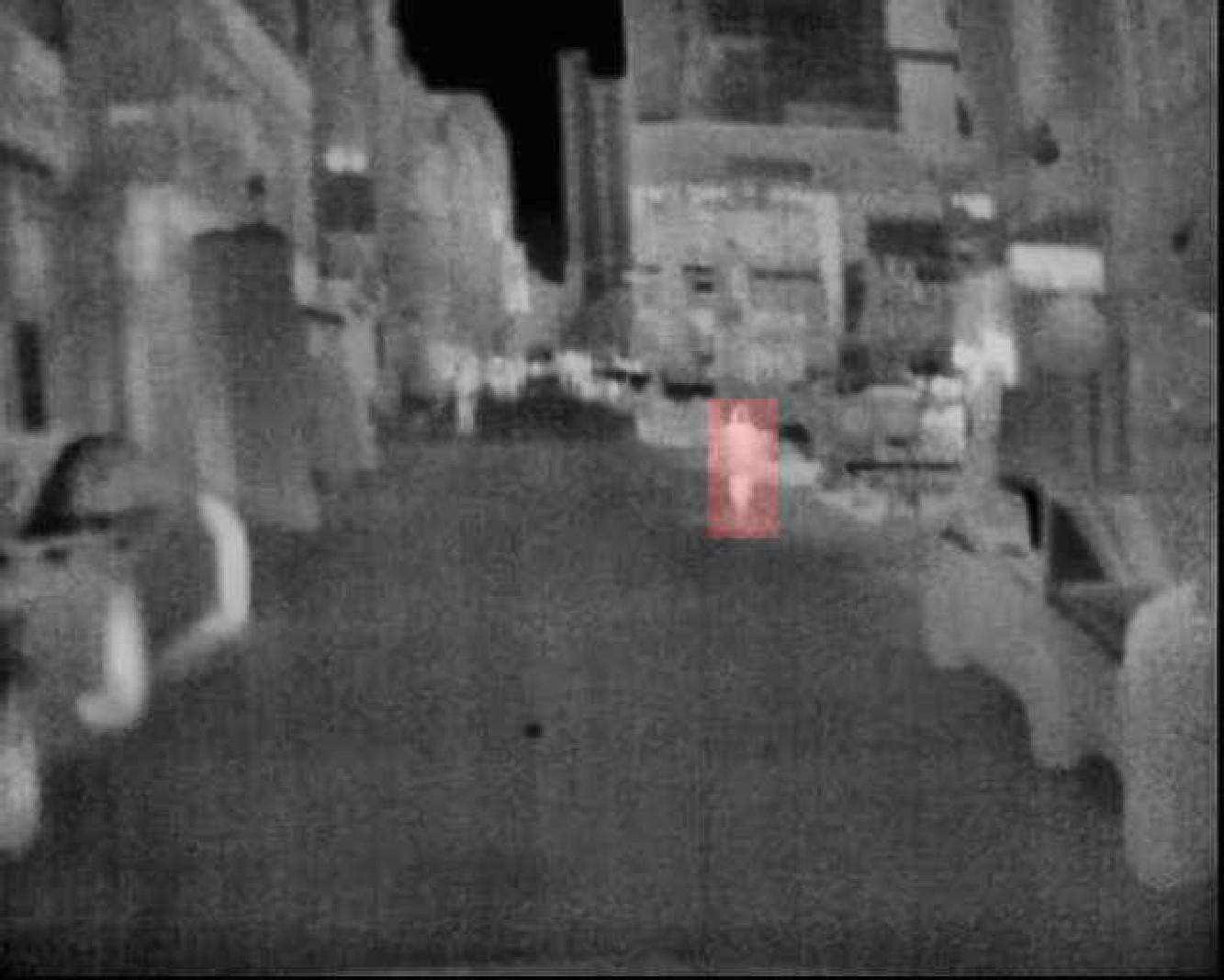}}
	\end{minipage}
	\begin{minipage}{0.16\linewidth}
		{MSDS-RCNN \cite{li2018multispectral}}
		\centering
		{\includegraphics[width=1\linewidth,clip]{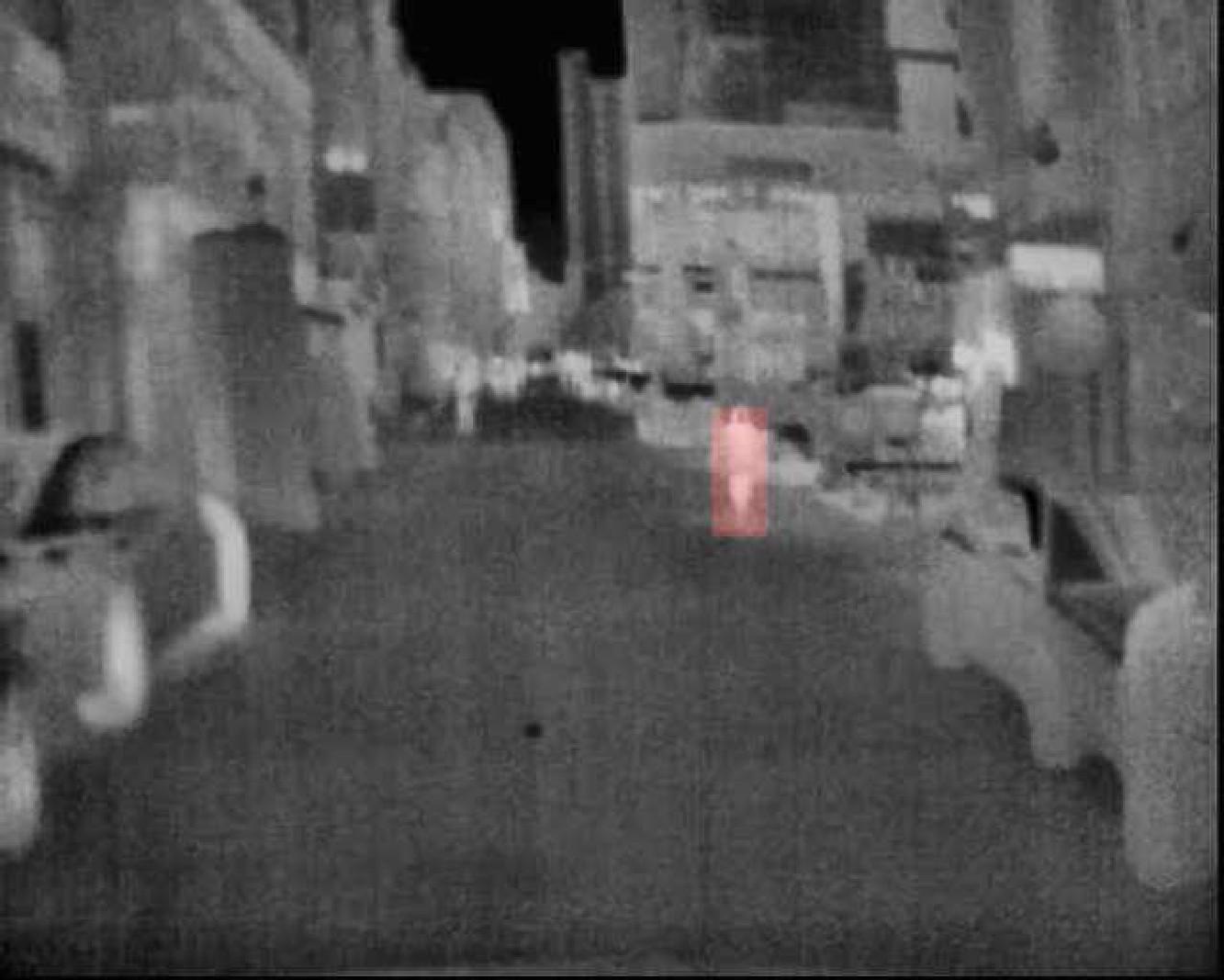}}
	\end{minipage}
	\begin{minipage}{0.16\linewidth}
		{\bf HMFFN-640 (ours)}
		\centering
		{\includegraphics[width=1\linewidth,clip]{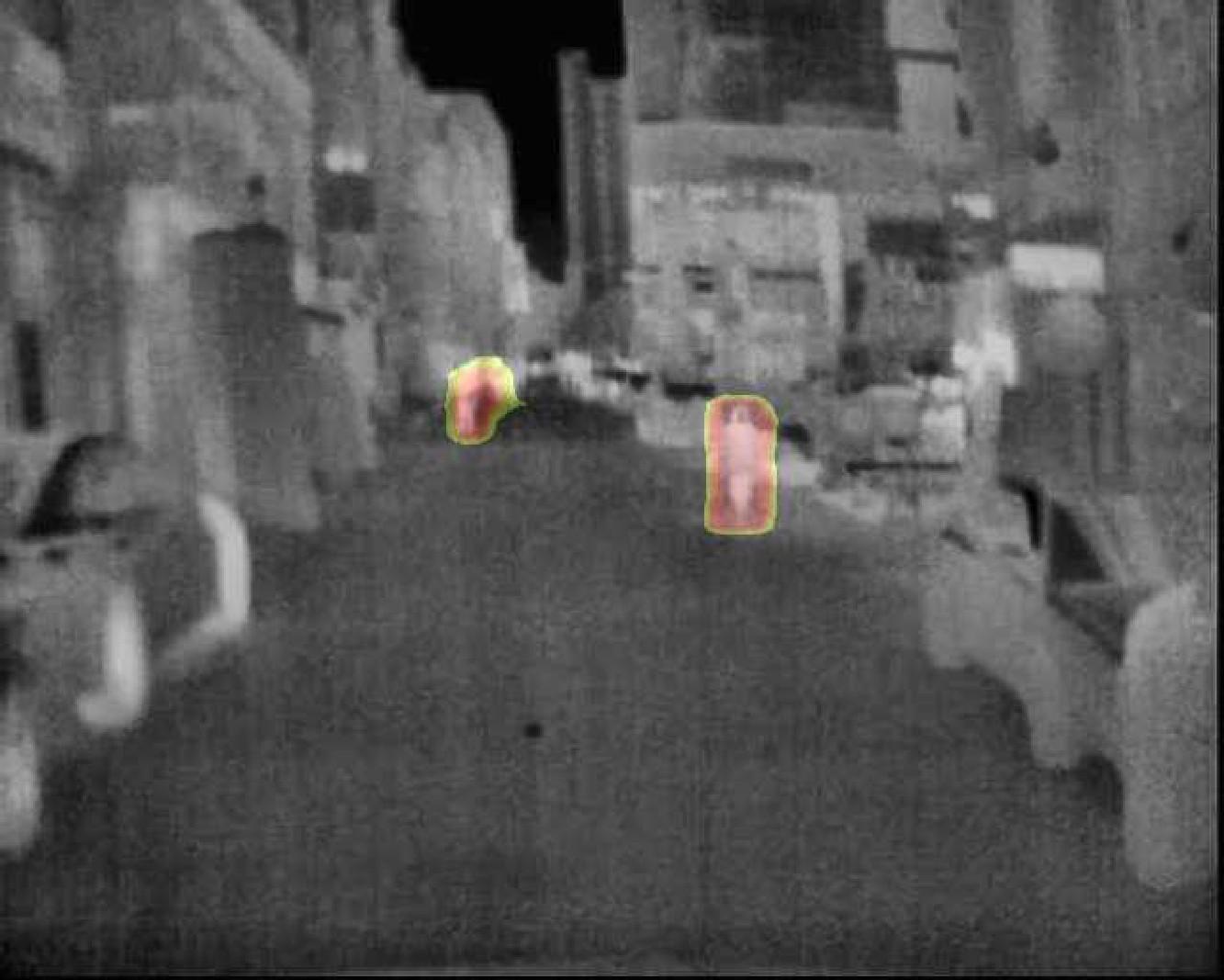}}
	\end{minipage}
	\begin{minipage}{0.16\linewidth}
		{\bf HMFFN-320 (ours)}
		\centering
		{\includegraphics[width=1\linewidth,clip]{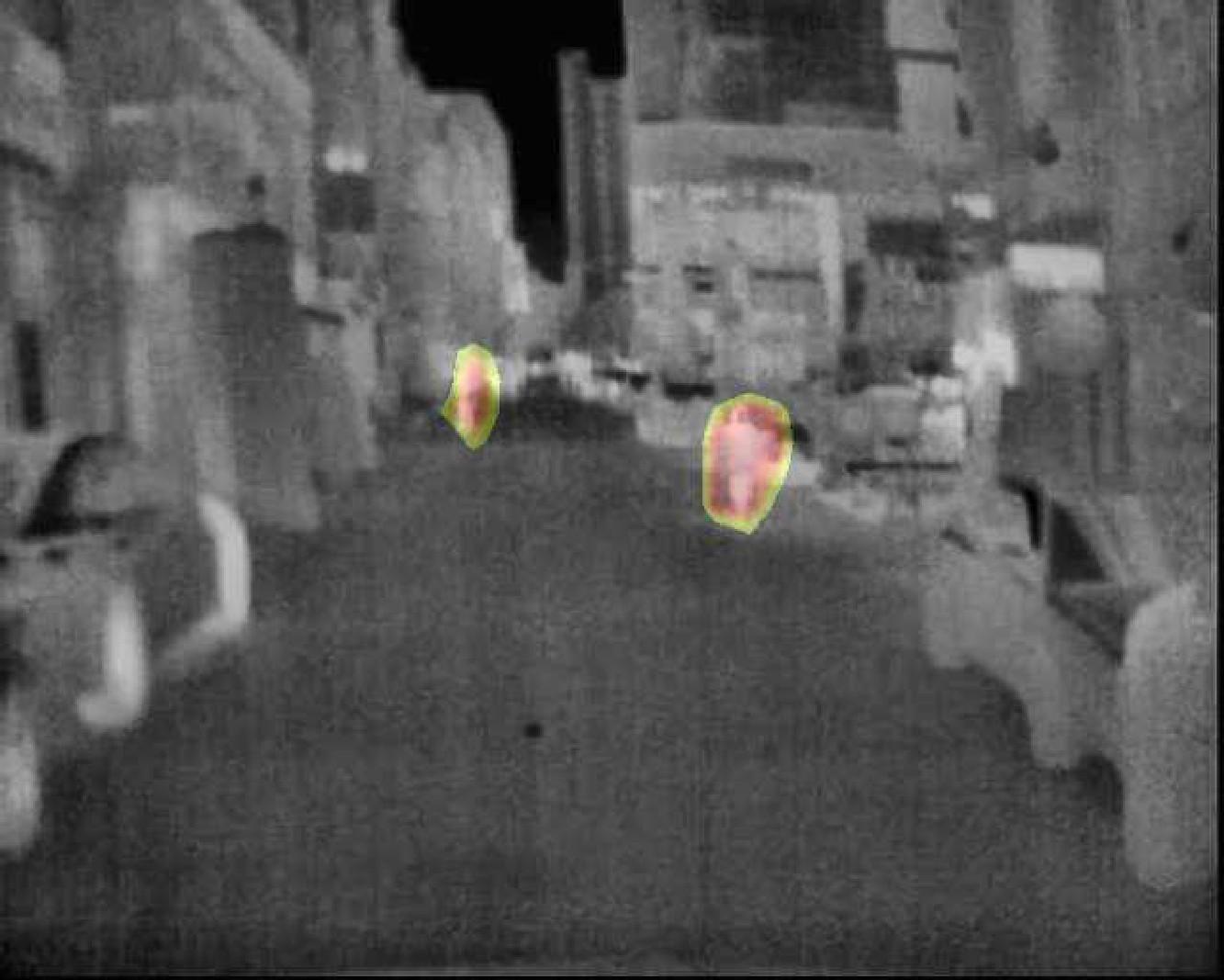}}
	\end{minipage}
	\vspace{1mm}\\
	\centering
	\begin{minipage}{0.16\linewidth}
		\centering
		{\includegraphics[width=1\linewidth,clip]{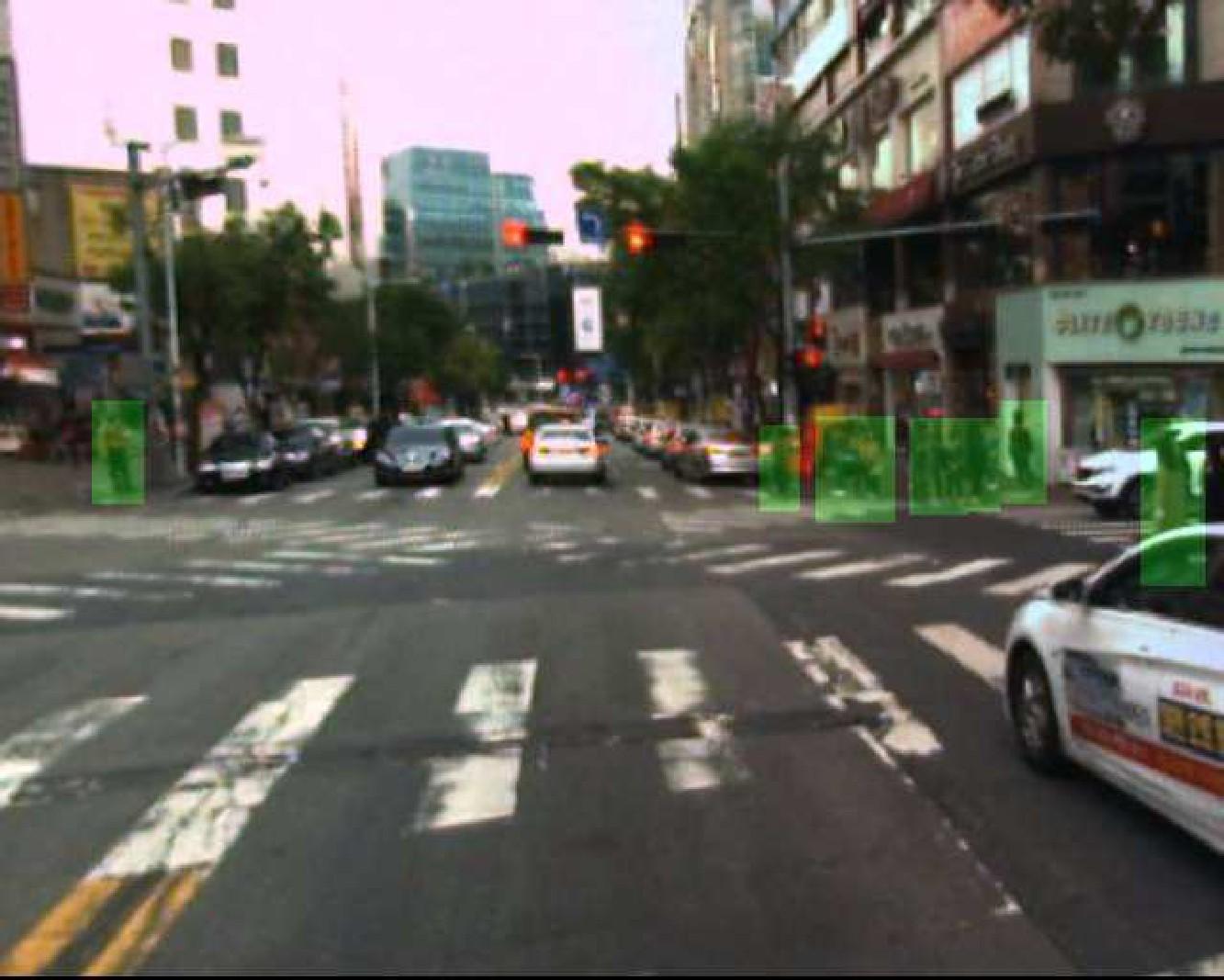}}
	\end{minipage}
	\begin{minipage}{0.16\linewidth}
		\centering
		{\includegraphics[width=1\linewidth,clip]{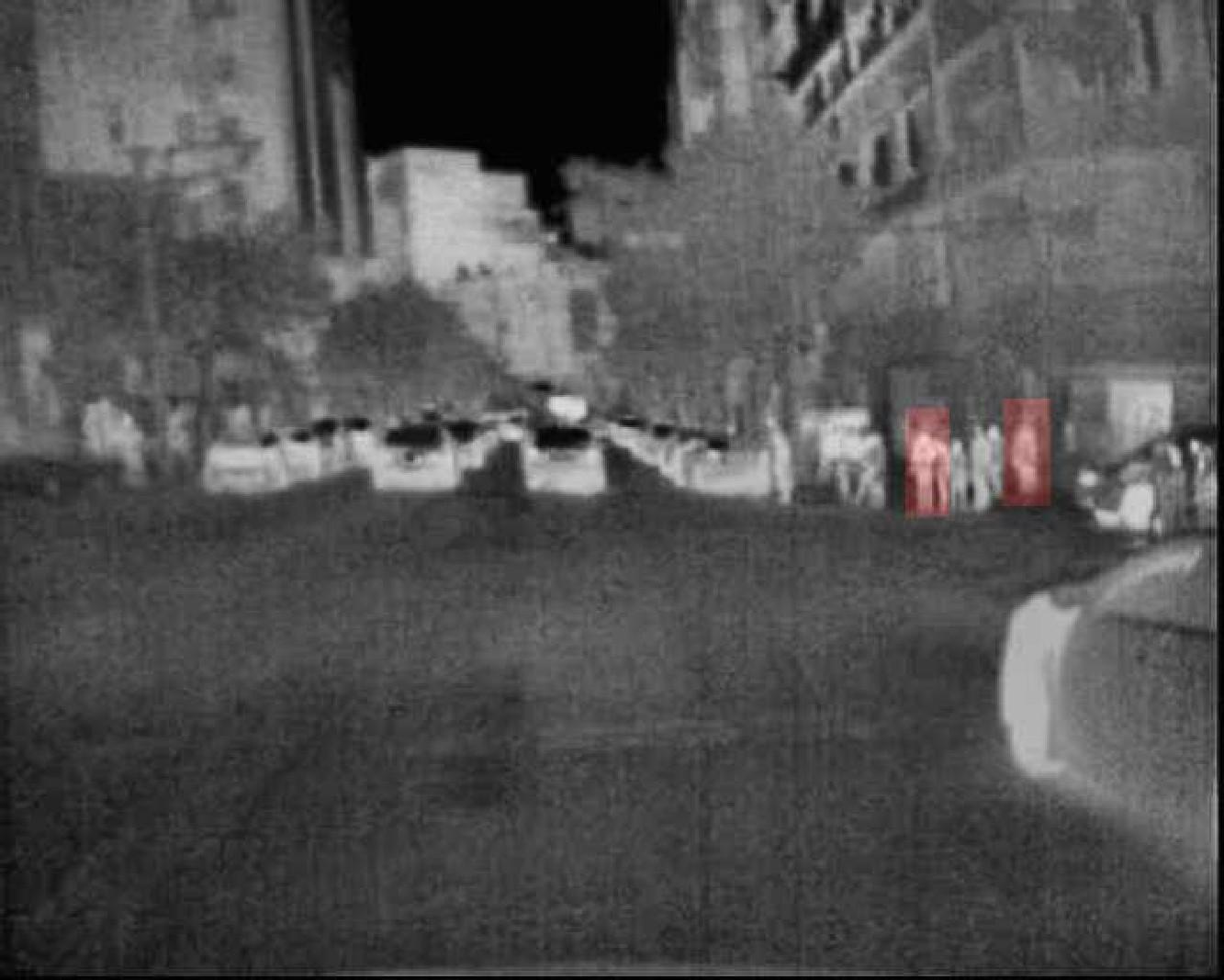}}
	\end{minipage}
	\begin{minipage}{0.16\linewidth}
		\centering
		{\includegraphics[width=1\linewidth,clip]{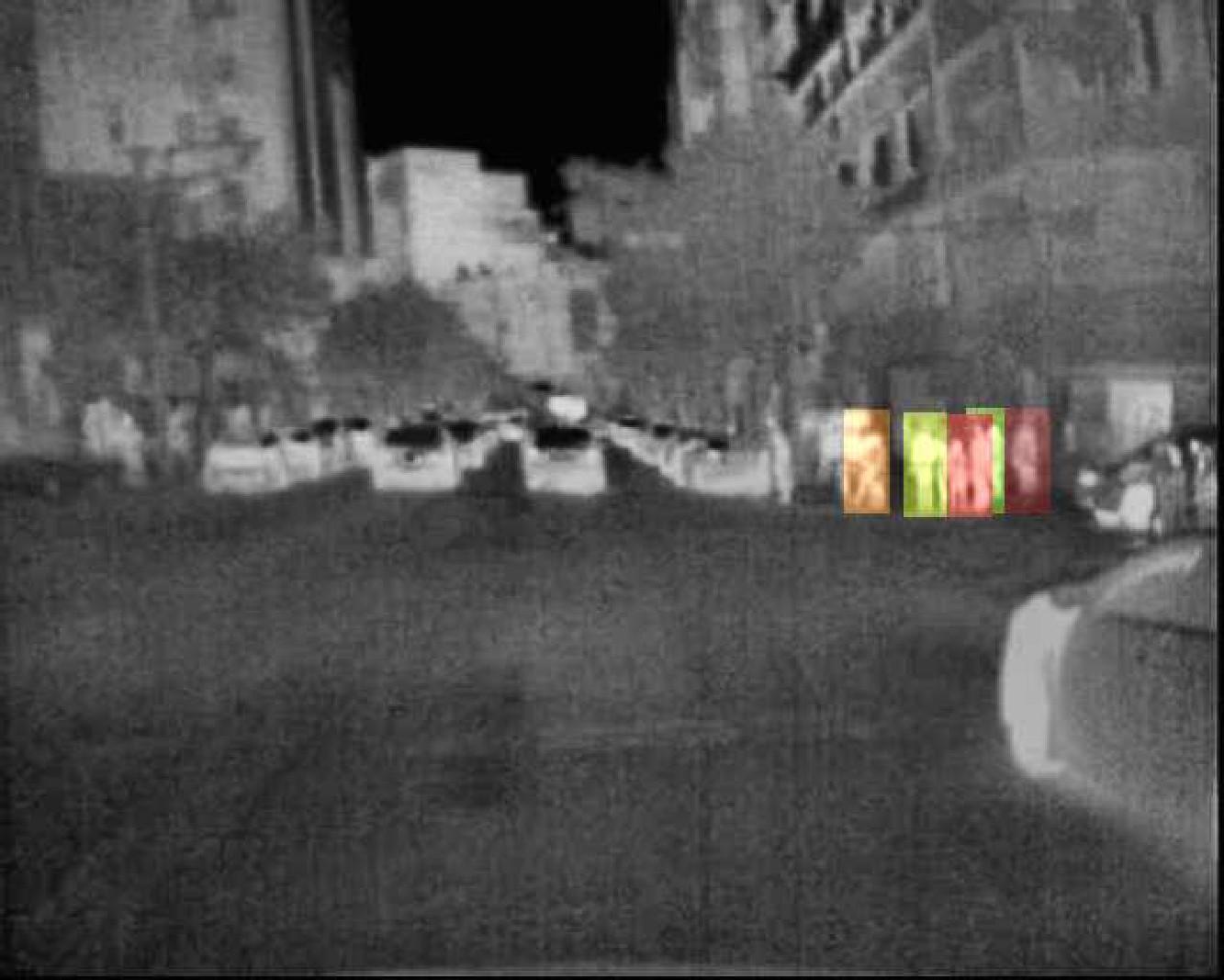}}
	\end{minipage}
	\begin{minipage}{0.16\linewidth}
		\centering
		{\includegraphics[width=1\linewidth,clip]{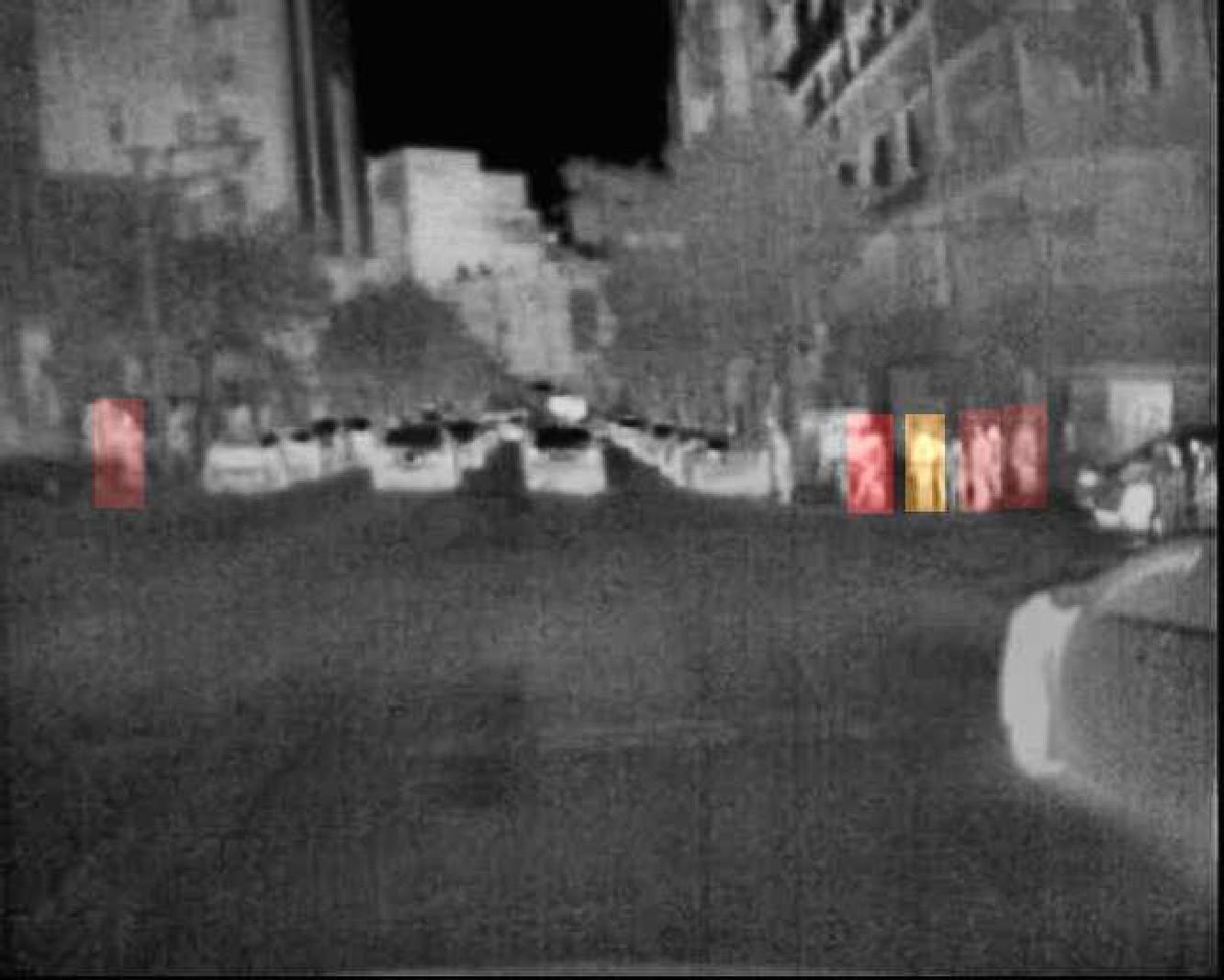}}
	\end{minipage} 	
	\begin{minipage}{0.16\linewidth}
		\centering
		{\includegraphics[width=1\linewidth,clip]{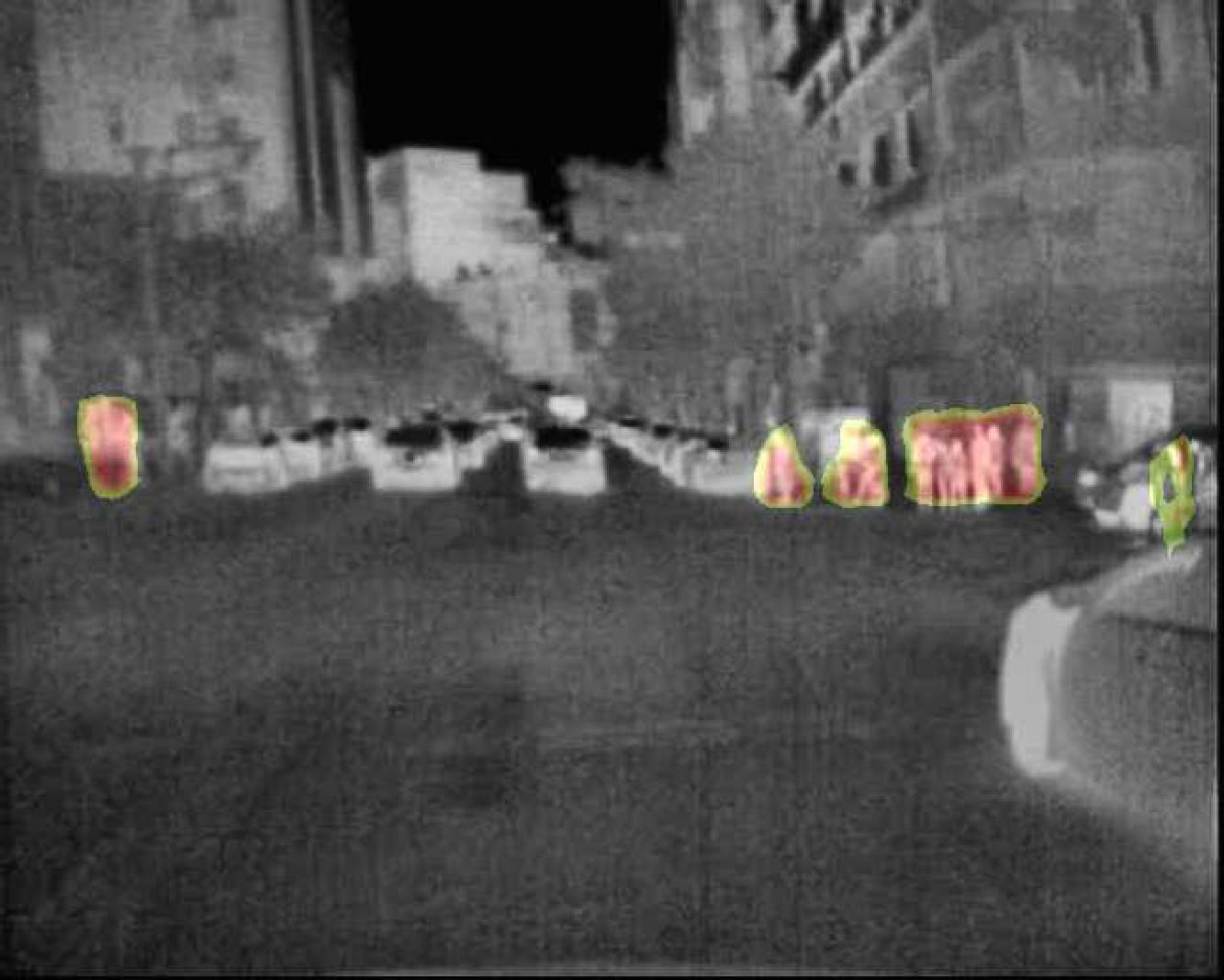}}
	\end{minipage} 
	\begin{minipage}{0.16\linewidth}
		\centering
		{\includegraphics[width=1\linewidth,clip]{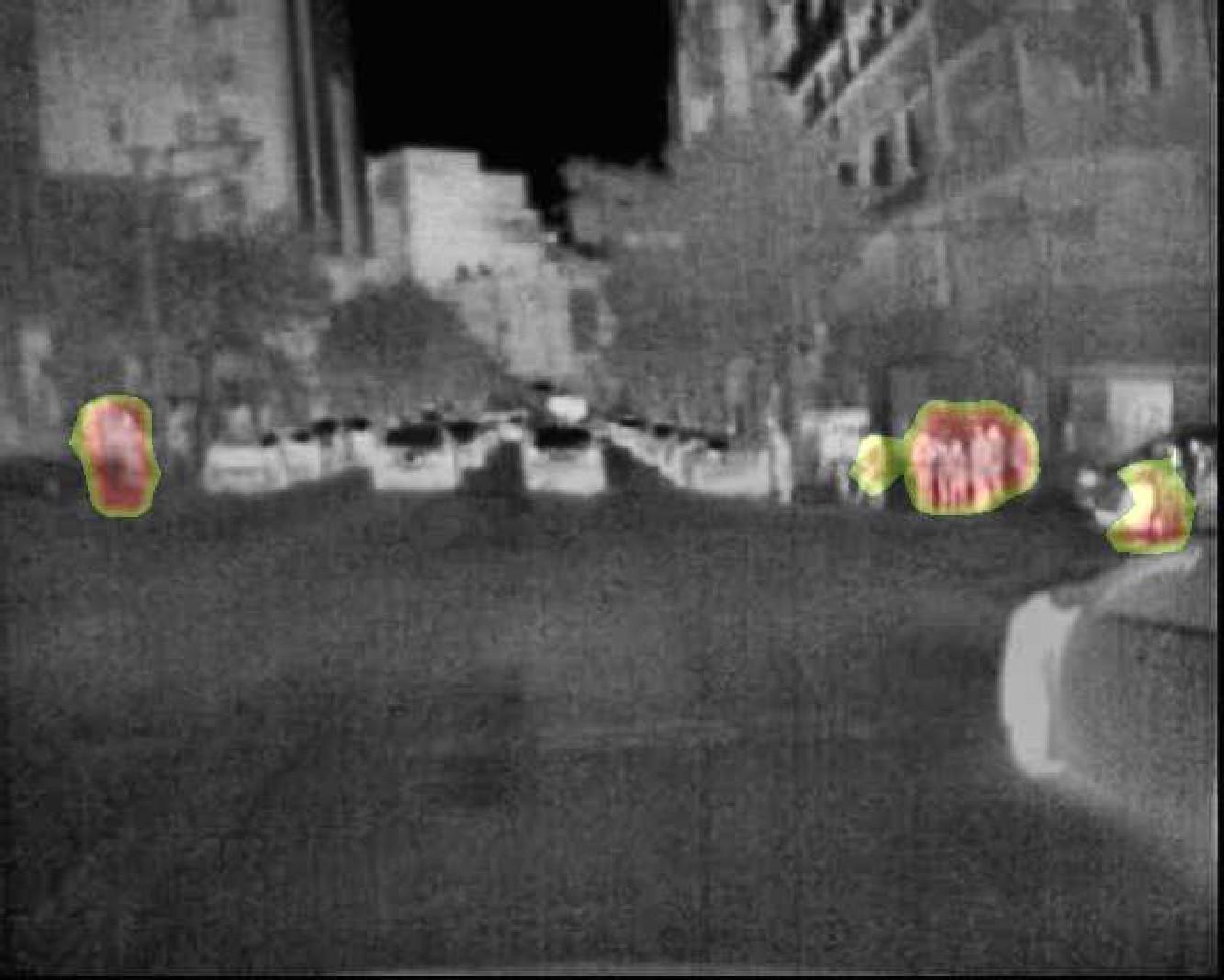}}
	\end{minipage} 
	\vspace{1mm}\\
	\begin{minipage}{0.16\linewidth}
		\centering
		{\includegraphics[width=1\linewidth,clip]{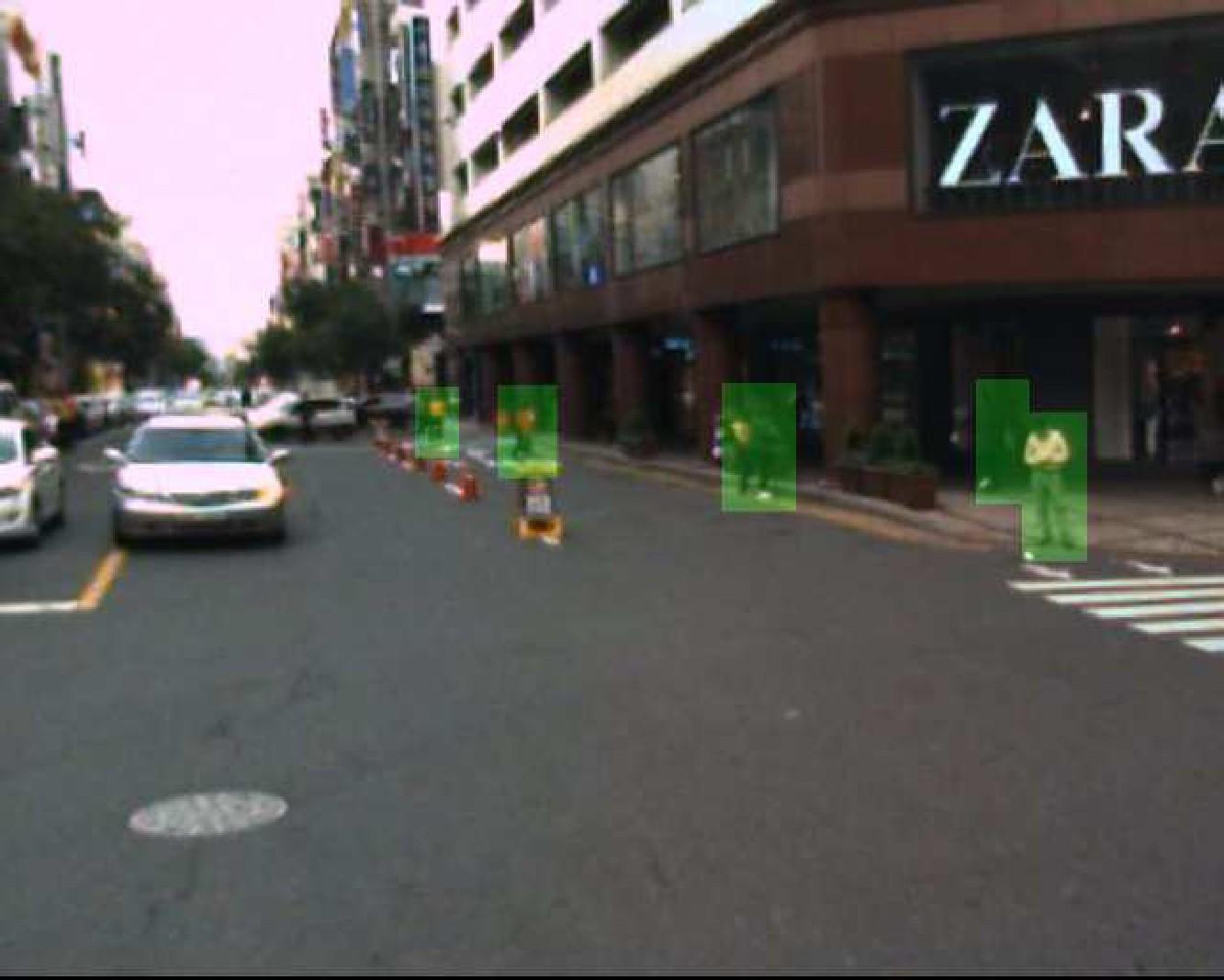}}
	\end{minipage}
	\begin{minipage}{0.16\linewidth}
		\centering
		{\includegraphics[width=1\linewidth,clip]{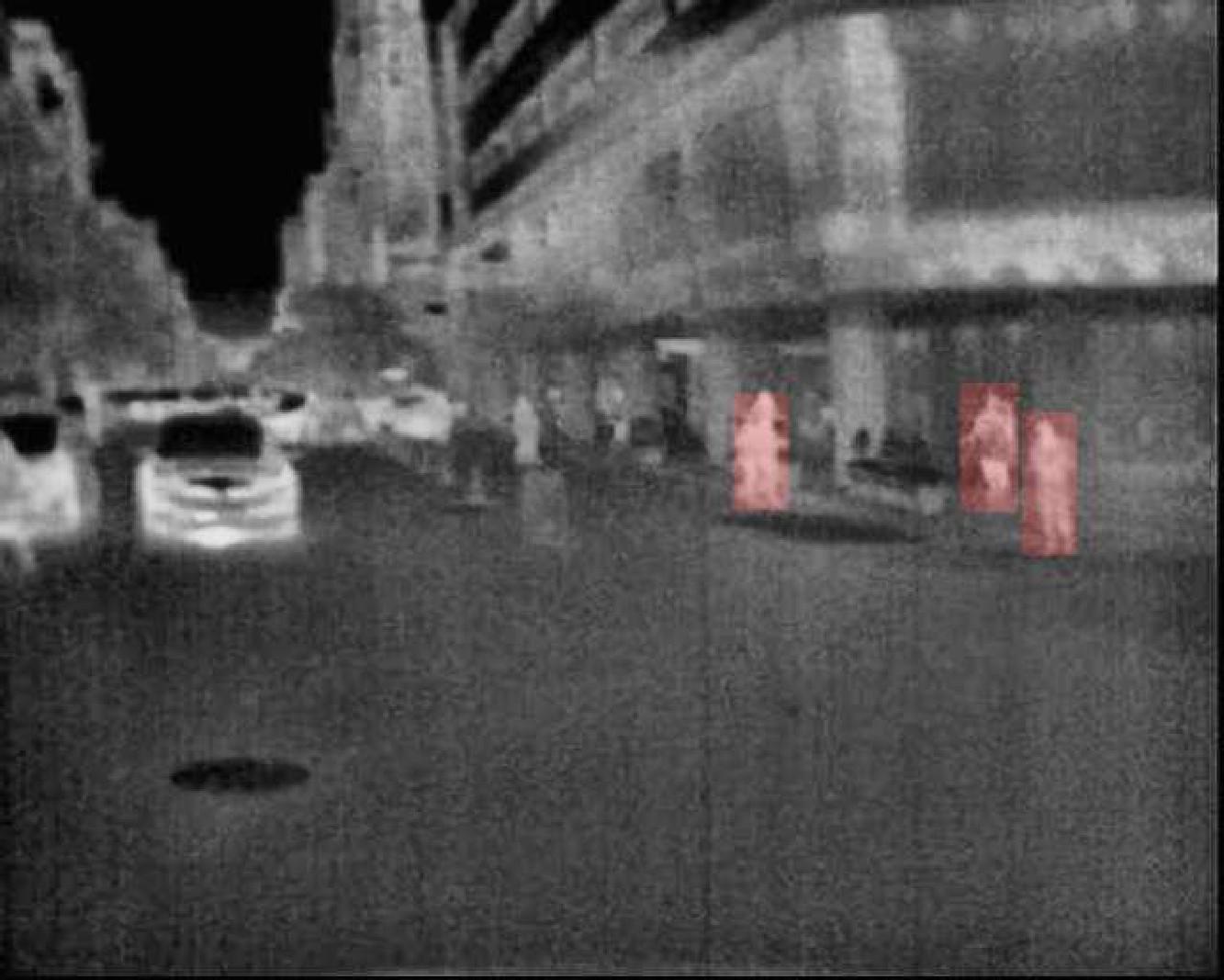}}
	\end{minipage}
	\begin{minipage}{0.16\linewidth}
		\centering
		{\includegraphics[width=1\linewidth,clip]{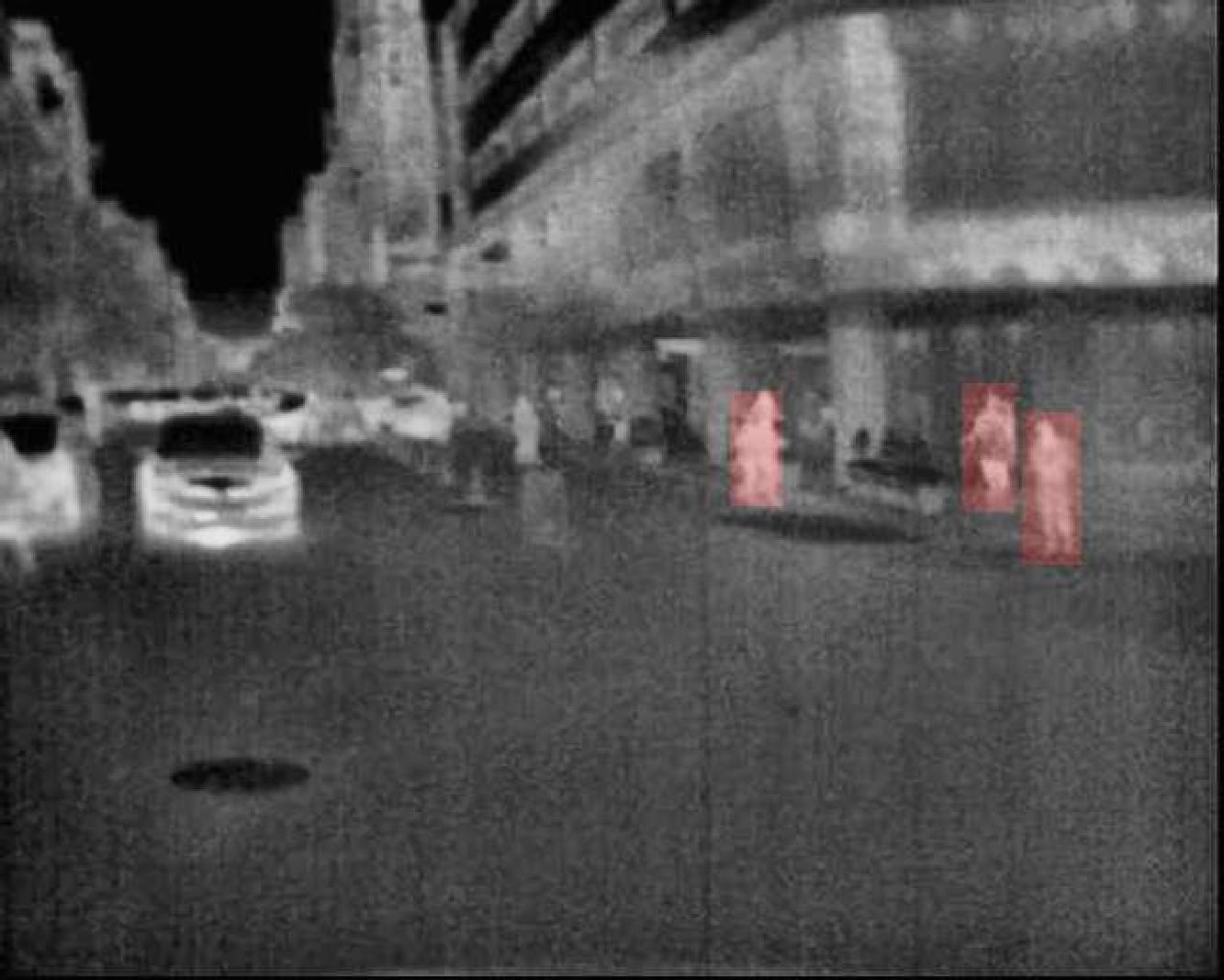}}
	\end{minipage}
	\begin{minipage}{0.16\linewidth}
		\centering
		{\includegraphics[width=1\linewidth,clip]{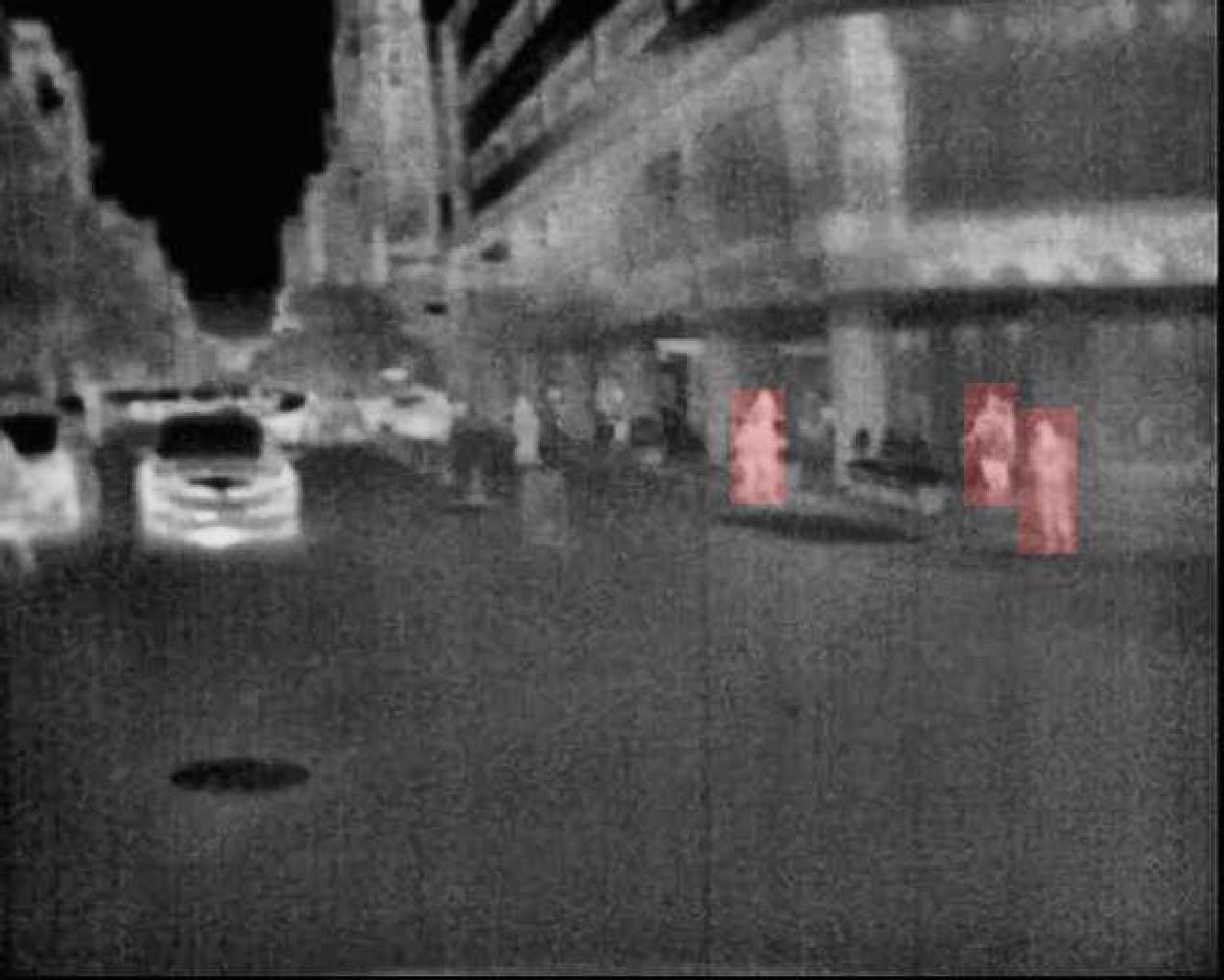}}
	\end{minipage} 	
	\begin{minipage}{0.16\linewidth}
		\centering
		{\includegraphics[width=1\linewidth,clip]{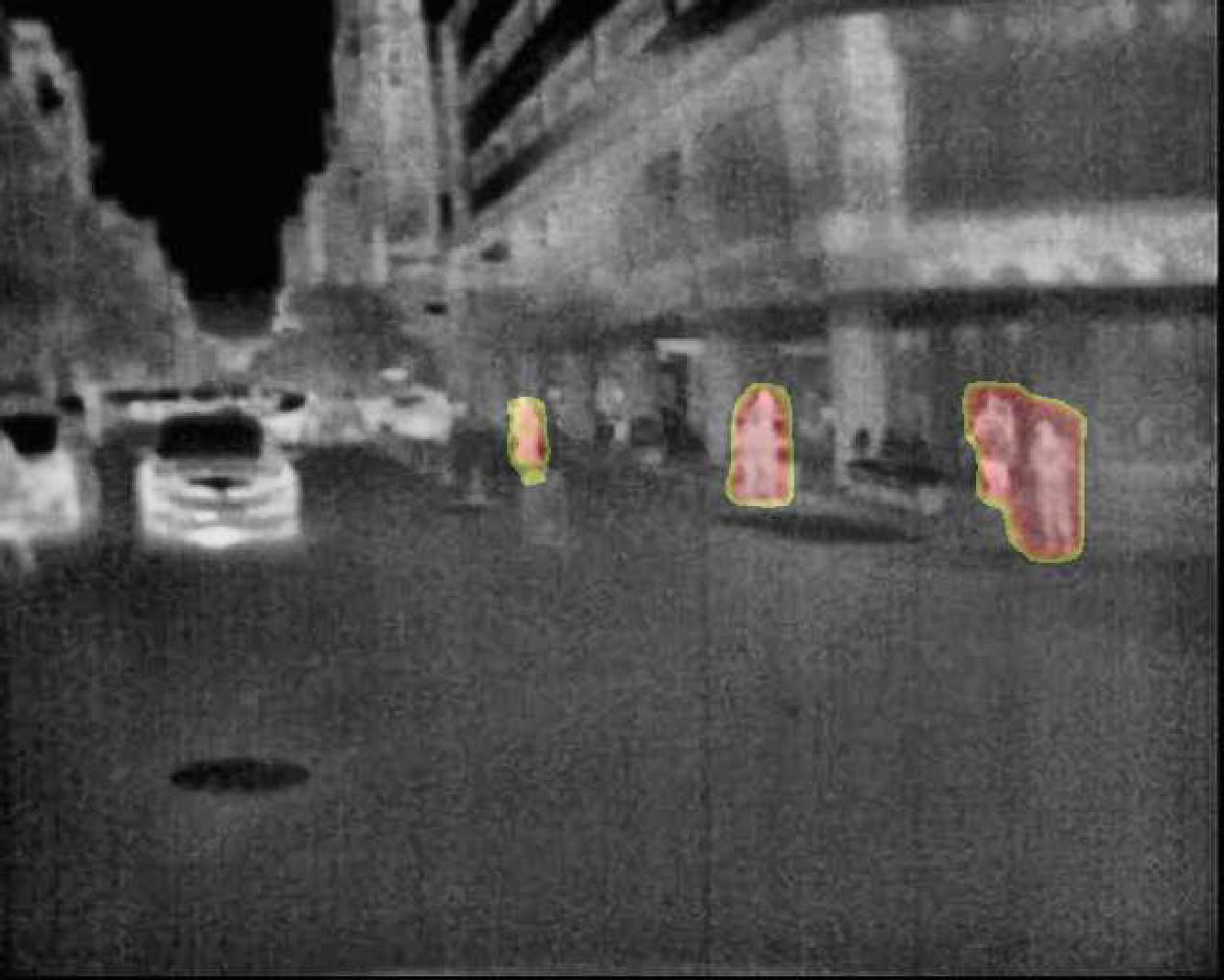}}
	\end{minipage} 
	\begin{minipage}{0.16\linewidth}
		\centering
		{\includegraphics[width=1\linewidth,clip]{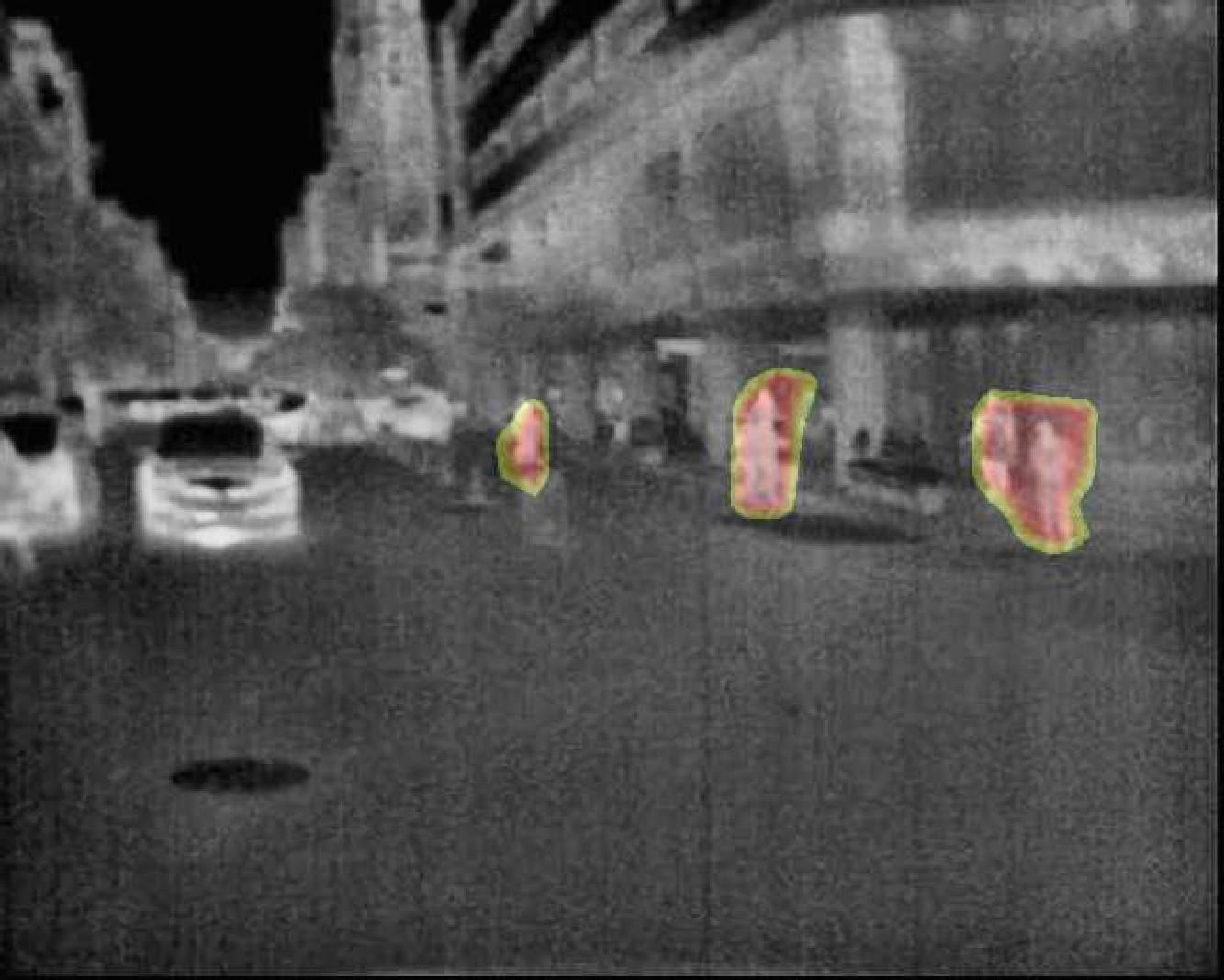}}
	\end{minipage} 
	\vspace{1mm}\\
		{\bf (a) Daytime}\\
	\centering
	\begin{minipage}{0.16\linewidth}
		\centering
		{\includegraphics[width=1\linewidth,clip]{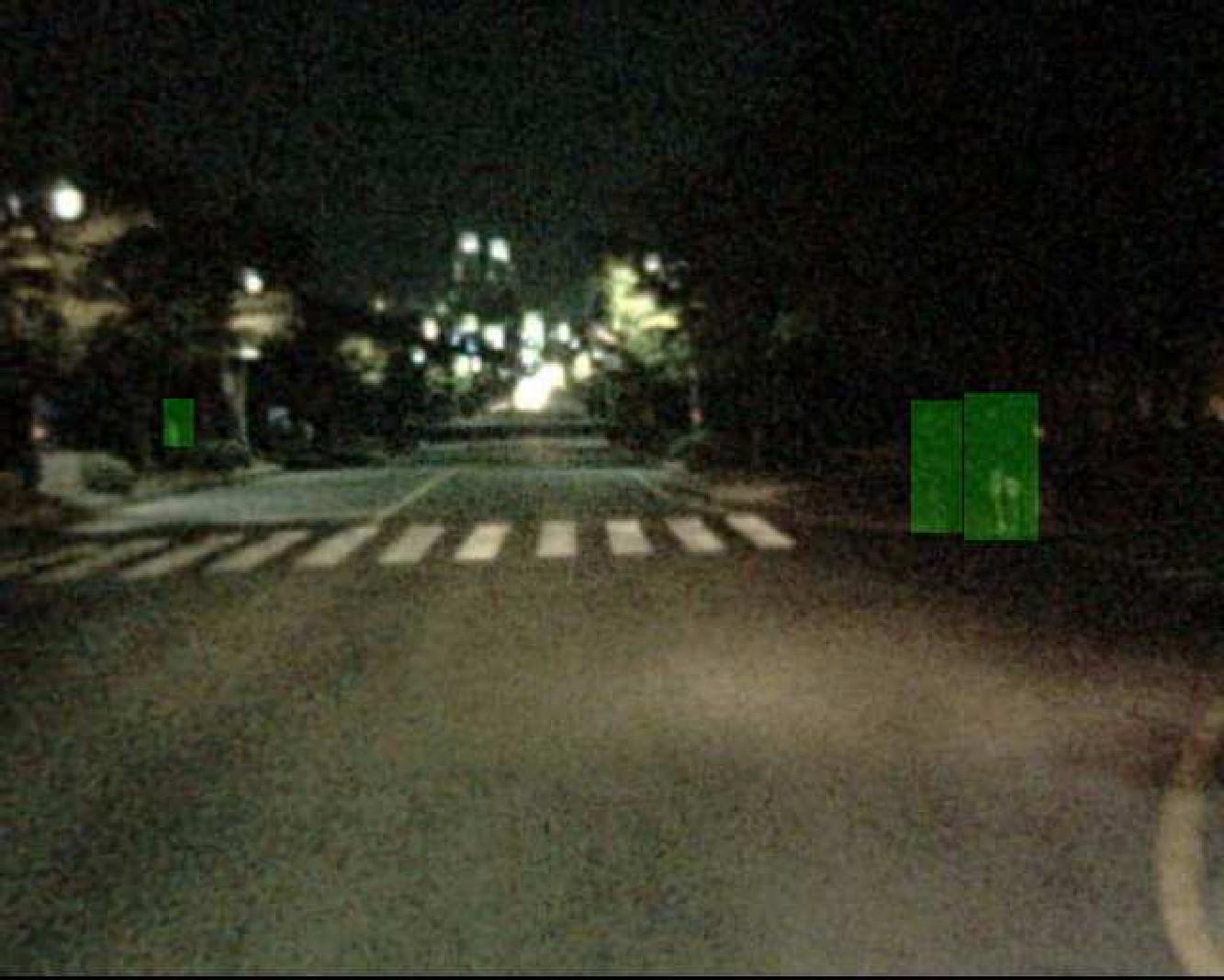}}
	\end{minipage}
	\begin{minipage}{0.16\linewidth}
		\centering
		{\includegraphics[width=1\linewidth,clip]{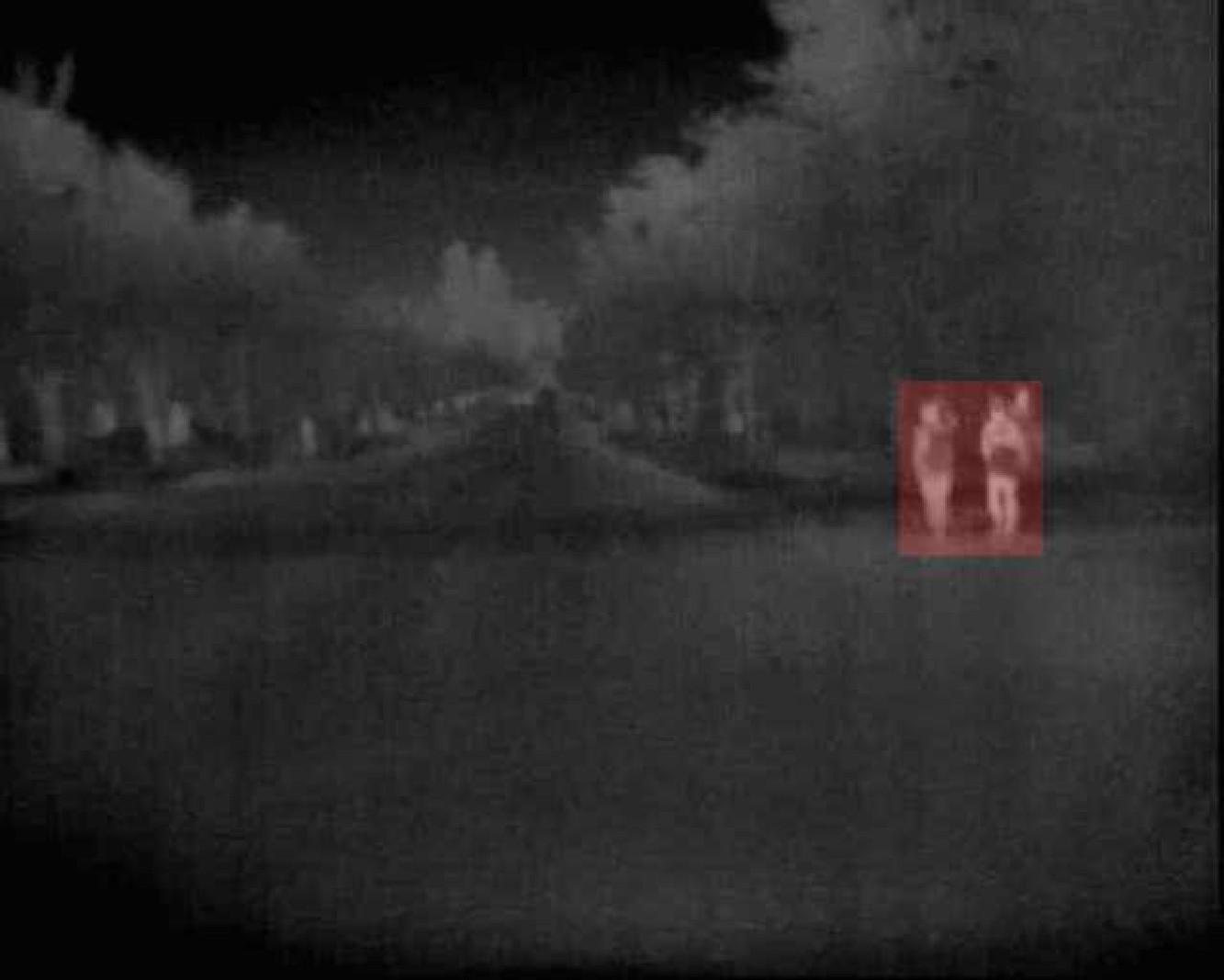}}
	\end{minipage}
	\begin{minipage}{0.16\linewidth}
		\centering
		{\includegraphics[width=1\linewidth,clip]{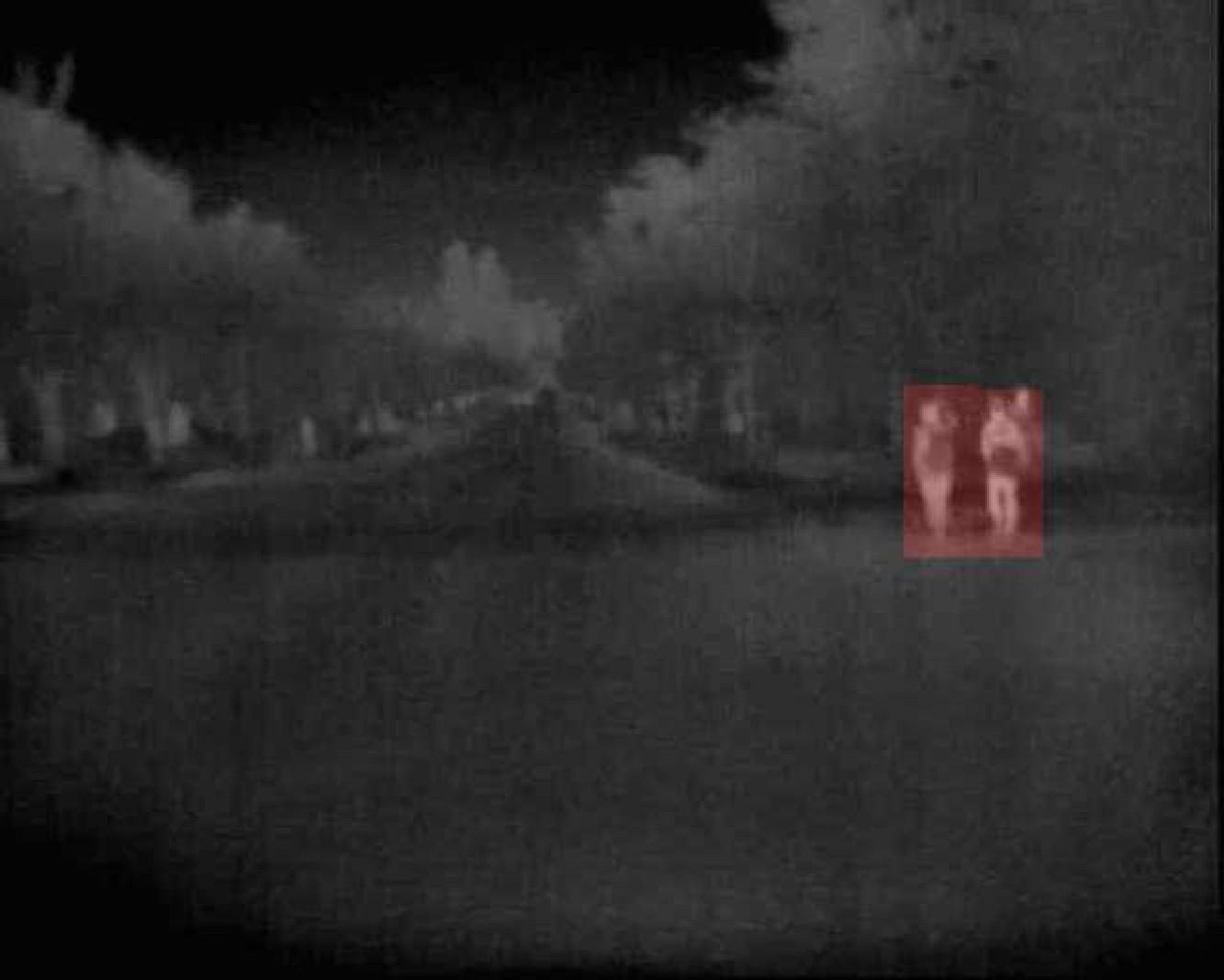}}
	\end{minipage}
	\begin{minipage}{0.16\linewidth}
		\centering
		{\includegraphics[width=1\linewidth,clip]{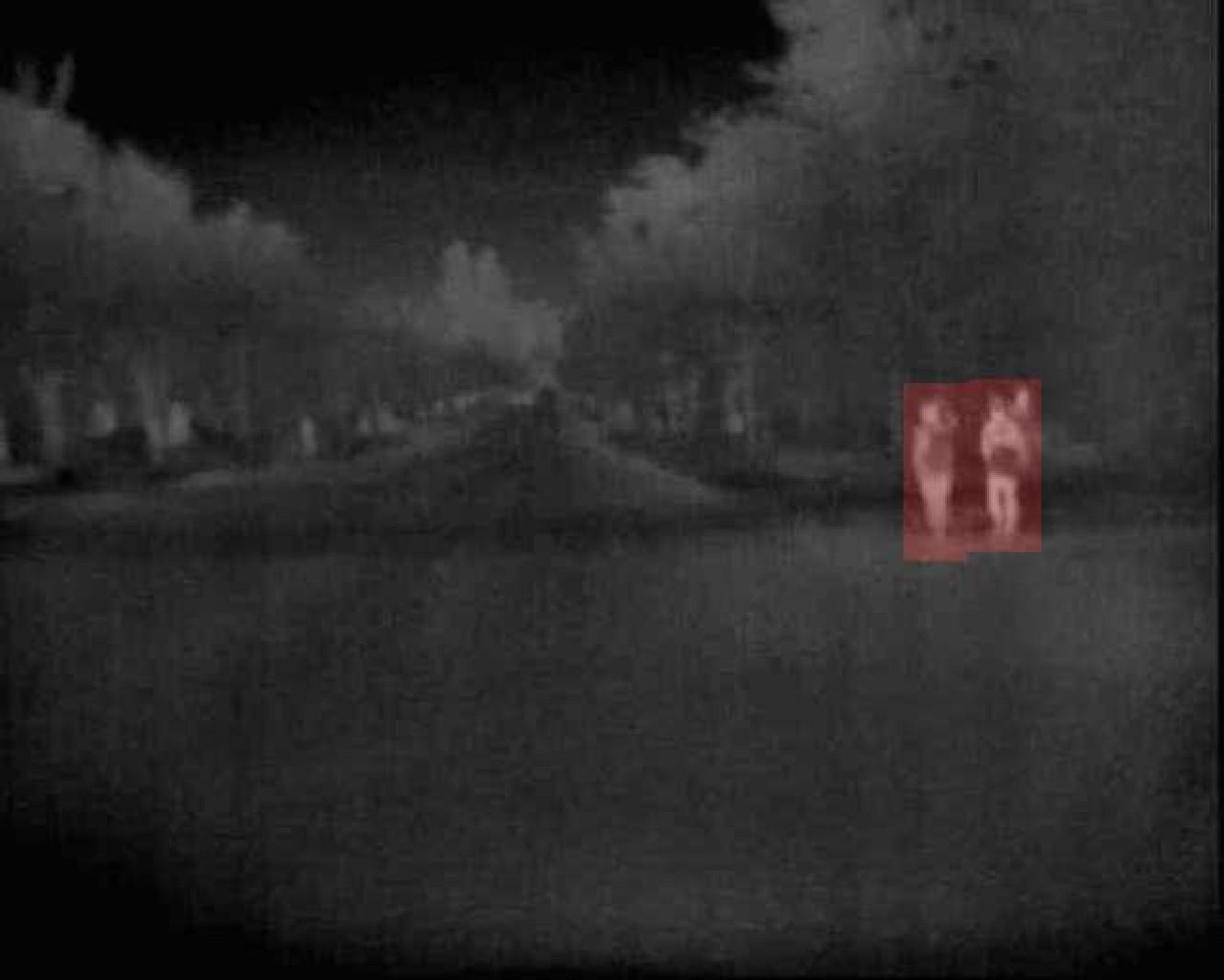}}
	\end{minipage} 	
	\begin{minipage}{0.16\linewidth}
		\centering
		{\includegraphics[width=1\linewidth,clip]{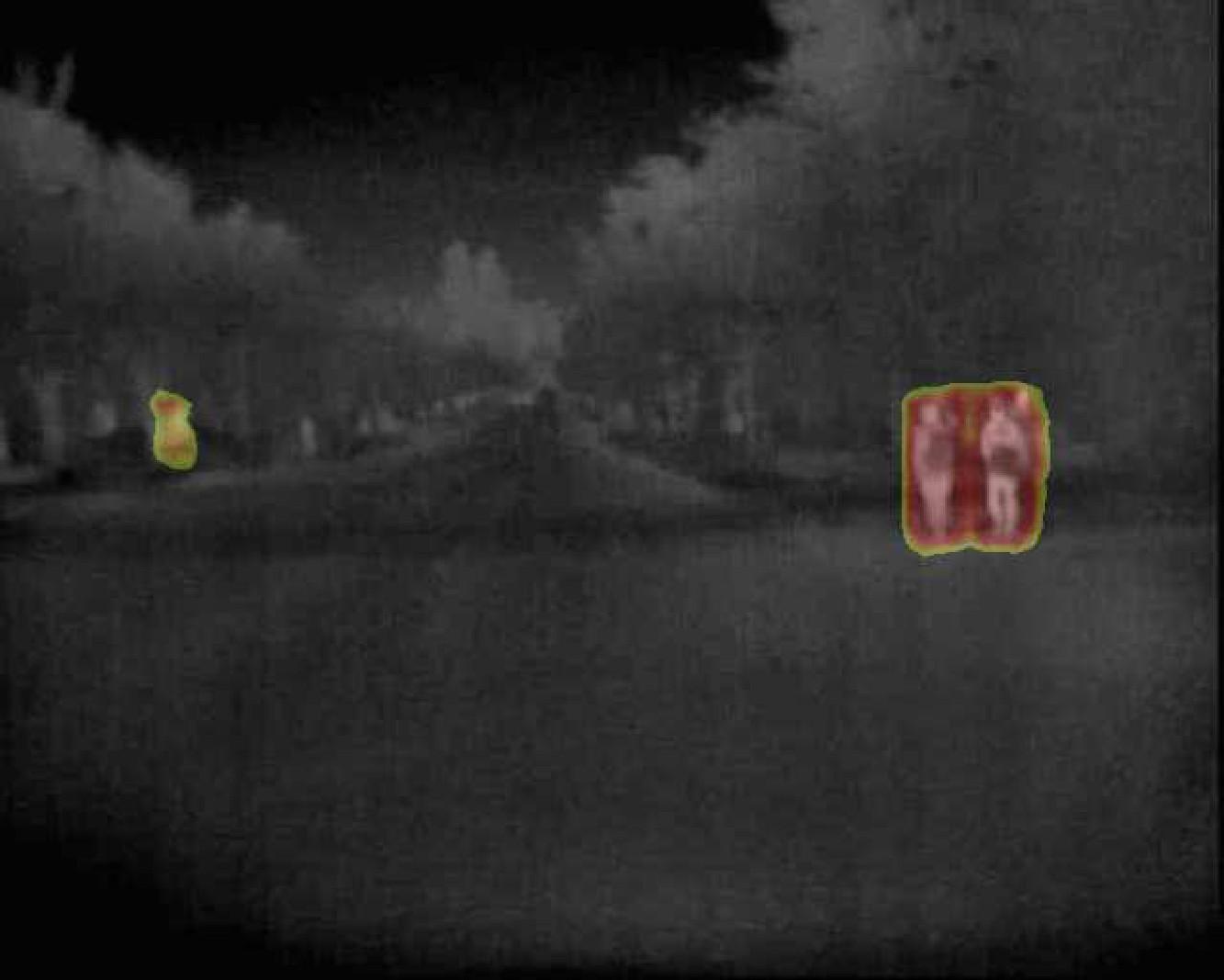}}
	\end{minipage} 
	\begin{minipage}{0.16\linewidth}
		\centering
		{\includegraphics[width=1\linewidth,clip]{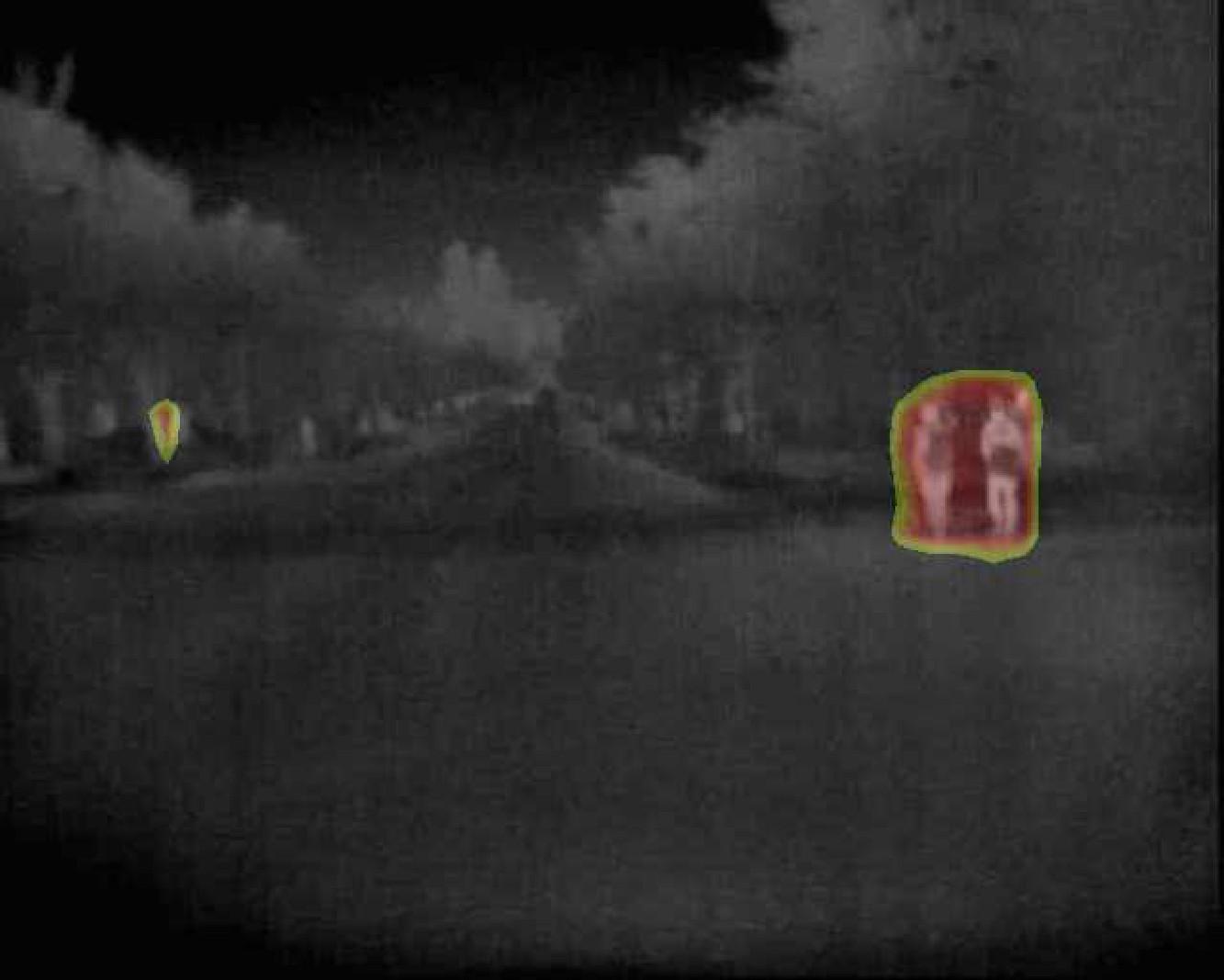}}
	\end{minipage} 
	\vspace{1mm}\\ 
	\centering
	\begin{minipage}{0.16\linewidth}
		\centering
		{\includegraphics[width=1\linewidth,clip]{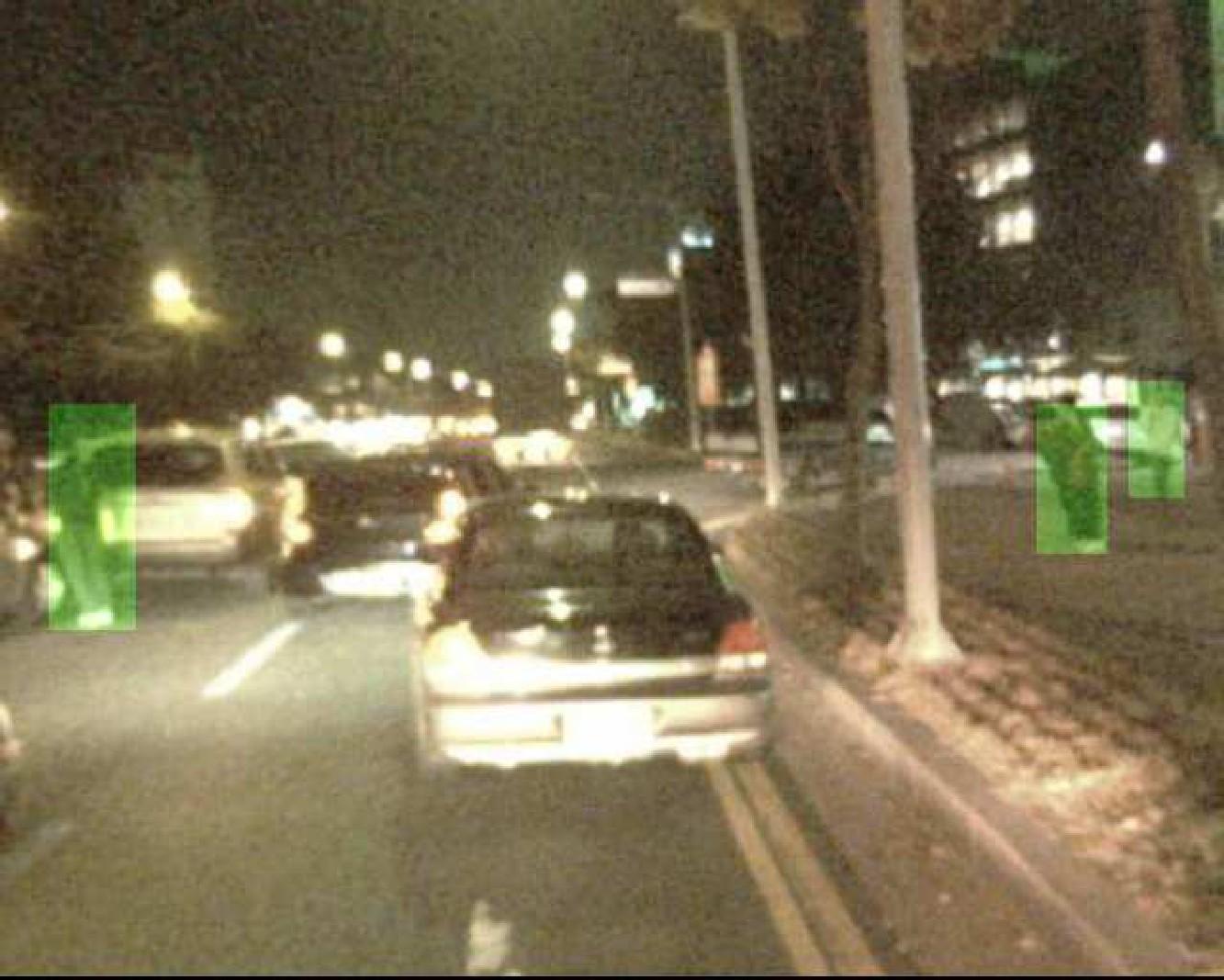}}
	\end{minipage}
	\begin{minipage}{0.16\linewidth}
		\centering
		{\includegraphics[width=1\linewidth,clip]{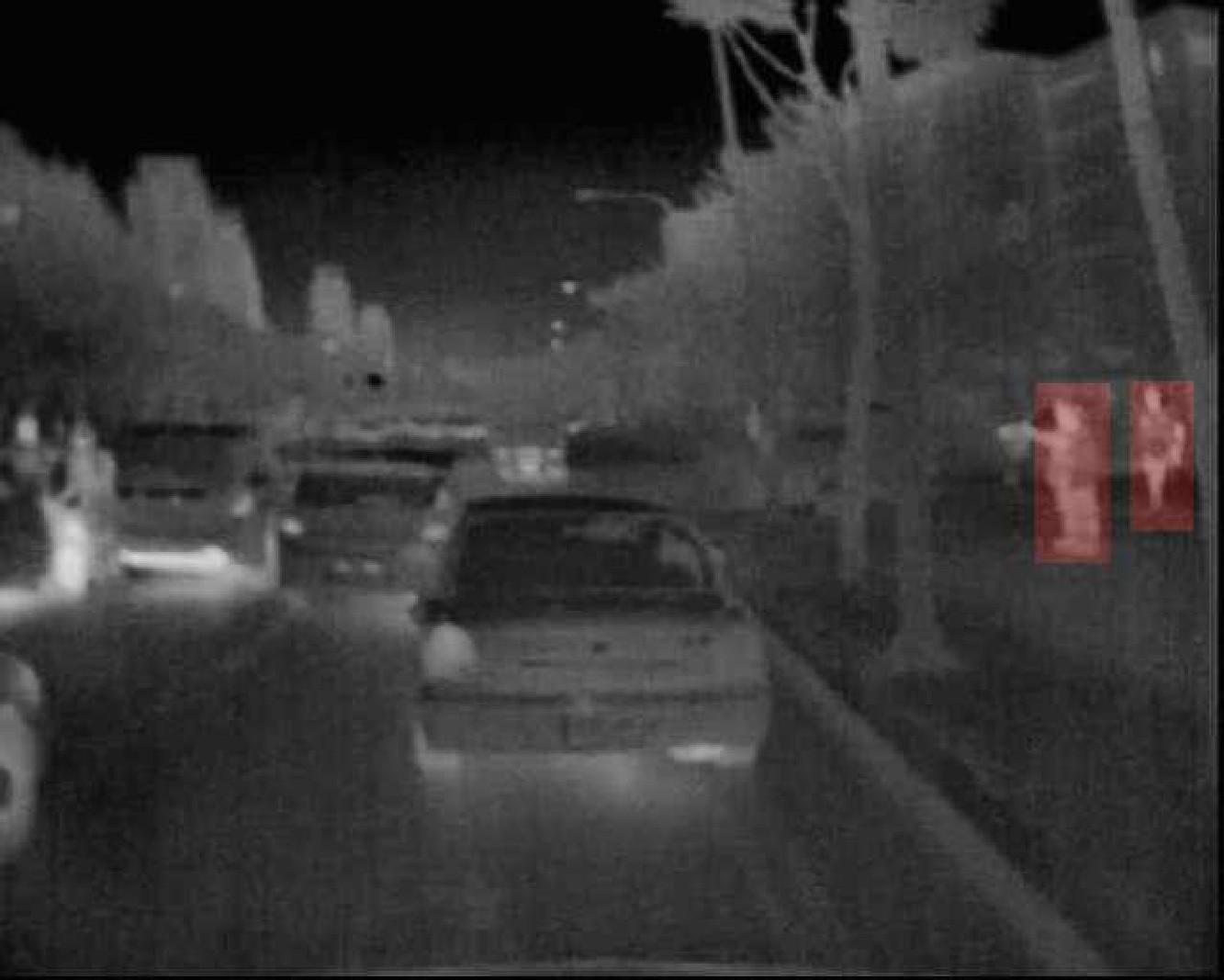}}
	\end{minipage}
	\begin{minipage}{0.16\linewidth}
		\centering
		{\includegraphics[width=1\linewidth,clip]{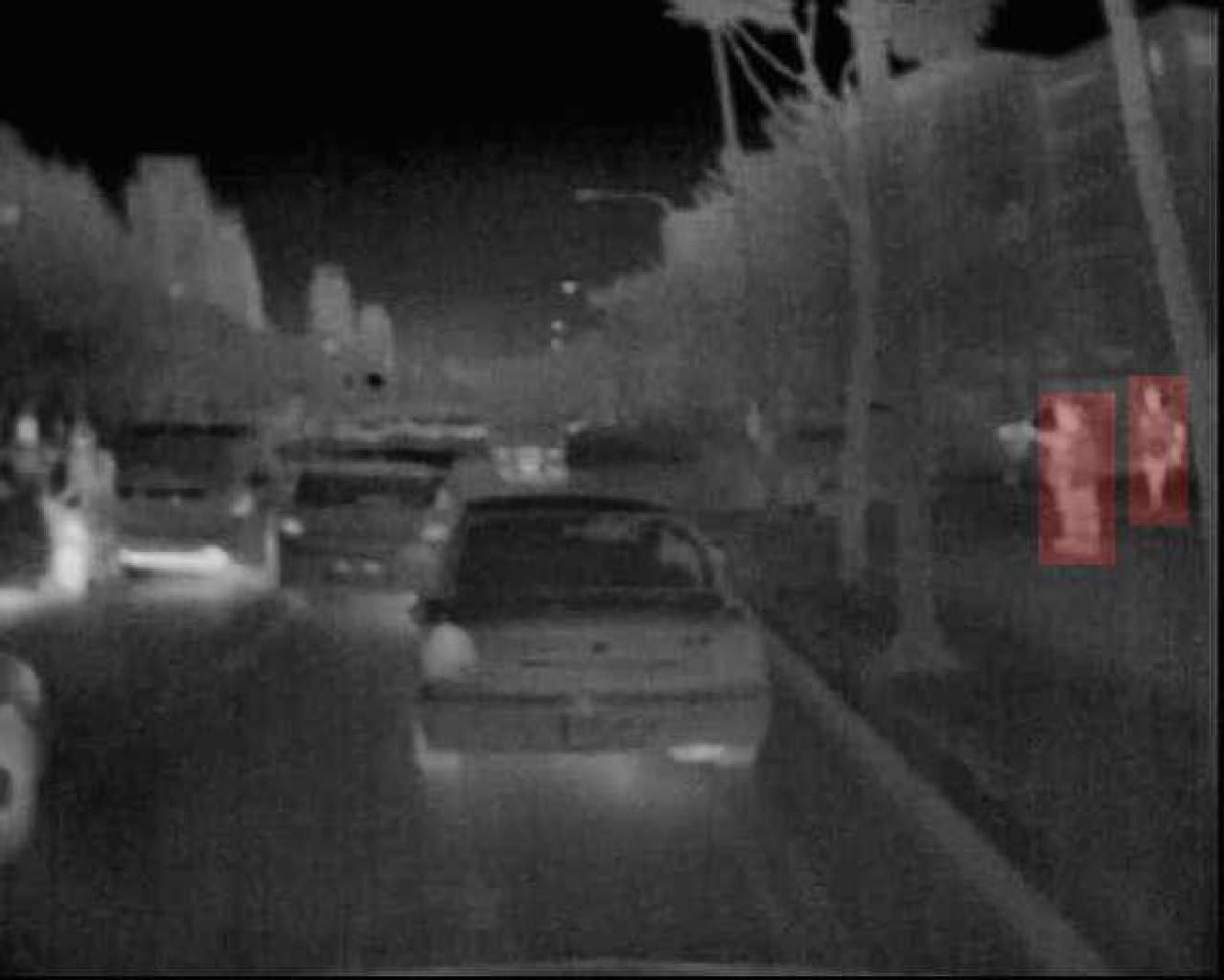}}
	\end{minipage}
	\begin{minipage}{0.16\linewidth}
		\centering
		{\includegraphics[width=1\linewidth,clip]{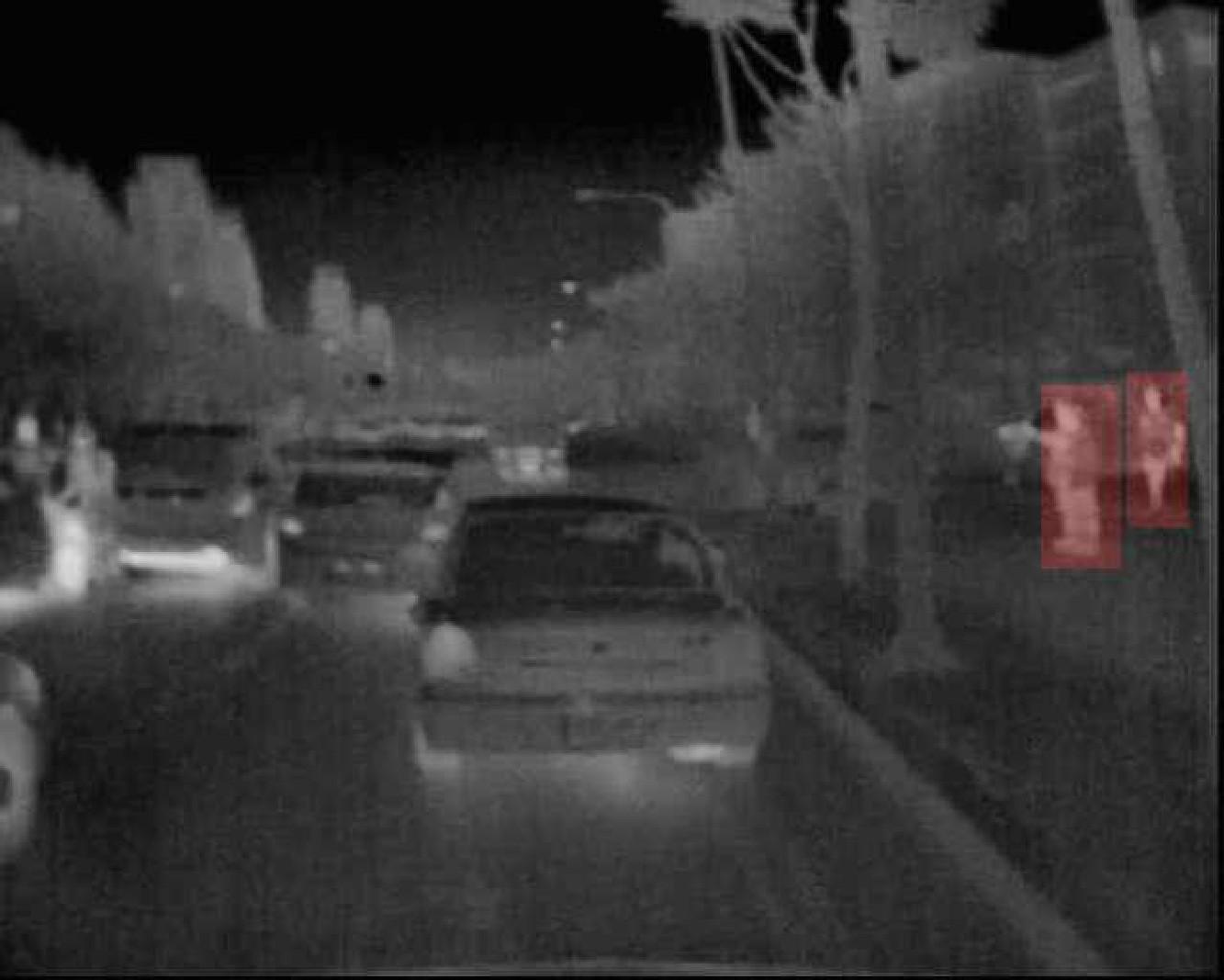}}
	\end{minipage} 	
	\begin{minipage}{0.16\linewidth}
		\centering
		{\includegraphics[width=1\linewidth,clip]{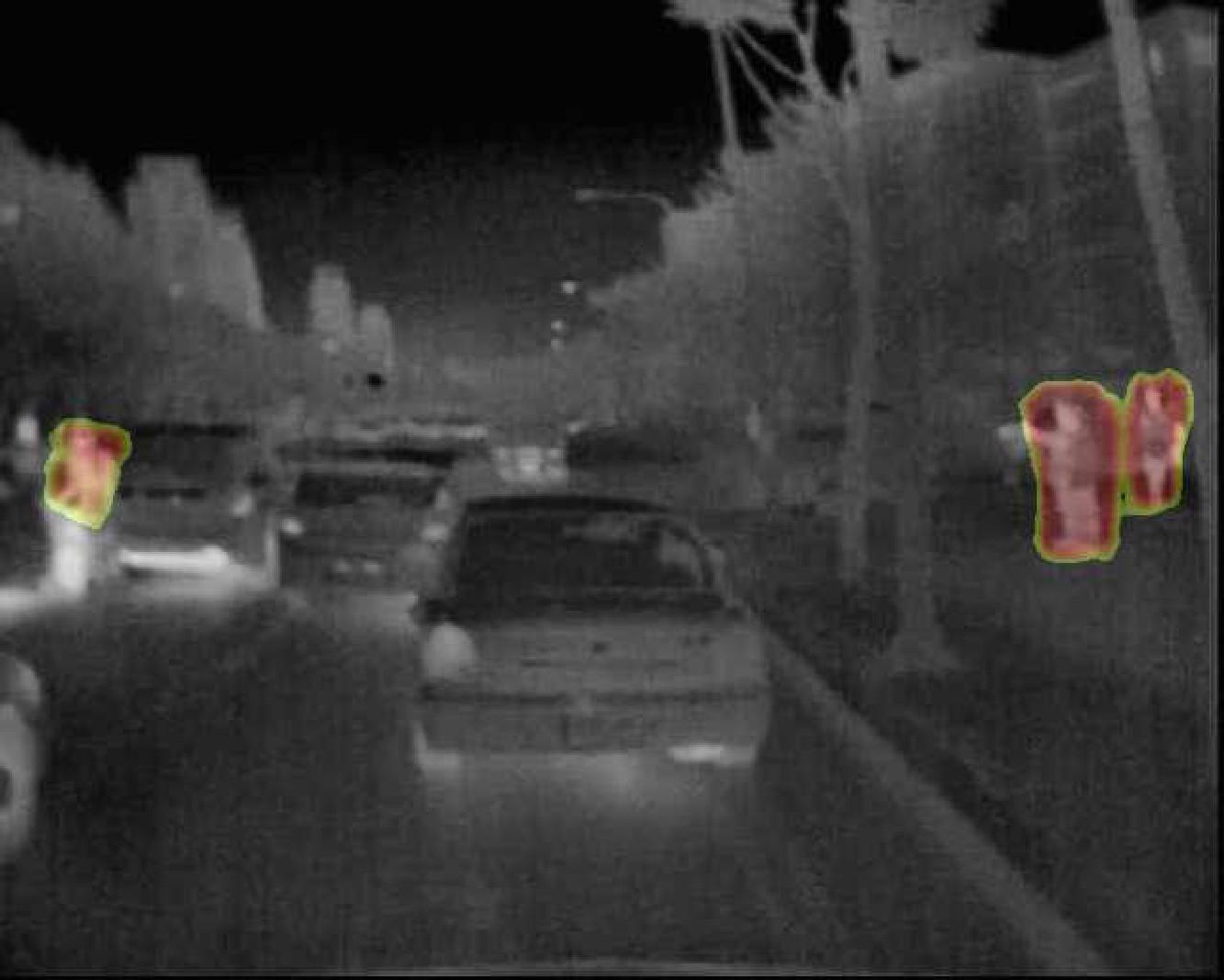}}
	\end{minipage} 
	\begin{minipage}{0.16\linewidth}
		\centering
		{\includegraphics[width=1\linewidth,clip]{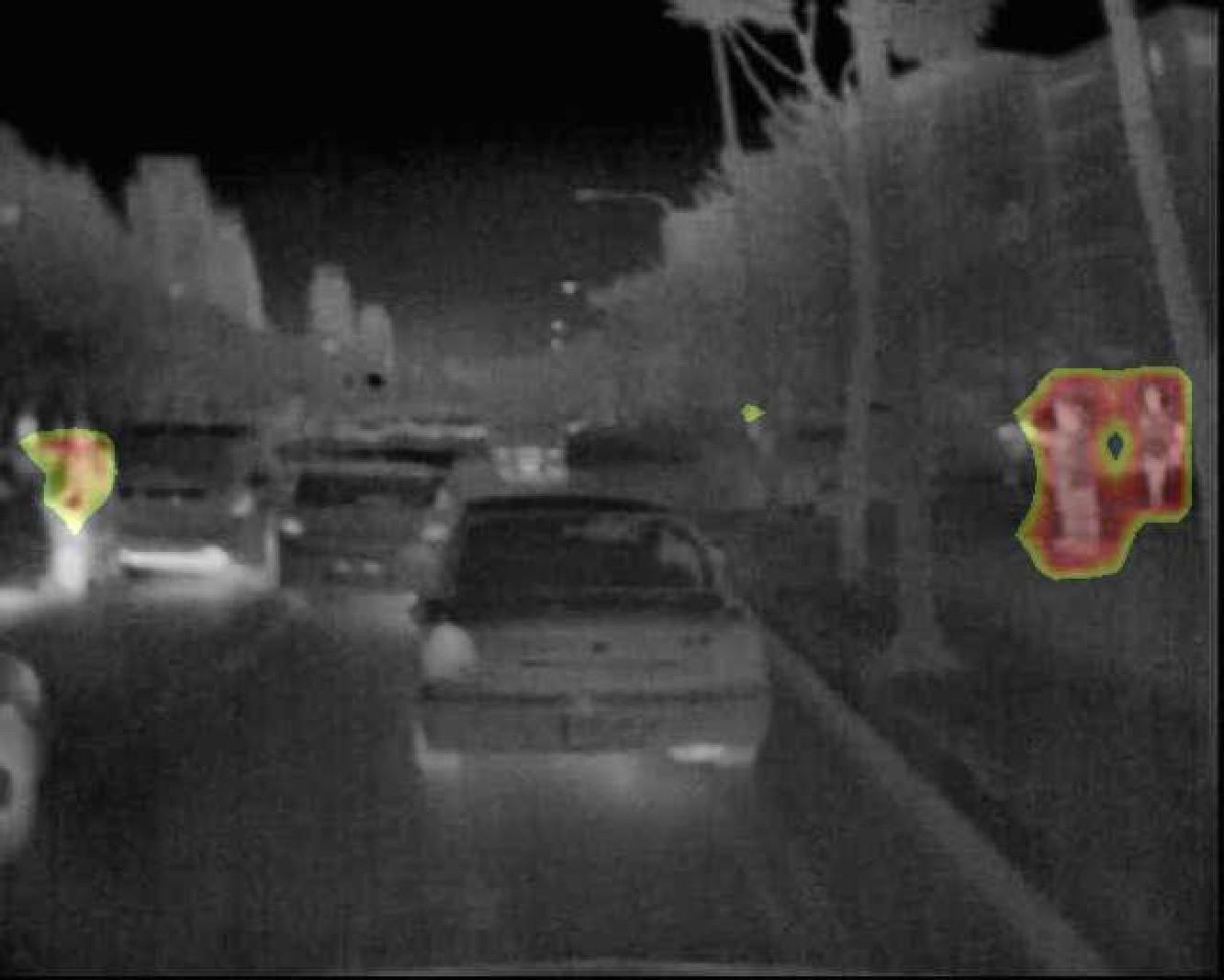}}
	\end{minipage} 
	\vspace{1mm}\\
	\centering
	\begin{minipage}{0.16\linewidth}
		\centering
		{\includegraphics[width=1\linewidth,clip]{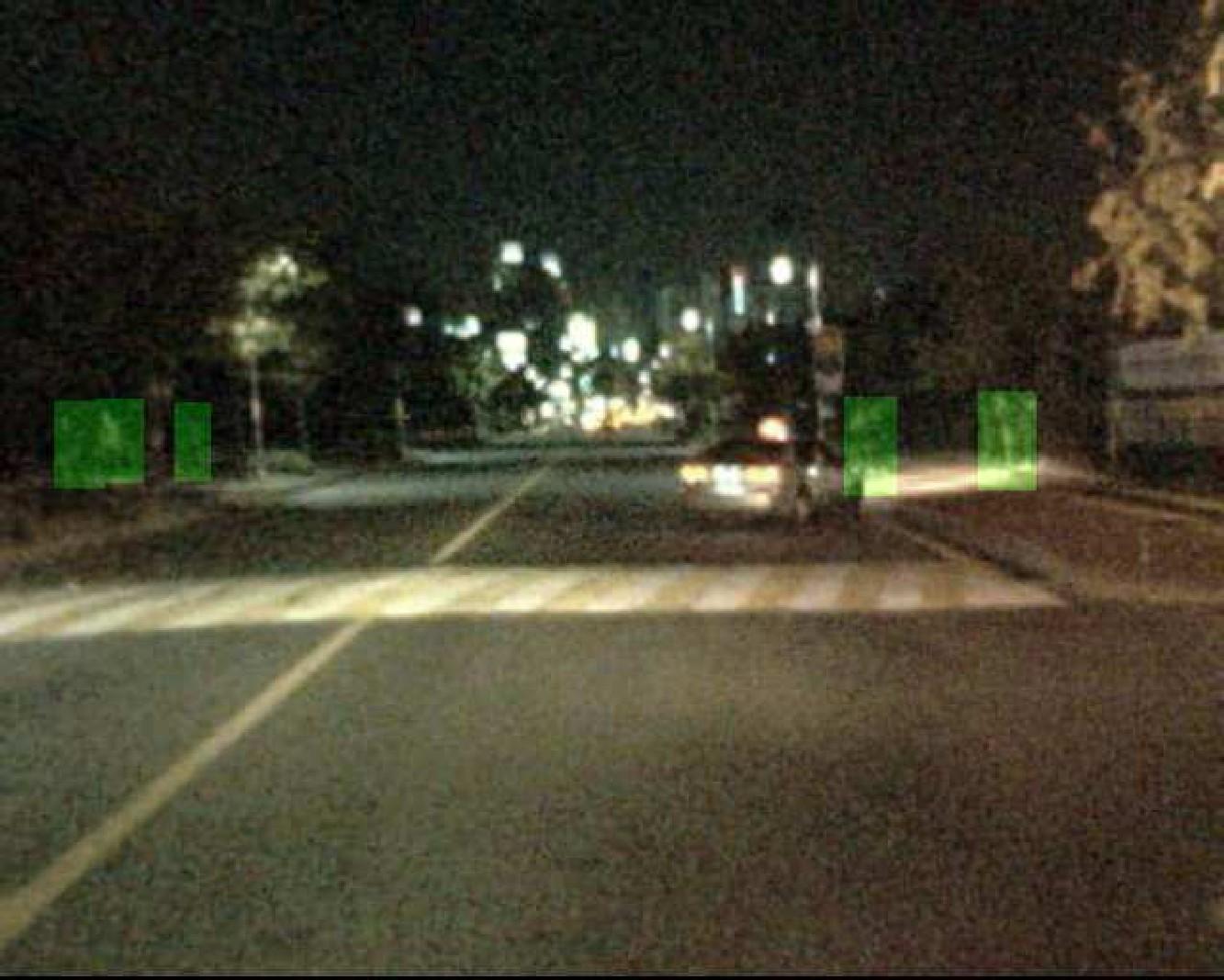}}
	\end{minipage}
	\begin{minipage}{0.16\linewidth}
		\centering
		{\includegraphics[width=1\linewidth,clip]{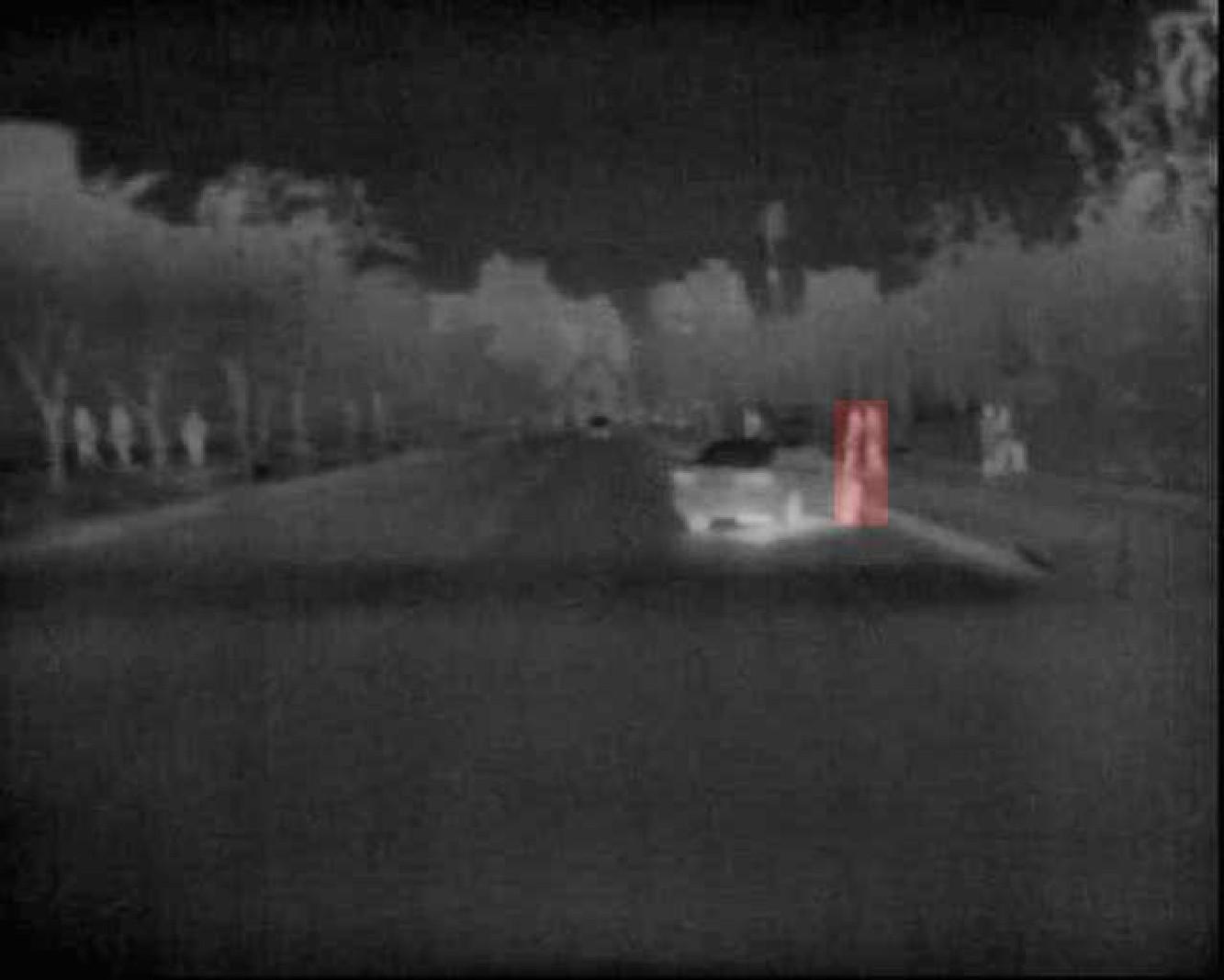}}
	\end{minipage}
	\begin{minipage}{0.16\linewidth}
		\centering
		{\includegraphics[width=1\linewidth,clip]{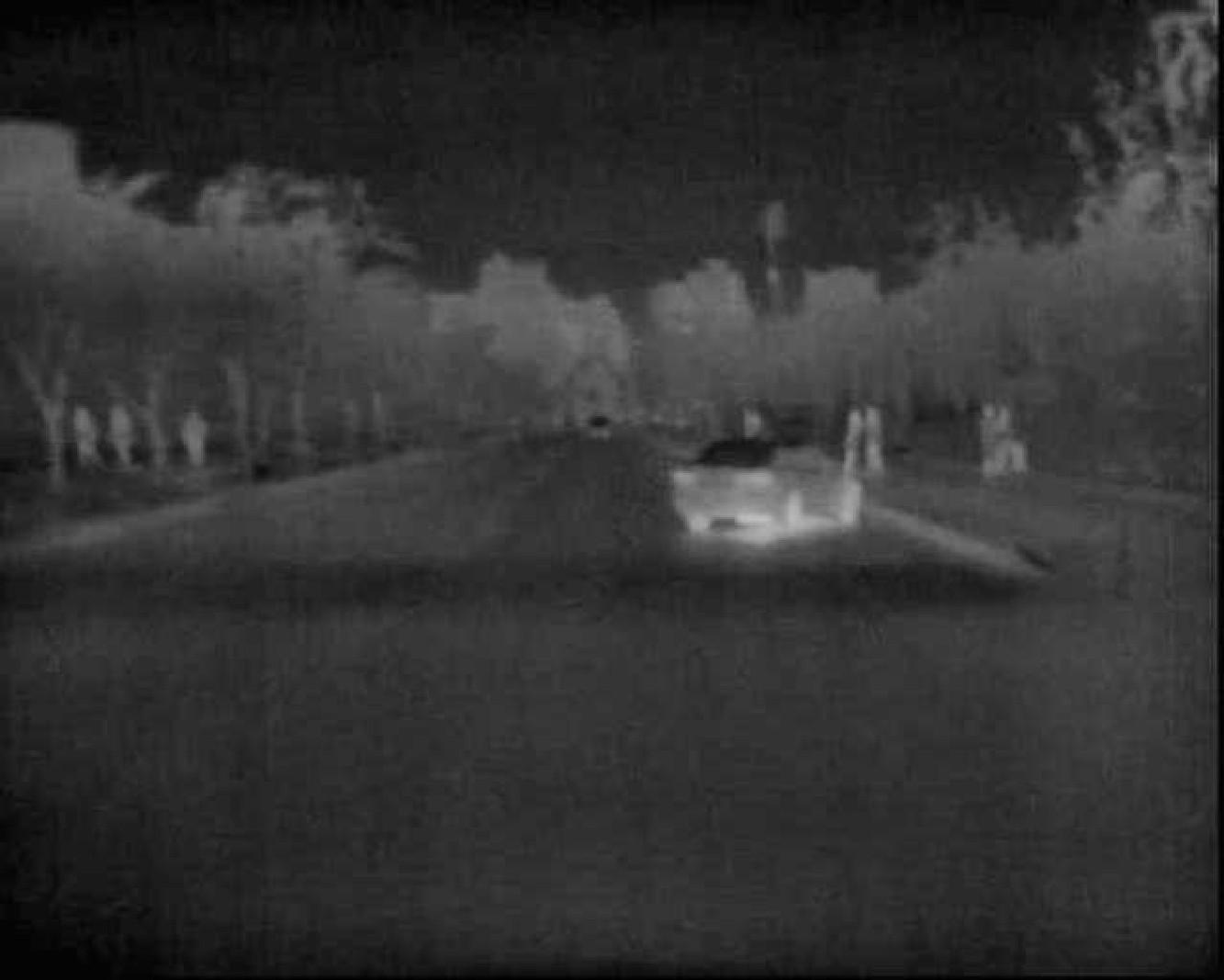}}
	\end{minipage}
	\begin{minipage}{0.16\linewidth}
		\centering
		{\includegraphics[width=1\linewidth,clip]{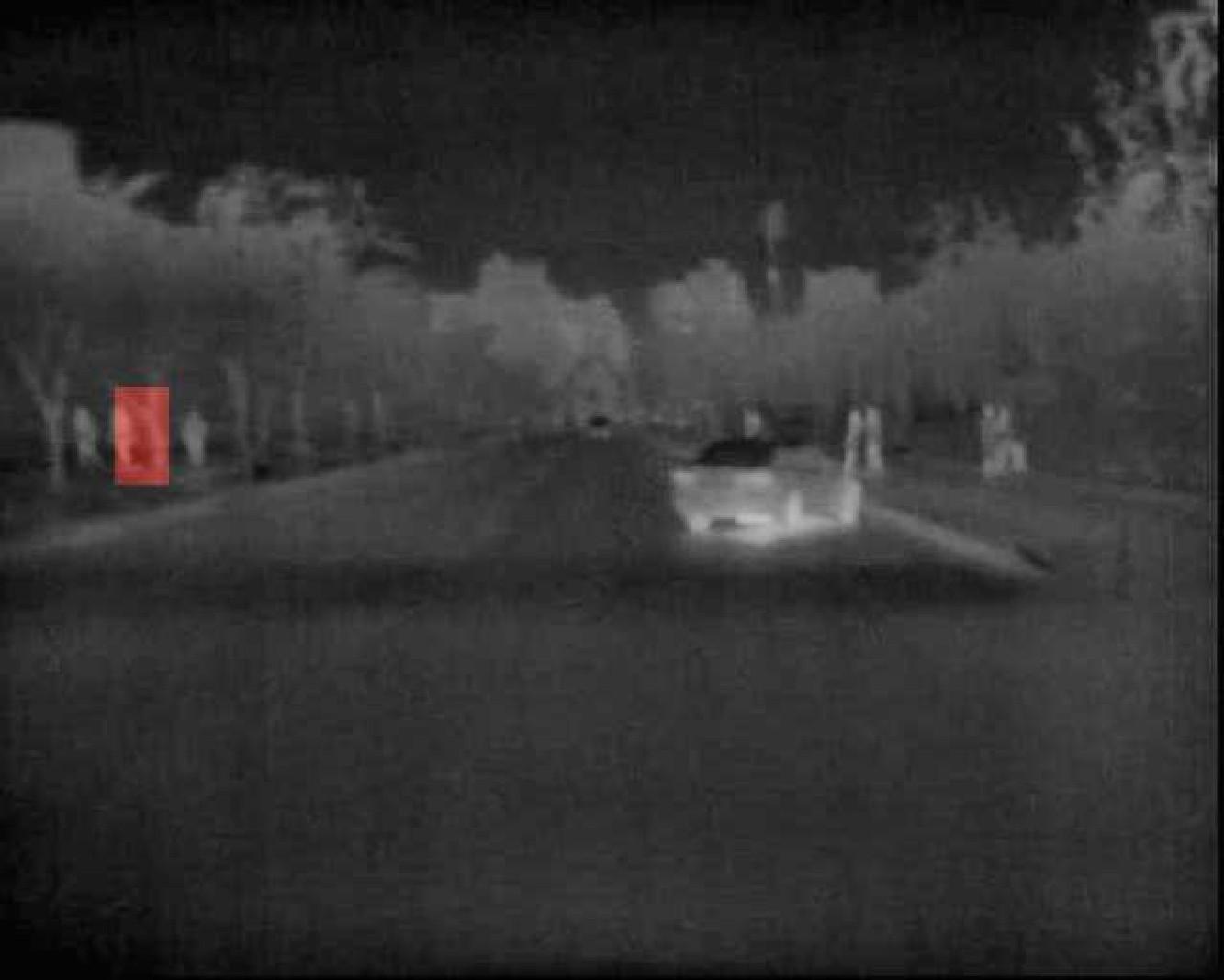}}
	\end{minipage} 	
	\begin{minipage}{0.16\linewidth}
		\centering
		{\includegraphics[width=1\linewidth,clip]{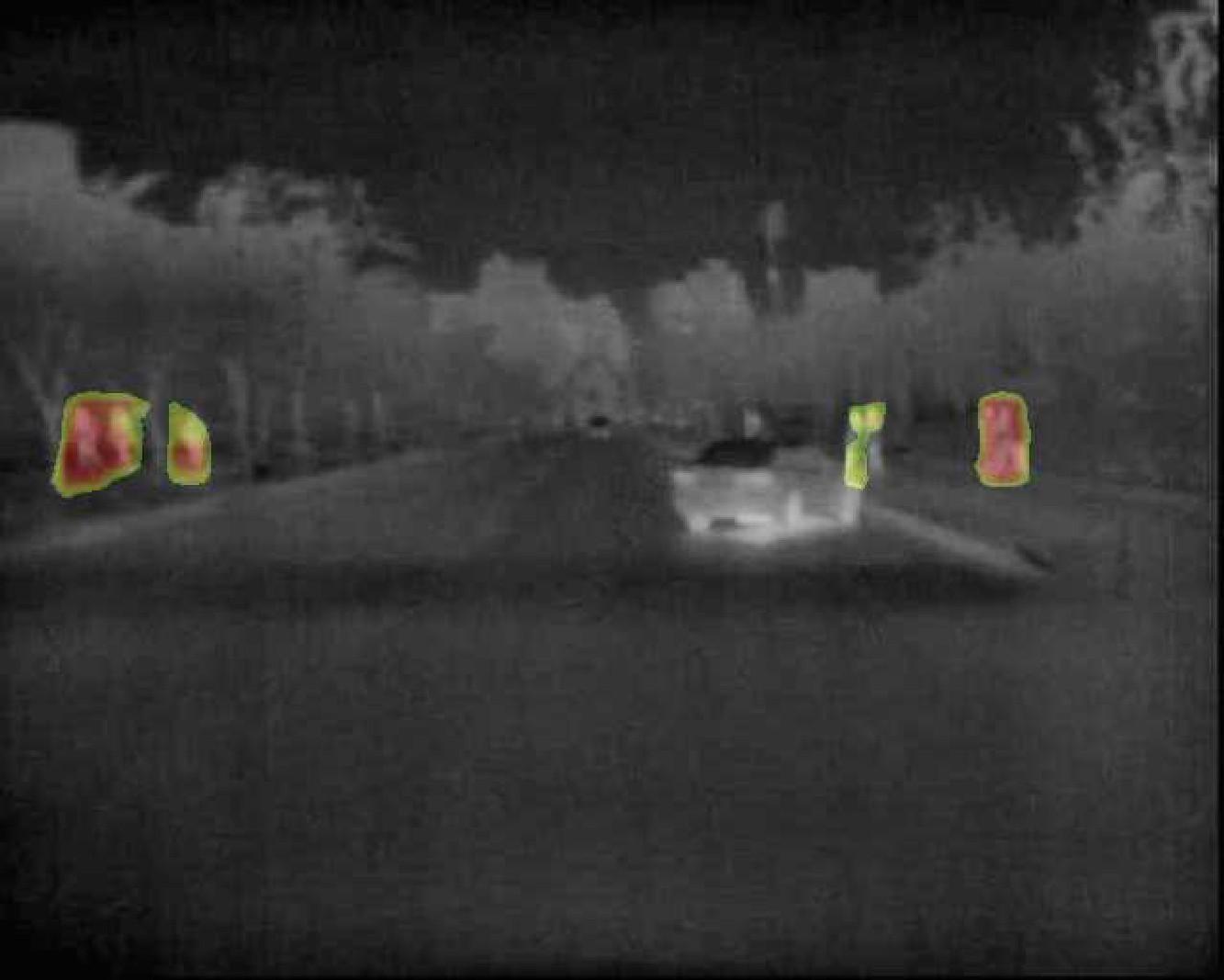}}
	\end{minipage} 
	\begin{minipage}{0.16\linewidth}
		\centering
		{\includegraphics[width=1\linewidth,clip]{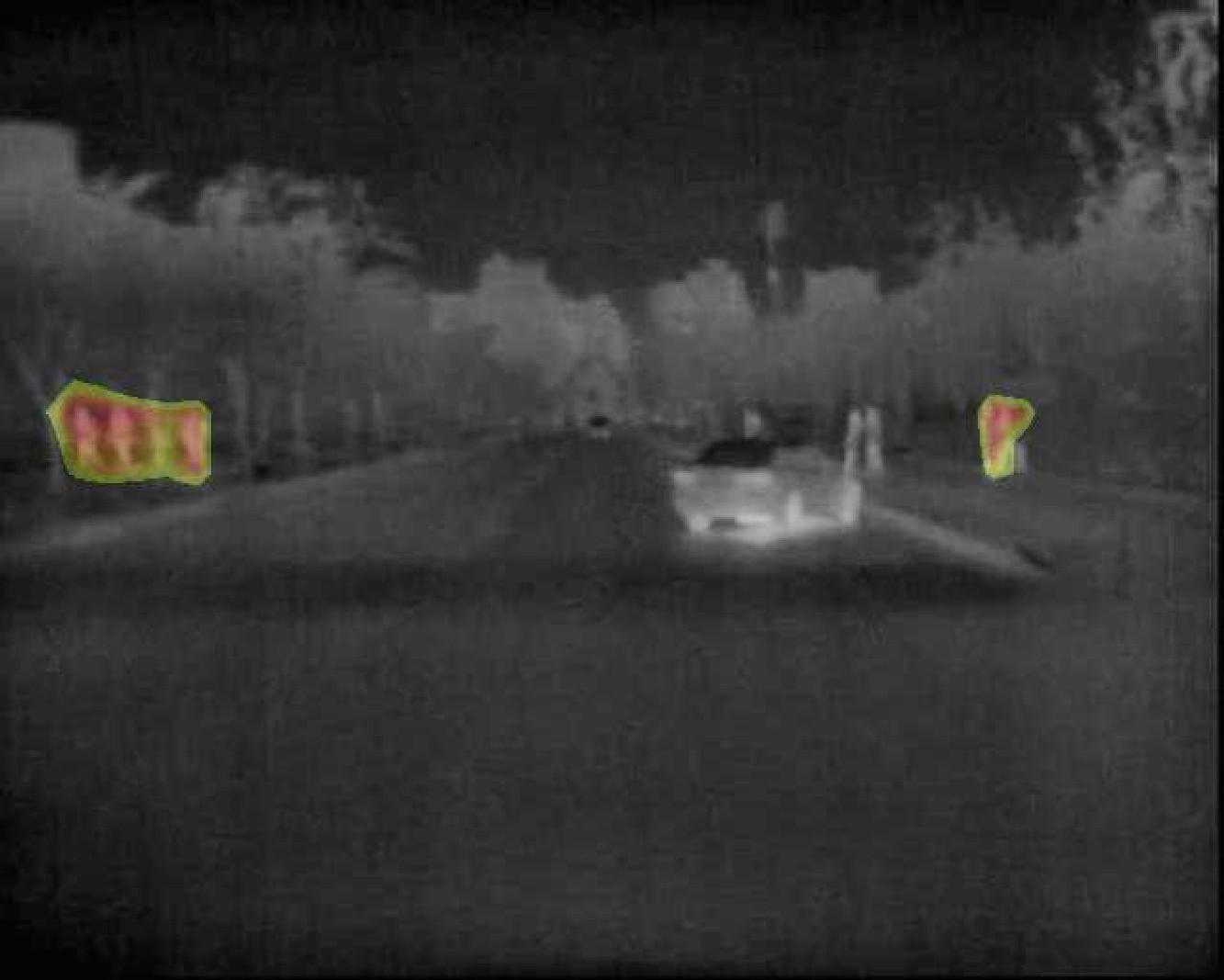}}
	\end{minipage} 
	\vspace{1mm}\\
			{\bf (b) Nighttime}\\
	\caption{Qualitative comparison of multispectral pedestrian detection results in the KAIST testing dataset with other state-of-the-art approaches. First column shows the ground truth (displaying using the visible channel) and the others show detection results of Fusion RPN+BDT \cite{konig2017fully},IATDNN+IAMSS \cite{guan2018fusion}, MSDS-RCNN \cite{li2018multispectral} and our proposed HMFFN-640 and HMFFN-320 respectively (displaying using the infrared channel). Note that the green regions represent ground-truth annotation masks which are generated based on manually labeled bounding boxes, and the detected pedestrian targets are visualized using the heat map representation with a 0.5 threshold. Best viewed in color.}
	\label{fig6}
\end{figure*}

\subsection{Evaluation of Multispectral Feature Fusion Schemes}
\label{MFFNvsHMFFN}
In this paper, we design two multispectral feature fusion schemes (MFFN and HMFFN ). The HMFFN model makes use of skip connections to associate the middle-level feature maps (output of Conv4-V/I layers) with the high-level ones (output of Conv5-V/I layers). We experimentally evaluate the performance gain by incorporating middle-level feature maps into the baseline MFFN model.
The quantitative performance {(pixel-level AP \cite{salton1986introduction})} of MFFN and HMFFN for different sizes of input images ($640 \times 512$, $480 \times 384$, and $320 \times 256$) are compared in Tab.~\ref{tab1}. 

We observe that better detection performance is achieved through the hierarchical multispectral feature fusion. Moreover, the performance gain is more obvious when handling small-size input images. By incorporating the middle-level feature maps, AP index significantly increases from 0.748 (MFFN-320) to 0.817 (HMFFN-320) for $320 \times 256$ resolution input images in the Reasonable-all subset, while the improvement is not obvious for $640 \times 512$ resolution input images (increasing from 0.844 to 0.854). The underlying reason is that the middle-level features from shallower layers (Conv4-V/I) encode rich small-scale image characteristics which are essential for accurate detection of small-size targets. Using a smaller size input image will significantly improve the computational efficiency for real-time autonomous driving applications.

Furthermore, we conduct the qualitative comparison of two multispectral feature fusion networks (MFFN-320 and HMFFN-320) by displaying detection results in various scenes in Fig.~\ref{fig7}. It is observed that performance gains can generally be achieved (in both daytime and nighttime scenes and on different scale and occlusion subsets) by integrating middle-level feature maps with high-level ones. We evaluate MFFN-320 and HMFFN-320 models on testing subsets of different scales. Although both MFFN-320 and HMFFN-320 work well on the near scale subset, HMFFN-320 can better identify pedestrian targets in the medium and far scale subsets through incorporating image details extracted in middle-level layers (Conv4-V/I). Moreover, we test MFFN-320 and HMFFN-320 models on different occlusion subsets and observe that HMFFN-320 generates more accurate detection results when target objects are partially or heavily occluded. A reasonable explanation of this improvement is that low-level features extracted in shallower layers (Conv4-V/I) provide useful information of human parts and their relationships to handle the challenging target occlusion problem \cite{shu2012part}. The experimental results verify the effectiveness of the proposed HMFFN architecture, capable of extracting informative multi-scale feature maps to achieve more precise object detection and remain more robust against scene variations. 

\subsection{Evaluation of Box-level Segmentation Supervised Framework}

In this subsection, we evaluate the performance gain of using box-level segmentation masks instead of anchor boxes to train deep convolutional neural networks for multispectral target detection. For a fair comparison, we make use of the same architecture in HMFFN for multispectral feature extraction/fusion as shown in Fig.~\ref{fig3} (b). Given the multispectral semantic features from Conv-Mul layer, the anchor box based detector RPN \cite{zhang2016faster} is utilized to generate confident scores and bounding boxes as detection results. In comparison, our proposed segmentation mask supervised method computes a prediction heat map to highlight the existence of human targets in a scene. 
The performances {(pixel-level AP \cite{salton1986introduction})} of our proposed box-level segmentation supervised method (HMFFN) and the one based on anchor boxes (RPN-HMFFN) on different sizes of input images ($640 \times 512$, $480 \times 384$, and $320 \times 256$) are quantitatively compared in Tab.~\ref{tab12}.

It is observed that HMFFN based on box-level segmentation masks performs better than RPN-HMFFN based on anchor boxes, achieving significantly higher AP indexes on various testing subsets and on images of different sizes (HMFFN-640 0.854 AP vs. RPN-HMFFN-640 0.756 AP on the reasonable all subset). Such improvements are particularly evident on some challenging detection tasks (HMFFN-640 0.166 AP vs. RPN-HMFFN-640 0.065 AP for far scale human target detection). Another advantage of our proposed HMFFN is that it directly computes a prediction heat map instead of confident scores and coordinates of bounding boxes, achieving faster inference speed (HMFFN-320 38.3 fps vs. RPN-HMFFN-320 32.0 fps).

Furthermore, we qualitatively show some sample detection results of HMFFN-640 and RPN-HMFFN-640 in Fig.~\ref{fig8}. { The output of our method is a full-size prediction heat map in which human target regions yields high confident scores. For a fair comparison, we also transform the bounding box detection results with different prediction scores to the heat map representation, utilizing different colors to show prediction scores of bounding boxes. Note we only show regions with confident scores larger than 0.5.} It is noted that HMFFN-640 generate more precise detection results and fewer false positives compared with RPN-HMFFN-640. The use of anchor boxes involves complex hyperparameter settings (e.g., box size, aspect ratio, stride, and intersection-over-union threshold) will cause severe imbalance between positive and negative training samples and damage the learning of human-related features \cite{law2018cornernet}. Moreover, we observe that HMFFN-640 can successfully identify some pedestrian instances on the far scale and heavy occlusion subsets, which are difficult to detect using the anchor box based RPN-HMFFN-640 or even based on visual observation. For small/occluded targets, it is difficult to generate enough positive samples using discretely distributed anchor boxes. In comparison, our proposed HMFFN takes the easily obtained bounding box annotation as input and produces an unambiguous box-level segmentation mask for learning to distinguish target objects from the background. Overall, our experimental results demonstrate that box-level approximate segmentation masks provide better supervision information than anchored boxes for the training of two-stream deep neural networks to learn human-relative characteristic features. 

\subsection{Comparison with the State-of-the-art}

We compare the proposed HMFFN-640 and HMFFN-320 models with a number of state-of-the-art multispectral pedestrian detectors including Halfway Fusion \cite{liu2016multispectral}, Fusion RPN+BDT \cite{konig2017fully}, IATDNN+IAMSS \cite{guan2018fusion}, FRPN-Sum+TSS \cite{guan2018exploiting}, and MSDS-RCNN \cite{li2018multispectral}. The Fusion RPN+BDT \cite{konig2017fully} model is re-implemented and trained according to the original papers, and the detection results of Halfway Fusion \cite{liu2016multispectral}, IATDNN+IAMSS \cite{guan2018fusion}, FRPN-Sum+TSS \cite{guan2018exploiting}, and MSDS-RCNN \cite{li2018multispectral} are kindly provided by the authors.

The quantitative evaluation results of different multispectral pedestrian detectors are shown in Tab.~\ref{tab2}. Our proposed HMFFN-640 and HMFFN-320 models both achieve higher AP values in all reasonable, scale, and occlusion subset of the KAIST testing dataset. These comparative results indicate that our propose multispectral pedestrian detector achieves more robust performances under various surveillance situations. We qualitatively compare different multispectral pedestrian detectors by visualizing some sample detection results in Fig.~\ref{fig6}. { The output of our method is a full-size prediction heat map in which human target regions yields high confident scores, while the bounding box detection results with different prediction scores are transformed to the heat map representation, utilizing different colors to show prediction scores of bounding boxes. Note we only show regions with confident scores larger than 0.5.} Different from the existing multispectral pedestrian detection methods which generate a number of bounding boxes, our method estimates a full-size prediction heat map to highlight the existence of pedestrians in a scene. It is observed that our approach is capable of generating accurate detection results even for small human targets and using small-size input images.

We also compare the computational efficiency of HMFFN-640 and HMFFN-320 with state-of-the-art methods. A single Titan X GPU is utilized to evaluate the computation efficiency. Please note that the current state-of-the-art multispectral pedestrian detectors \cite{konig2017fully, Liu2016BMVC, guan2018fusion, guan2018exploiting, li2018multispectral} typically perform image up-scaling to achieve their optimal detection performances. For instance, input sizes of \cite{liu2016multispectral}, Fusion RPN+BDT \cite{konig2017fully}, IATDNN+IAMSS \cite{guan2018fusion}, FRPN-Sum+TSS \cite{guan2018exploiting}, and MSDS-RCNN \cite{li2018multispectral} models are $750 \times 600$, $960 \times 768$, $960 \times 768$, $960 \times 768$, and $750 \times 600$, respectively. In comparison, HMFFN-640 directly takes $640 \times 512$ multispectral data as input without image up-scaling thus run much faster (10.8fps vs. 4.4fps). Moreover, our HMFFN-320 model takes small-size $320 \times 256$ images as input and achieves 38.3fps which is sufficient for real-time autonomous driving applications. Please note HMFFN-320 achieves more accurate detection results than the current state-of-the-art multispectral pedestrian detection methods. 

\section{CONCLUSIONS}
\label{conclusion}

In this paper, we propose a powerful box-level segmentation supervised learning framework for accurate and real-time multispectral pedestrian detection. To the best of our knowledge, this represents the first attempt to train multispectral pedestrian detectors without using anchor boxes. Extensive experimental results verify that box-level approximate segmentation masks provide useful information for distinguishing human targets from the background. Also, we design a hierarchical multispectral feature fusion scheme in which the middle-level feature maps (small-scale image characteristics) and the high-level ones (semantic information) are incorporated to achieve more accurate detection results, particularly for far-scale human targets. Experimental results on KAIST benchmark show that our proposed method achieves higher detection accuracy compared with the state-of-the-art multispectral pedestrian detectors. Moreover, this efficient framework achieves real-time processing speed and processes more than 30 images per second on a single NVIDIA Geforce Titan X GPU. The proposed methods can be generalized to other object detection task with multispectral input and facilitate potential applications (e.g., path planning, collision avoidance, and target tracking) in autonomous vehicles.

\bibliographystyle{elsarticle-harv}
\bibliography{manuscript.bbl}

\end{document}